\documentclass{iopjournal}
\usepackage[english]{babel}
\usepackage{amsmath}
\usepackage{subcaption}
\usepackage{graphicx}
\usepackage{csquotes}
\usepackage{orcidlink}
\usepackage{hyperref}  
\hypersetup{
    colorlinks=true,
    linkcolor=black,
    citecolor=black,
    urlcolor=black,
    breaklinks=true  
}
\usepackage[style=apa,backend=biber]{biblatex}  
\begin{document}
\articletype{Topical Review}
\title{More than MACs: Exploring the Role of Neuromorphic Engineering in the Age of LLMs}
\author{Wilkie Olin-Ammentorp$^1$\orcidlink{0000-0002-2472-9862}\\}
\affil{$^1$Mathematics \& Computer Science Division, Argonne National Laboratory, Lemont, USA}
\email{wolinammentorp@anl.gov}

\abstract{The introduction of large language models has significantly expanded global demand for computing; addressing this growing demand requires novel approaches that introduce new capabilities while addressing extant needs. Although inspiration from biological systems served as the foundation on which modern artificial intelligence (AI) was developed, many modern advances have been made without clear parallels to biological computing. As a result, the ability of techniques inspired by ``natural intelligence'' (NI) to inflect modern AI systems may be questioned. However, by analyzing remaining disparities between AI and NI, we argue that further biological inspiration can contribute towards expanding the capabilities of artificial systems, enabling them to succeed in real-world environments and adapt to niche applications. To elucidate which NI mechanisms can contribute toward this goal, we review and compare elements of biological and artificial computing systems, emphasizing areas of NI that have not yet been effectively captured by AI. We then suggest areas of opportunity for NI-inspired mechanisms that can inflect AI hardware and software.}

\section*{Introduction: The AI Explosion}

Large language models (LLMs) are neural networks consisting of billions of parameters (or more) that can carry out numerous natural language tasks with high levels of skill---meeting, or in some cases exceeding, human performance in certain tasks~\parencite{hendrycks_measuring_2021, bubeck_sparks_2023, katz_gpt-4_2023}. The utility of LLMs has expanded from the domain of natural language to include audio, images, video, and more~\parencite{rombach_high-resolution_2021, anil_gemini_2023, brooks_video_2024}. However, issues in modern AI persist: these systems' long-term learning capabilities are limited, novel sources of training data are becoming scarce, capital and operational expenditures required for AI hardware remain high, and high performance on benchmarks has not consistently translated into real-world success,~\parencite{villalobos_will_2024, epoch2024trainingcomputeoffrontieraimodelsgrowsby45xperyear, reuters_microsoft_openai_2024, economist_staff_data_2024}. Future evolutions of AI technologies seek to address these issues, enabling deployment of AI systems capable of more diverse and unpredictable tasks and enabling them to safely interact with, learn from, and explore the physical world.

Biological systems are firmly rooted in interaction with the environment and address challenges such as limited data, new tasks, and uncertainty while thriving in the natural environments that remain challenging for AI. Given this contrast,  it seems a simple conclusion to draw is that the development of physical AI could benefit from the inclusion of biological computing methods that enable ``natural intelligence'' (NI). However, although many foundations of modern LLM systems, such as connectionist networks, were founded from biological principles, many more modern developments have been developed without explicitly adhering to this source of inspiration. In light of this contrast, a key question which arises is ``what, if any, principles can be leveraged from NI to address issues with today's AI systems?''

Applying principles from NI has occasionally been criticized as the art of ``gluing feathers to the wings of an airplane.'' In other words, copying principles that appear effective in nature to engineered, artificial systems is superficial and does not address underlying differences in applications, materials, and commercial viability. For instance, while birds and airplanes are both capable of heavier-than-air flight, only airplanes are capable of transporting hundreds of humans at hundreds of miles per hour 35,000 feet above the Pacific Ocean. Nevertheless, biological systems can persevere and thrive in niches unaddressed by engineered solutions; until the development of drones, airplanes could not fill the role of delivering small, individual parcels from one point to another---a role successfully filled for centuries by carrier pigeons. 

Today, AI fills a similar role as large aircraft in this analogy: developing and deploying AI models at scale is expensive in capital, human, and energetic costs. Just as no average individual can afford a Boeing 747, ``homebrewing'' a capable LLM is---currently---infeasible. But in contrast to flight, NI remains capable of performing all the same tasks as an AI, albeit not as quickly or with equal access to information. 

In this work, we posit that rather than focusing on the ability to more efficiently calculate operations --- such as multiply-accumulates (MACs) --- which are already being aggressively optimized by industrial concerns, biologically-inspired approaches are more likely to significantly inflect the trajectory of AI systems by focusing on other areas of opportunity. These include exploring radically different approaches to integrating dense, inexpensive memory with compute, alternative methods for in-situ learning, and avoiding conversions between the analog and digital domains. Developments in these areas could lead to entirely new classes of AI systems which could be effectively employed outside of datacenters, opening new application domains to development. To justify this position, we begin by re-examining the relative properties of biological and artificial systems and examine the resulting conditions under which these systems are effective.

\section{Re-Examining Biological Efficiency}\label{sec_bio_efficiency}

A long-standing motivation of neuromorphic engineering stems from the high efficiency of biological systems carrying out complex tasks \parencite{mead_how_2020, schuman_survey_2017, christensen_2022_2022}. For many years, this claim was self-evident, as in many areas --- such as image recognition and natural language processing (NLP) --- artificial systems could not replicate the capabilities of NI. However, with the introduction of deep neural networks, LLMs, and ``chat'' agents, both AI systems and humans are now capable of carrying out advanced tasks such as textual conversation and code completion. This advance allows us to re-evaluate the relative efficiencies of artificial and biological systems. 

\subsection{Token-Level Efficiency}

One argument that AI has not yet reached energy efficiency competitive with NI is based on rates of energy consumption. While a modern processor such as an NVIDIA H100 requires hundreds of watts (J/s), metabolic measurements estimate the human brain consumes approximately 20 watts ~\parencite{balasubramanian_brain_2021}. However, this comparison neglects the relative throughput of these systems. To provide a fair comparison, we must normalize against the amount of informational ``work'' carried out by each system. 

A significant amount of labor in the modern economy centers on NLP:  sharing, explaining, and summarizing knowledge expressed verbally or in writing. Expert knowledge often consists of translating information between natural language and specialized domains, such as designers taking a client’s requests and feedback to create visuals or radiologists examining MRI images to respond to a request for a diagnosis. LLMs excel at NLP tasks, and the number of alternate domains (such as images) they can interact with skillfully continues to increase~\parencite{anil_gemini_2023}. 

\begin{figure}[htbp]
    \centering
    \includegraphics[width=1\linewidth]{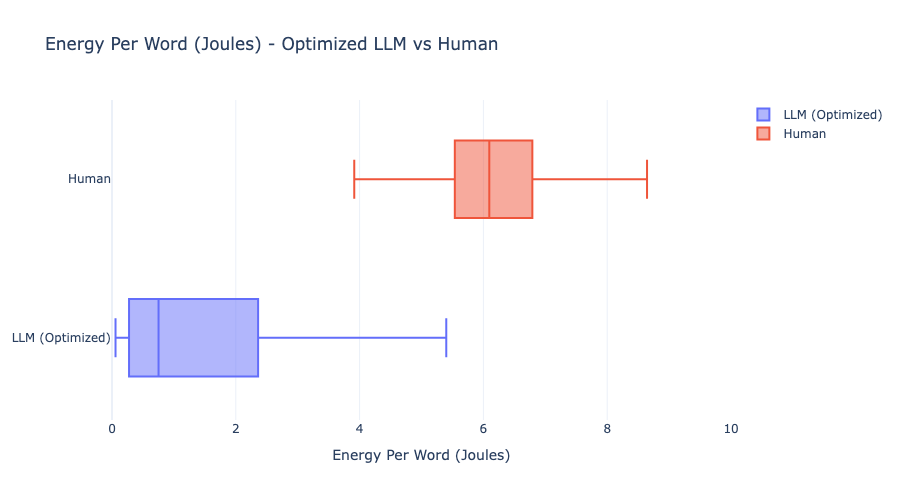}
    \caption{Estimates of the average efficiency of humans at processing natural language derived from speech rates and metabolic limits suggest that LLMs running on commodity GPU hardware currently reach similar or greater efficiencies. Estimates for LLM efficiency are sourced from 12 LLMs publicly benchmarked via ML.Energy \parencite{mlenergy-neuripsdb25}.}
    \label{fig:energy_comparison}
\end{figure}

Humans converse at a median rate of approximately 200 words per minute, with most rates of speech falling between 100 and 300 words per minute ~\parencite{yuan_towards_2006}. This rate of information---on the order of ten bits per second---applies not only to speech but also to many other tasks, such as solving a Rubik's cube or playing video games~\parencite{zheng_unbearable_2025}. By assuming speaking rates follow a normal distribution between 100 and 300 words per minute and dividing by the metabolic budget for the human brain, we arrive at a median estimate of 6 joules expended per word. While the brain does not utilize its entire metabolic budget for processing words (as is the case for a GPU), and in fact is simultaneously maintaining homeostasis, processing visual and auditory stimuli, and so forth, these processes are not ``optional'' and must be considered part of the total energy budget per token. 

LLMs do not interact natively with words, but accept digital text inputs and convert them to “tokens”---points in a high-dimensional space representing the semantics of natural language~\parencite{pennington_glove_2014, wu_googles_2016}. Generally, tokenizers represent words with an average of two tokens ~\parencite{google_gemini_tokens, gptspace_gpt_tokens_guide}. By using public benchmarks of the energy expended per token and multiplying by the approximate conversion factor of two tokens per word, we can estimate the same energetic cost expended per word in AI systems ~\parencite{mlenergy-neuripsdb25}. If humans still maintain a distinct advantage in efficiency over artificial systems, we would expect the approximate ``energy per word'' to be distinctly less for humans than for AI. However, we see that the opposite is actually true: compared with humans, LLMs executing on commodity hardware (NVIDIA H100 or B100) use distinctly less energy per token (U-Test, $p<0.001$) (Figure \ref{fig:energy_comparison}). However, the high efficiencies of these AI systems are dependent on a number of conditions which we will soon explore in more detail (Section \ref{sec_memory}).

\subsection{Synaptic-Level Efficiency}
Our previous ``top-down'' estimate on joules per word can be challenged on several levels. For instance, it assumes that language is a necessary basis for effective thought. To complement this imperfect comparison, we supplement it with an independent ``bottom-up'' estimate of the energy dissipated per each synaptic operation in artificial and biological systems. While comparing the temporal, stochastic release of neurotransmitters through a synapse to a simple multiplication of two real-valued numbers is undoubtedly reductive, this equation is the basis of modern AI systems and we leave a detailed discussion of the impacts of this reduction to a later section (\ref{sec_synapses}) ~\parencite{rosenblatt_perceptron_1958, lecun_deep_2015}.

Artificial systems provide clear benchmarks on achievable ``synaptic'' operations (ops.) per second, with maximum and effective number of ops. per watt publicly available~\parencite{reuther_lincoln_2023, EpochMachineLearningHardware2024}. These figures can also be calculated for the human brain by estimating its total power (20 W), mean spiking rate (0.7 Hz), number of neurons (86 billion), and number of synapses (between 0.1 and 1.0 quadrillion)~\parencite{balasubramanian_brain_2021,herculano-houzel_human_2009, bahney_search_2017, buzsaki_log-dynamic_2014, Koch2024}. By multiplying mean spiking rate by the mean number of synapses per neuron and dividing by metabolic energy, we estimate the efficiency of synaptic operations in the brain between 3.5 and 35 tera-ops. per watt (TOPS/W). 

\begin{figure}[htbp]
    \centering
    \begin{subfigure}[t]{0.48\linewidth}
        \includegraphics[width=\linewidth]{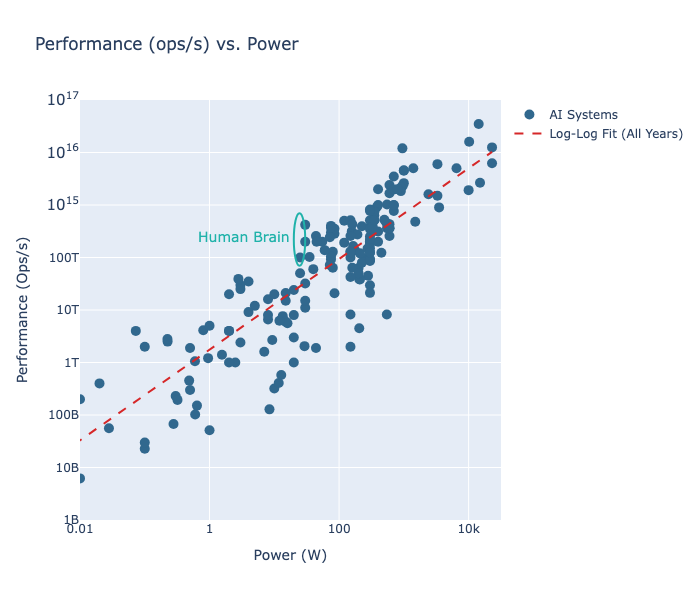}
        \caption{Number of ops. sustained per second versus power dissipation for AI accelerators, colored by year.}
        \label{fig:tops_w_a}
    \end{subfigure}
    \hfill
    \begin{subfigure}[t]{0.48\linewidth}
        \includegraphics[width=\linewidth]{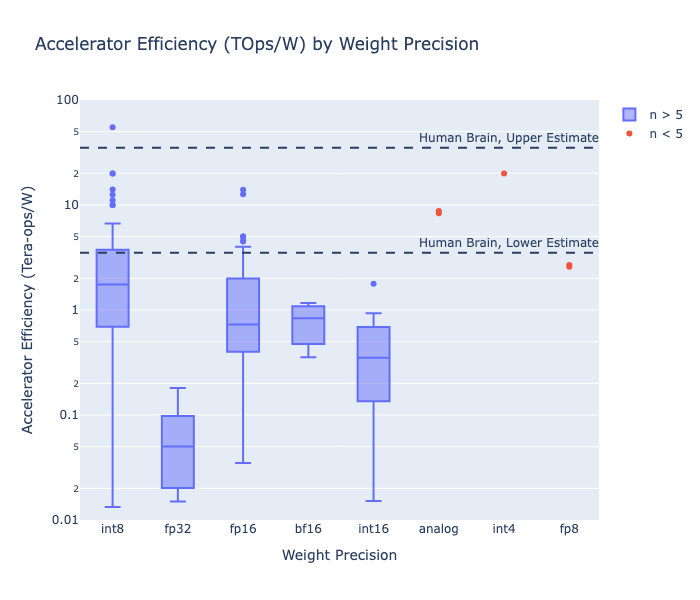}
        \caption{Per-operation (op) efficiency estimates of over a hundred AI accelerators, grouped by data representation type.}
        \label{fig:tops_w_b}
    \end{subfigure}
    \begin{subfigure}[t]{0.48\linewidth}
        \includegraphics[width=\linewidth]{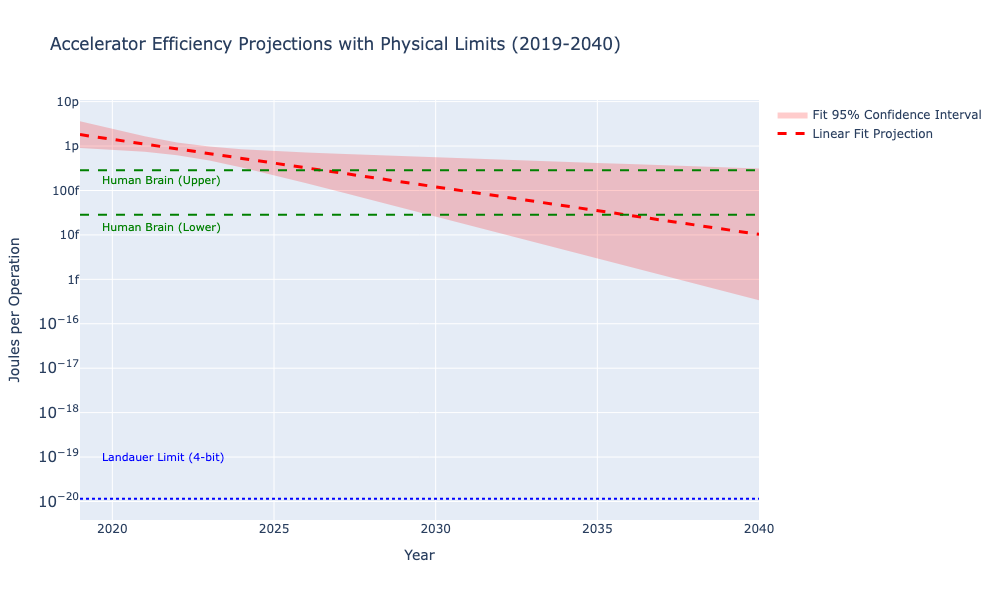}
        \caption{Extrapolating current trends, a digital multiplication could dissipate equivalent or lesser energy as a synaptic operation over the next decade.}
        \label{fig:tops_w_c}
    \end{subfigure}
    \caption{By leveraging alternative coding strategies for precision of weights and activations, AI accelerators have made large strides in per-op efficiencies, and some approach estimated per-synapse efficiencies for the human brain. Data from 170 AI accelerators is sourced from the Lincoln AI Computing Survey \parencite{reuther_lincoln_2023}.} 
    \label{fig:tops_w}
\end{figure}

Of a survey of 170 AI accelerators, the majority of these systems remain below even the lower estimate on this efficiency threshold (Figure \ref{fig:tops_w_a}). However, some systems --- particularly those which leverage lower-cost representations of numbers such as 4- and 8-bit integers --- approach or exceed the lower limit of 3.5 TOPS/W (Figure \ref{fig:tops_w_b}). Furthermore, continued progress in hardware and software is likely to reduce this gap to the point where it may cease altogether over the next decade (Figure \ref{fig:tops_w_c}) ~\parencite{dally_hardware_2023}. 

From these comparisons of the increasingly comparable per-token and per-op efficiencies of AI and NI, it may appear that the gap between the two is rapidly closing---or already has closed. As impressive as this progress is, however, it leaves out many important factors: high costs of artificial memory, a resulting dependency on batching, and limited learning capabilities. We now detail and quantify these factors in greater detail to argue that they --- more than joules per token or synaptic operation --- are issues which must be addressed through novel approaches.

\subsection{Memory Constraints}
\label{sec_memory}
In practice, the execution of LLMs is often bottlenecked not by available compute, but instead by available memory \parencite{mlenergy-neuripsdb25}. Static Random-Access Memory (SRAM), constructed from 6 transistors, is extremely fast and is commonly integrated as caches colocated with compute units. However, SRAM cells require a relatively large amount of area which has not decreased significantly with recent semiconductor production processes \parencite{schor_iedm_2022}. As a result, many compute systems such as graphics processing units (GPUs) co-package slower, denser dynamic random access memory (DRAM) in high-bandwidth memory (HBM) packages. For instance, the NVIDIA H100 contains 34 MB of low-level registers, 50 MB of Level 2 Cache, and 80 GB of HBM memory. For an LLM with 10 billion (100B) 8-bit parameters, 10 GB of information plus intermediate results must be stored - the vast majority of which at any time is resident in the highest level DRAM package, with variables relevant to the current step of computation streaming at extremely high rates (up to 3.3 TB/s) through the lower level cache and registers  \parencite{nvidia_h100_gpu_whitepaper}. Despite advances in co-packaging and high-bandwidth memory, data movement through this hierarchy carries significant costs and greatly influences the execution strategy used for large models \parencite{jun_hbm_2017}.

Alternative architectures to GPUs are one potential method to reduce data transit during execution and avoid the associated energetic costs. Systolic arrays are one such architecture, in which an array of independent cores allow program parameters remain resident in local memory, passing only transient execution data. However, these systems remain dependent on the hierarchy and integration of current memory technologies, with most utilizing fast but low-density SRAM. Even architectures which have been scaled to entire wafer-scale systems, such as the Cerebras CS-3, may require external memories to supplement available caches \parencite{hall_training_2023}. Alternately, multiple systems may be linked together to provide sufficient memory to store program data \parencite{abts_groq_2022}. However, the necessity of these work-arounds demonstrates that even architectures designed specifically to avoid the ``memory wall'' remain constrained by current technologies. 

\begin{figure}[htbp]
    \centering
    \includegraphics[width=0.70\linewidth]{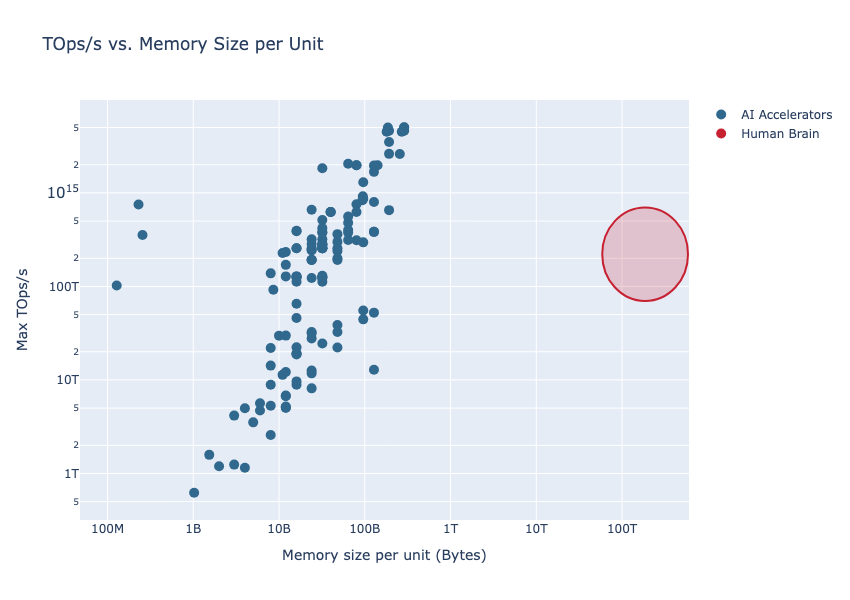}
    \caption{Computational systems optimized for execution of LLMs and other parallel programs contain variable amounts of memory, from megabytes to gigabytes. Available memory often corresponds with computational throughput, with exceptions including Groq's LPU and Facebook's MTIA (left). While the largest of these 170 systems can store 288 GB of data, synapses in the human brain encode 200-2000 times more information.}
    \label{fig:memory}
\end{figure}

As a consequence of the constraints of memory technologies available today, no individual artificial system contains memory comparable to the long-term capacity of the human brain (Figure \ref{fig:memory}). Multiplying estimates of the number of synapses in the human brain (0.1-1.0 quadrillion) by entropic estimates of synaptic variability (4.7 bits), we estimate the long-term informational content of the brain (disregarding transient electronic \& chemical gradients) between 59 and 590 TB; this far exceeds the amount of information available in any current AI accelerator \parencite{Koch2024, bartol_nanoconnectomic_2015}. Given this enormous divide in available memory and the requirement for data transport between caches, how is it possible that operations in GPUs can approach or exceed per-token energy efficiency with the brain? Effectively, this is only possible given a major condition on the execution of LLMs: parallel, or ``batched,'' execution.

\begin{figure}[htbp]
    \centering
    \begin{subfigure}[t]{0.48\linewidth}
        \includegraphics[width=\linewidth]{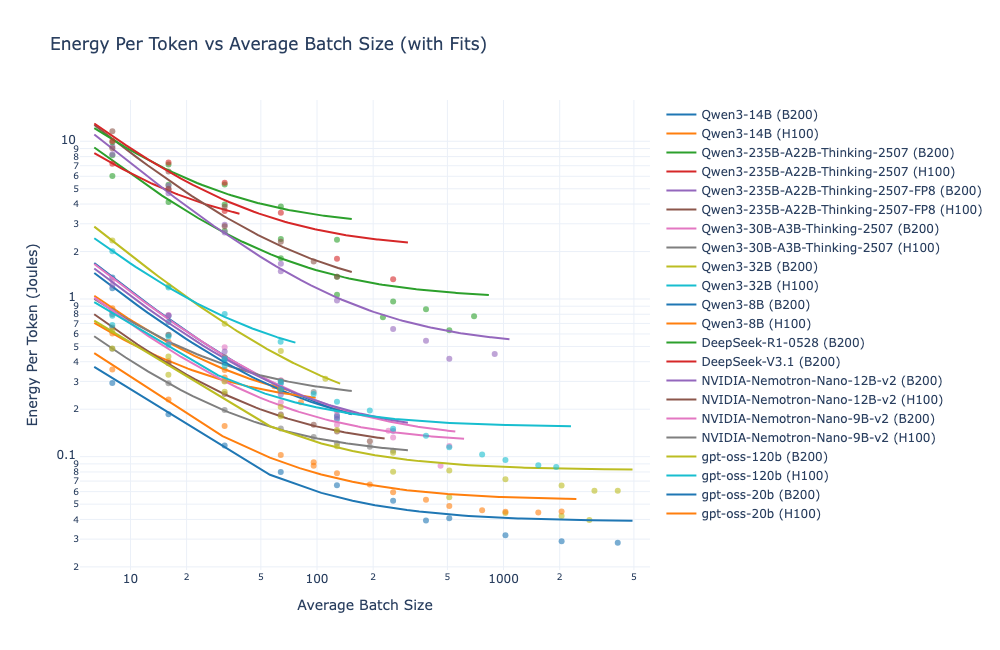}
        \caption{The energy per token of 12 LLMs is plotted against the batch size used during execution on an NVIDIA H100 or B200. }
        \label{fig:batching_latency_a}
    \end{subfigure}
    \hfill
    \begin{subfigure}[t]{0.48\linewidth}
        \includegraphics[width=\linewidth]{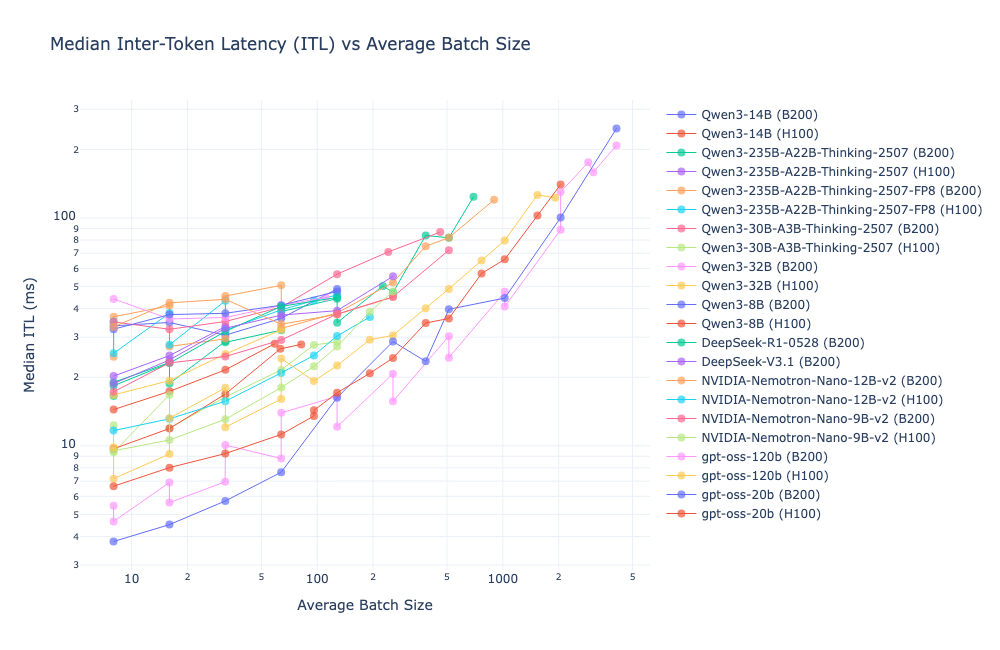}
        \caption{The median inter-token latency (ITL) of 12 LLMs is plotted against the batch size. Larger batches require more memory and movement between caches, increasing the time taken between successive outputs of the model.}
        \label{fig:batching_latency_b}
    \end{subfigure}
    \caption{``Batching'' multiple inputs together to be calculated simultaneously in a parallel processor allows the cost of moving information through its memory to be amortized (a). However, this comes at the cost of increasing the latency between successive outputs of the model (b). Data plotted is sourced from the public ML.Energy benchmark \parencite{mlenergy-neuripsdb25}.}
    \label{fig:batching_latency}
\end{figure}

Batched execution in AI hardware systems allows simultaneous processing with multiple inputs. By extending parallel execution technologies over multiple inputs as well as multiple parameters, the cost of loading program data from memory can be amortized. Given a number of users presenting requests to an AI program at random times, this creates a trade-off between latency and efficiency: by combining more inputs, increased batching can be used to lower the average cost of the movement of static program data at the cost of more computational overhead and end-user responsiveness. This is illustrated quantitatively in Figure \ref{fig:batching_latency}. This relationship can be simply modeled by a linear transformation of batch size divided by its value ($E_{token}(b) \approx (mb+k)/b$). The optimal batch number may shift by the specific model used, but each model exhibits a trade-off between an under-batched regime where loading static program data dominates energy usage, and an over-batched regime where increasing the number of inputs causes higher latency without decreasing average energy usage due to the increase in computational overhead. High efficiency per token and low inter-token latency are thus dependent on finding the optimal point on the Pareto frontier between these two regimes \parencite{semianalysis_inference_race_2023}. 

In contrast to AI, the human brain remains efficient without using batching while providing constant low-latency processing and containing orders of magnitude more possible ``parameters'' than AI~\parencite{thorpe_speed_1996}. Furthermore, while parallel processing in AI enables simultaneous processing of multiple inputs, parallelism in NI takes a wider variety of forms; while handling parallel, independent requests in the same manner as AI is not possible for humans, other forms of parallelism such as integrating information from multiple sources, planning, and execution of actions are accomplished simultaneously~\parencite{pashler1994dual}. Additionally, in contrast to AI, NI does not require explicit ``training'' and ``inference'' phases; information acquired from the environment can be immediately integrated and applied. This inability can greatly increase the cost of deploying a model in environments with many unknowns, where the system must ``ground'' new hypotheses to gain information via testing or integrate new behaviors and/or sources of information. We next quantify these costs for comparison against NI. 

\subsection{Learning Costs}

The conditions which spurred the development of advanced learning capabilities in NI and AI stem from different motivations. In biology, an evolutionary perspective suggests that the capability to learn and modify behaviors over time is useful to help living creatures adapt and survive in an uncertain and dangerous environment. Following this motivation, humans' learning capabilities span over multiple timescales: these include gaining short-term memories through direct interaction with the environment or instruction (seconds), integrating knowledge through sleep (hours), acquiring long-term skill improvement and muscle memory (days to  months), and modifying the ``foundational'' model of the brain's genetic code through epigenetic markers and, ultimately, evolution (years to decades)~\parencite{kudithipudi_biological_2022}. By multiplying these time periods with an average metabolic rates for the entire human body (approximately 100 W) --- as the entire body may be involved in generating signals and gathering information necessary for learning --- we approximate the costs of learning in humans~\parencite{henry_basal_2005}. 

The emergence of modern AI systems stems from a different set of motivating factors than survival in a natural environment. Given the rise of digital computing, recordkeeping, the internet, and telecommunications, the amount of digital  data available has grown exponentially. Tools to navigate this mountain of data, such as search engines, are highly valuable. But while ``classic'' (pre-LLM) search engines allowed humans to navigate data, supplementary capabilities such as summarizing information, synthesizing reports, and constructing new insights were limited. The latent need for automated tools to fill this niche, combined with the ever-growing amount of data, created the conditions for massively-parallel computing hardware and autoregressive neural networks to win the ``hardware lottery'' by creating solutions to these issues using available hardware \parencite{hooker_hardware_2021} . 

\begin{figure}[htbp]
    \centering
    \includegraphics[width=0.70\linewidth]{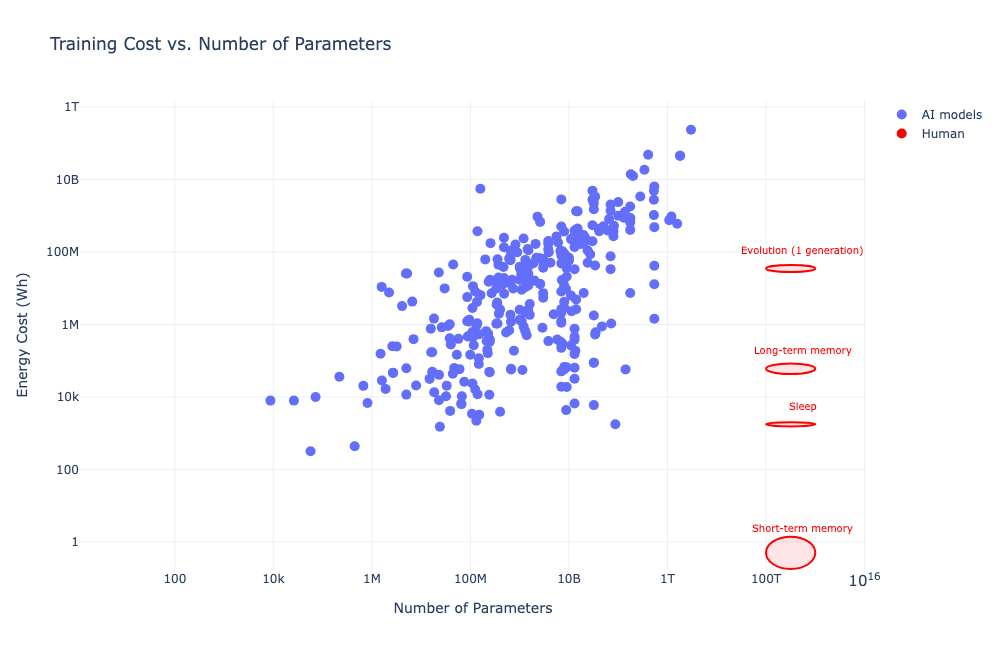}
    \caption{Costs to train AI models `from scratch' scale with the number of parameters utilized in the model, and reach into billions of watt-hours (Wh). In contrast, the brain has numerous learning mechanisms available acting in tandem to enable learning at a variety of timescales with a minuscule cost when averaged over the entirety of the brain.}
    \label{fig:training}
\end{figure}

Specifically, extant algorithms for allowing the error between prediction and truth to be gradually minimized scaled well on parallel systems; this allowed massive compute resources to create language models trained to approximately recreate the available corpus of human language in years, rather than millennia, and extended these networks to assistive ``chat'' agents to carry out a variety of tasks. However, this training is resource intensive, requiring millions or billions of watt-hours (Wh) to be expended in order for a model to be trained~\parencite{epoch2024trainingcomputeoffrontieraimodelsgrowsby45xperyear}. Using public data on the number of parameters, required training time, and energy usage of LLMs, we compare the approximate cost of training AI models from scratch with a variety of learning mechanisms available within the brain (Figure \ref{fig:training}).

An extreme divide remains between the efficiency of training AI systems and learning in NI, with 25 years of human life (approximately a generation) being energetically equivalent to training a 10B parameter model. This may appear artificially inflated by plotting all learning methods available to the brain as involving all possible synapses, but this points towards one of the key advantages of these methods: while they have the potential to explore an enormous space of possible changes in the brain, these mechanisms interplay to selectively update only regions of the brain involved in specific experiences (Section \ref{sec_modules}). Despite the exploration of a variety of alternative algorithms and architectures, no method enabling a self-selective, efficient, and effective update over all parameters in modern LLM systems has yet emerged~\parencite{ororbia_brain-inspired_2023}. As a result, the ``brute force'' approach of applying backpropagation and variations of gradient descent over all parameters and all available data remains the (expensive) choice for producing LLMs and many related architectures. 

\subsection{Summary}
\label{sec_intro_summary}

Both NI and AI are capable of processing highly complex natural information and producing useful responses. However, the environments in which they can carry out this feat vary greatly. Being embedded at a single point moving through space and time and interacting with noisy, analog inputs, NI does not suffer from the inability to simultaneously process multiple, independent chains of thought, but it does require the ability to rapidly change responses through time to survive. Given the same context, knowledge from previous episodic experiences can immediately affect responses. Even without feedback, responses to the same situation may change, and the underlying computation has a large stochastic component~\parencite{gerstner2002spiking, habenschuss_stochastic_2013}. 

In contrast, to ``thrive'' in the environment of a data center, AI must be able to consistently process many digital inputs from a dozen independent users or more to amortize the cost of loading and storing data. While the users' perspectives remain the same, thousands of queries from different positions, times, and contexts are collated into a single batched workload executing across one or more systems. Similarly, data is not ``learned'' serially to build up knowledge incrementally; all possible knowledge is ingested simultaneously with the goal of being reproduced. One emergent property of this process is the potential for linking and generalization of concepts as training progresses, but another is the production of ``hallucinatory'' responses ungrounded in reality. 

To truly enable future goals such as constructing self-driving laboratories or exploring ``physical'' AI --- where automation, AI, and robotics can enable automated discovery of new materials, structures, and more --- artificial systems must be able to move beyond their current niche. Real-world systems involving feedback must be able to react in time to unpredictable stimuli and learn to cope under changing conditions. Meanwhile, communication bandwidth, space, cooling, and available power cannot be supplied as freely as they are in the datacenter. 

In this section, we shown that while LLM's efficiencies have grown tremendously over previous years, requiring feats such as unbatched processing and rapid, online learning remains infeasible. Toward the goal of enabling these or other advanced AI systems to one day address these needs, we examine the ways in biological and artificial agents have---or have not---converged to achieve effective and efficient computation. 
\section{Comparing Biological and Artificial Computation}\label{sec_comparing}

Although knowledge of how information is processed within the human brain and general NI remains incomplete, decades of groundbreaking research have revealed key fundamental mechanisms and effectively served as the inspiration for connectionist AI models that form the foundation of modern techniques ~\parencite{rosenblatt_perceptron_1958, lecun_deep_2015, vaswani_attention_2017}. As a result, the basic features of NI may already be surprisingly familiar to practitioners of modern deep learning, who build networks from \textit{synaptic} weights and \textit{neural} activation functions.

As the success of connectionist AI exploded, it entered a phase of rapid evolution in both software architectures and hardware platforms. Although some of developments in these areas were explicitly informed by biology, many were not, stemming from non-biological domain knowledge, mathematical reasoning, and empirical success. Despite this departure in origins, however, can artificial systems---by pursuing the goal of efficient processing of natural information---converge on principles similar to biological systems? We present an overview of established principles of biological computation, organized by scale, and compare them against relevant AI primitives. These principles are separated into three broad categories: physical substrates, the signals they express and/or process, and methods used to modulate these signals. 

\subsection{Synapses}\label{sec_synapses}

\begin{figure}[htbp]
    \centering
    \includegraphics[width=1\linewidth]{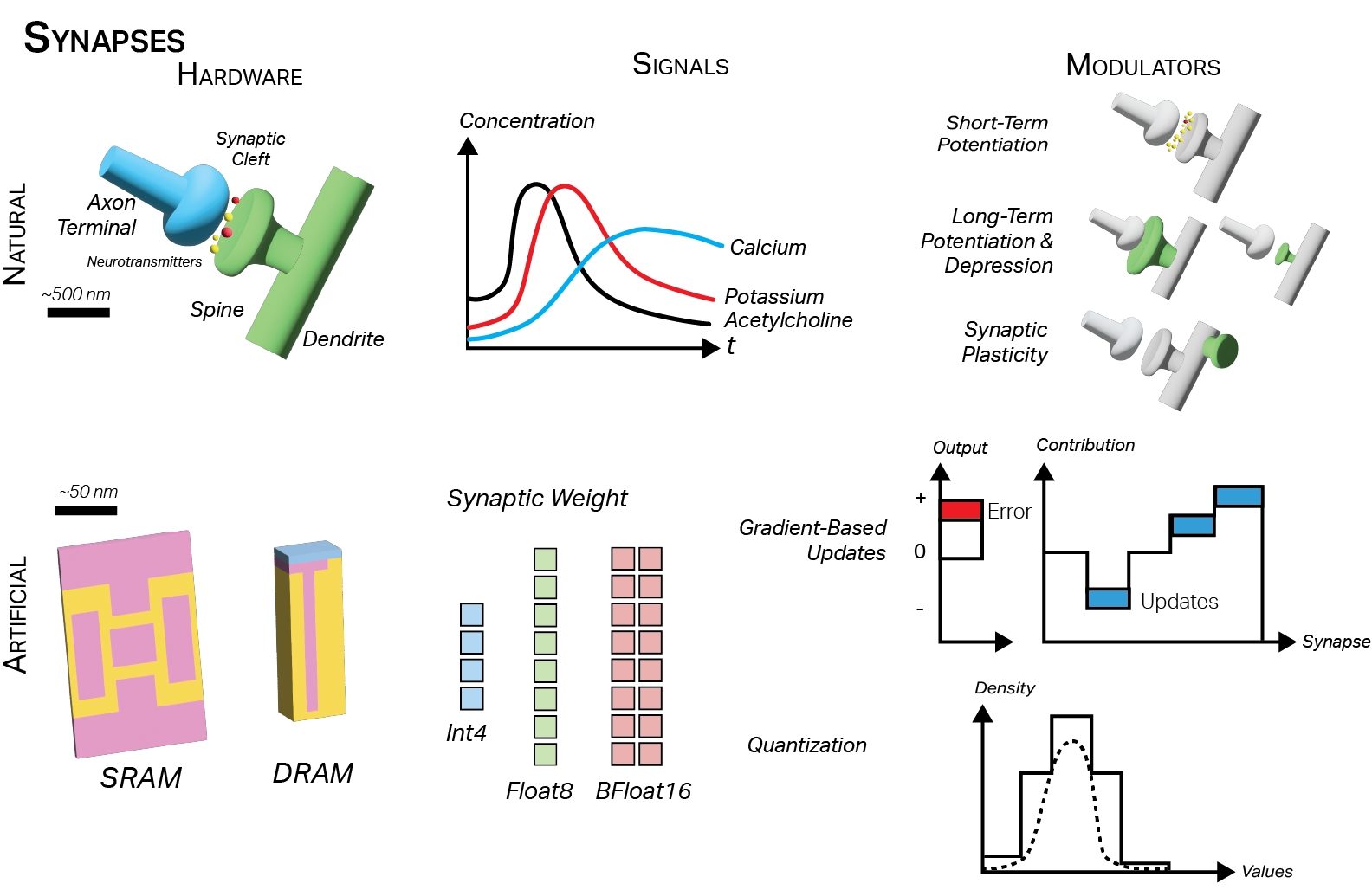}
    \caption{Synapses modulate the strength of connections made in neural networks. In NI, synapses consist of directional chemical, electrical, and physical junctions made to a neuron. The majority of synapses are neuron-neuron connections, in which one neuron may encourage or discourage another neuron to fire. This influence is achieved primarily through the release of neurotransmitters from the presynaptic neuron (blue) across the synaptic cleft. Neurotransmitters (such as glutamate) may bind to the postsynaptic neuron (green) and influence its short-term behavior by manipulating the ionic concentration between sodium and potassium maintained across neuronal membranes. Additional signals, such as calcium or norepinephrine, may vary in their concentration more slowly. This information can be used to track the long-term activity of a synapse and provide feedback to vary synaptic strength through its receptivity to chemicals, physical shape, and more. Artificial systems remove much of this complexity by expressing synaptic values as real numbers represented by digital values of varying precision. Artificial synaptic weights are modulated by gradient-based updates to decrease an error signal or quantization to smaller representations.}
    \label{fig:synapses}
\end{figure}

\subsubsection{Biological}
The most abundant feature in the nervous system is synapses: estimates place the number in the human brain between 100 and 1000 trillion total synapses, or a median value of approximately 6,000 per neuron ~\parencite{santuy_estimation_2020, herculano-houzel_human_2009}. A synapse (Figure \ref{fig:synapses}) is a connection made between two specialized, electrically active cells---neurons---that allows the electronic activity of one neuron to influence another. When a neuron is electrically active (described in more detail in Section \ref{sec_neurons}), it sends a signal down a long ``axon'' to synaptic terminals. These terminals release chemicals, such as glutamate or gamma-aminobutyric acid (GABA). These chemicals, contained in synaptic vesicles, diffuse across the ``cleft'' between the two cells, and a portion will be received by a receiving partner on the other neuron: a dendritic ``spine,'' named for its appearance under magnification~\parencite{hering2001dentritic}. Each chemical exerts a unique influence on the postsynaptic neuron: some, such as glutamate, are excitatory, whereas  others, such as GABA, are inhibitory. Across the nervous system, an approximate prevalence of 20\% inhibitory and 80\% excitatory synapses is observed~\parencite{meunier_modulation_2017}. The amount of neurotransmitters emitted, transported, and received through the synapse subject to variability, as each step of the transmission process is probabilistic, rather than deterministic \cite{branco_probability_2009}. 

Synapses demonstrate intrinsic dynamics, such as short-term potentiation, long-term potentiation, and long-term depression, that influence the relative ``strength'' of each synaptic connection. These dynamics involve signals such as the concentration of calcium ions that can accumulate near synapses, the amount of neurotransmitters available, and the concentration of  modulatory chemical signals such as acetylcholine and norepinephrine~\parencite{kudithipudi_biological_2022}. The interplay of these dynamics is thought to be responsible for a large part of the brain's learning capability; by adjusting the size of dendritic spines, the release of neurotransmitters, and even physical connections  between neurons, the propagation and interplay of electronic activity within the nervous system can be changed over time~\parencite{meunier_modulation_2017, doya_metalearning_2002, smart_regulation_2000}.  Synapses can remain highly stable over time, allowing the brain to store long-term information~\parencite{holtmaat_transient_2005}. 

\subsubsection{Artificial}
The complex, temporal dynamics of the synapse are simplified into effective `weights' in artificial neural networks, modulating the influence that an input has over an artificial neuron's response. LLMs are commonly trained with 33--70 billion weights, and state-of-the-art models may contain over a trillion ~\parencite{epoch2023announcingupdatedpcddatabase}. These weights are generally initialized from random distributions and adjusted via gradients obtained by minimizing a numerical error value at a given task (Section \ref{sec_modules}).

Large-scale artificial neural networks were  implemented with weights encoded into single- or double-precision floating-point values (32 or 64 bits)---making them effectively real-valued and continuously variable. However, the cost of multiplying these values scales nonlinearly with the number of bits in the representation. This scaling strongly motivated developers of AI systems to reduce the precision of weights \cite{dally_hardware_2023}. Currently, weights can be quantized to representations with as little as 4 bits---a coding that is similar in precision to the estimates of variability in biological synapses~\parencite{wu_understanding_2023, bartol_nanoconnectomic_2015}. 

Although biological synapses are immobile on the timescale of spiking activity, weights in digital hardware are highly mobile---transported on digital buses and stored in static random access memory (SRAM) registers during calculation, and stored on dynamic random access memory (DRAM) or slower nonvolatile storage media such as flash~\parencite{rochefort_dendritic_2012}. The extra cost associated with transporting, storing, and waiting for weight data in digital hardware has motivated the development of numerous architectures that target this inefficiency, compiling representations of neural network programs into graphs that can be executed via systolic array architectures. Essentially, during execution, weights stay in local processor memories, and only input-dependent data, such as activations, are transported across the system~\parencite{chow_intermediate_2013, reuther_lincoln_2023}. But as previously mentioned, weight data may exceed the amount of local memory available to each processor, requiring weights to be swapped in and out of local memory~\parencite{hall_training_2023}. 

\subsubsection{Comparison}

The compact and stationary nature of a biological synapse remains a stark contrast with artificial equivalents. As previously mentioned, scaling digital systems to store information on the order of the number of synapses in the brain remains infeasible for current high-performance memory technologies such as SRAM and DRAM. Furthermore, this information - 4.7 bits per synapse - encodes only the shape of a dendritic spine, and disregards other unique properties of a synapse such as the probability of synaptic release and local chemical gradients, properties which can introduce useful behaviors such as stochastic sampling and local learning mechanisms. However, in contrast, the simple, real-valued behavior of the artificial synapse allows for theoretical tractability, and digitization of their values allows for perfect replicability and transport across arbitrary distances in time and space.

\subsection{Neurons}
\label{sec_neurons}

\begin{figure}[htbp]
    \centering
    \includegraphics[width=1\linewidth]{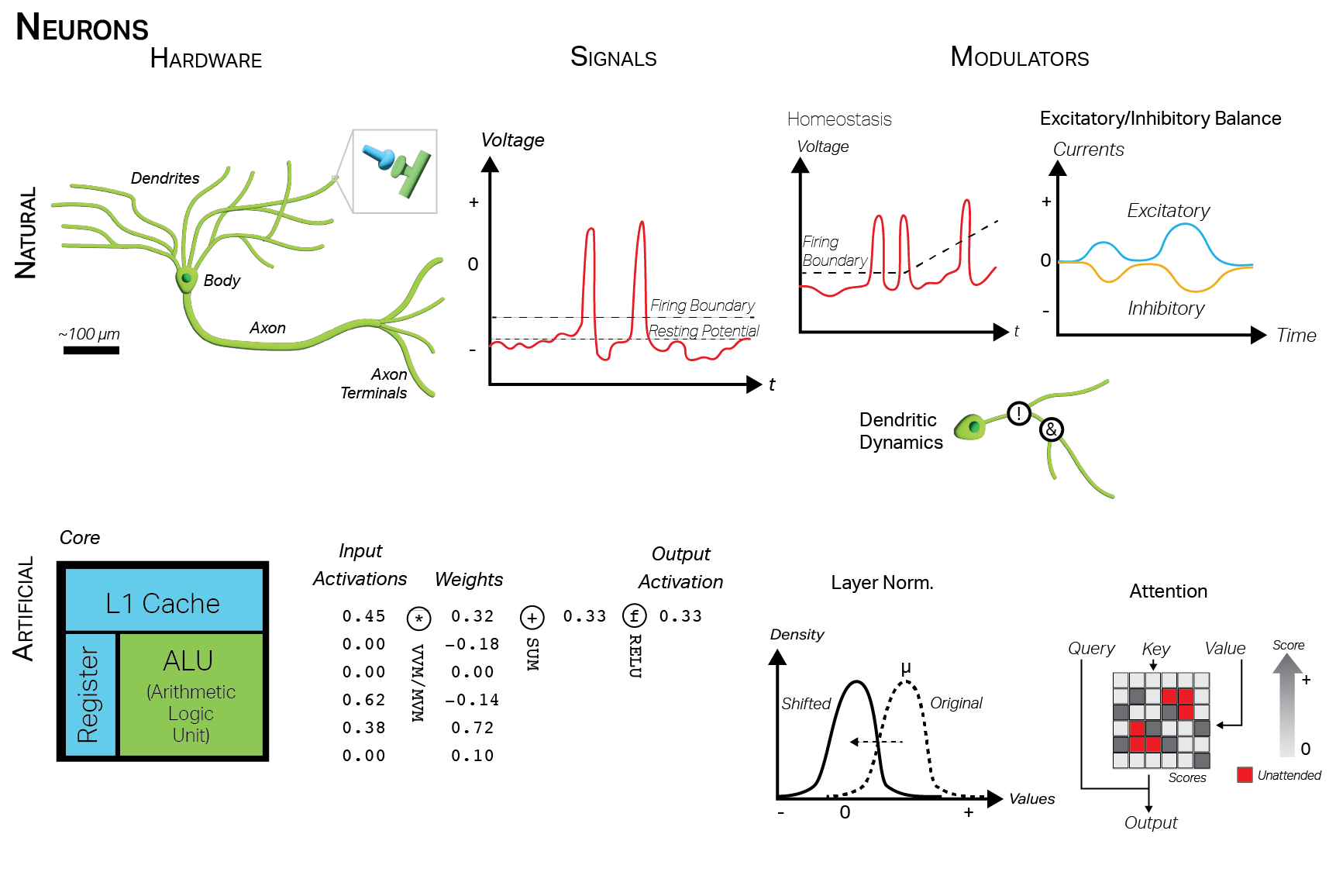}
    \caption{Biological neurons are complex, three-dimensional structures whose shape and behavior vary with an extensive number of subtypes. Fundamentally, however, each is responsible for maintaining a resting voltage potential that is influenced by inputs. As the potential of a neuron crosses a boundary in the phase space of its internal dynamics, it produces a voltage pulse that propagates from the neuron body down its axon and across synapses to other neurons. Neurons self-regulate their activity through mechanisms such as homeostasis, a precise balance between inhibitory and excitatory inputs, and complex gating behavior within dendritic structures. Most artificial neurons simplify this behavior by assuming neurons reach a steady ``firing rate,'' and assign this a real ``activation'' value. These can be calculated quickly in digital logic by multiplying each real input by a synaptic weight, summing it, and applying a nonlinear ``activation function'' that originates from the nonuniform response of biological neurons to constant inputs. Similar to biology, the activations of neurons are modified through techniques such as layer normalizations and attention mechanisms to both tame numerical complexities of these systems and enable them to address shifting contexts.}
    \label{fig:neurons}
\end{figure}

\subsubsection{Biological}
Synapses are hosted on neurons (Figure \ref{fig:neurons}). Neurons are specialized biological cells that actively maintain a gradient of ions, primarily sodium and potassium, across the interior and exterior of the cellular membrane. This ``resting potential'' is approximately 60--70 mV~\parencite{Chrysafides2024}. Neurotransmitters received by synapses on dendrites (or other sources) influence this potential, respectively moving it toward or away from a voltage ``threshold'': a variable boundary past which the neuron's body will produce a rapid, self-propagating membrane voltage inversion called an action potential or ``spike'' that propagates down the axon~\parencite{gerstner2002spiking}. The physiology, resting potential, excitability, and other specific parameters of a neuron can vary greatly, with genetic techniques identifying thousands of subtypes within the human brain~\parencite{siletti_transcriptomic_2023}. 

In contrast to the information stored in synapses, the ``spiking'' electrical signals produced by all neurons are mobile and transient,  lasting only milliseconds~\parencite{lemon_classification_2021}. These complementary characteristics support the widely held view that synapses in the brain encode memories and spikes encode short-term, transient information reflecting the environment and current state of mind. Although the amplitude of ``graded'' spikes can encode information, most are held to be binary, all-or-nothing signals~\parencite{juusola_coding_2007}. These binary spikes can encode discrete and real-valued information through various coding strategies.

Foundational studies of large, isolated neurons  focused on coding real values via the frequency with which spikes were emitted from a neuron. Thus, these models were founded on the interpretation that the more frequently a neuron ``fired,'' the stronger its ``activation''~\parencite{hodgkin1952quantitative}. However, more modern studies of neurons have revealed that a variety of coding methods are used, including the relative timing of a spike's arrival, its phase relative to a wave, and other strategies~\parencite{auge_survey_2021}. These alternative coding techniques allow for fast, efficient transmission of complex information that can be implemented via relatively simple neuron models~\parencite{izhikevich_resonate-and-fire_2001}. Even irregular or noisy neuronal firing can be utilized to allow neurons to communicate uncertainty or amplify small signals~\parencite{buesing_neural_2011, plesser_noise_2000}. 

As producing an electrical ``spike'' consumes significant metabolic energy, neurons exhibit several behaviors that allow them to self-regulate activity. The firing rates of neurons across the human brain are log-normally distributed, with a mean firing rate of 0.7 spikes per second~\parencite{buzsaki_log-dynamic_2014}. In other words, despite receiving approximately 4,200 inputs per second from 6,000 synapses, on average a neuron will fire only  once every 1.4 seconds. This sparse activation process is not a matter of chance but emerges from multiple interacting systems that support it and maintain the brain's metabolic activity. 

One crucial mechanism enabling this sparsity is the fine balance between excitatory and inhibitory inputs to a neuron, which tightly track one another through feedback to implement efficient coding and make firing a rare event~\parencite{iascone_whole-neuron_2020, Pastore2018, deneve_efficient_2016}. Damage to this balance is dangerous and can lead to the generation of damaging events such as seizures~\parencite{timofeev_neocortical_2004}. 

Furthermore, synaptic inputs to neurons are arranged and compartmentalized in complex structures called ``dendrites.'' Rather than simply transporting information toward the main neuron body, dendrites actively engage in information processing, nonlinearly combining the contributions of individual spikes~\parencite{london_dendritic_2005}. This gives greater complexity to individual neurons than was captured in the direct synaptic inputs and simple weights of the perceptron model, and it supports the ability to discriminate between a wide variety of synaptic input patterns to support a neuron's sparse activity. 

Even if a neuron's inputs change (e.g., through growth or damage), the homeostatic behavior of individual neurons---metabolic constraints, activity of ion channels, and other factors---collectively can allow the activity of a neuron to restabilize at a ``target level'' even given changing inputs or other external factors~\parencite{harnack_stability_2015, oleary_neuronal_2011}. 

\subsubsection{Artificial}

The intricate, time-varying behavior of even early neuronal models, such as the Hodgkin-Huxley model, encouraged pioneers of connectionist architectures to abstract and simplify the behavior of an individual neuron~\parencite{hodgkin1952quantitative}. At the time, the ``firing rate'' of a neuron---how often it spikes given a set of inputs---was thought to be the primary method by which signals were coded in the brain~\parencite{ rosenblatt_perceptron_1958}. This inspired the ``activation function'' used in neurons, where the currents received by a neuron are summed and transformed via a nonlinear function to a neural activation, which  originally represented a biological neuron's firing rate.

Although this view of ``quasi-static'' firing rates diverges from more modern views into biological computation, in which noise, self-sustained activity, rhythms, and feedback play essential roles, firing rates have nonetheless proved an extremely effective basis for perceptrons and their modern descendants~\parencite{buzsaki_brain_2019, lecun_deep_2015}. However, as perceptrons and activation functions departed from their biological inspiration, artificial neurons became capable of transmitting negative as well as positive activations, since they can assist in propagating gradients through error functions~\parencite{lecun_deep_2015, kunc_three_2024}. 

Modern artificial neurons are most commonly implemented in digital electronic logic, where local caches of SRAM allow for the storage of synaptic weights, which can be digitally multiplied by incoming input values, and transformed into activations. These manipulations are carried out via instructions and data stored in fast SRAM memories and executed in  algorithmic logic units (ALUs). One ``core'' with storage caches, registers, and ALU(s) may model many neurons during a single step through parallelization.

In ANNs, sparse activations are explicitly preserved throughout computation; ANNs generally produce activations that are highly sparse~\parencite{gale_state_2019}. This sparsity is achieved by utilizing a variety of activation functions, dropout layers, regularization techniques, and---similar to homeostatic biological mechanisms---regularly placed normalization layers that shift activations by recent or historical means and deviations~\parencite{kunc_three_2024, srivastava2014dropout, sakai_dropout_2019, huang_normalization_2023}.

In parallel, the complexity of neuronal behaviors has grown to support applications such as LLMs. The crucial advance that enabled these systems was ``attention.'' Attention mechanisms enables a set of inputs to a neuron to cross-modulate their strengths through ``scores,'' allowing the neuron to ``attend'' to certain sets of inputs. Depending on how this mechanism is implemented and deployed, this may enforce sparse, local comparisons in space, or enable vast, all-to-all comparisons through space, time, or other domains~\parencite{guo_attention_2022, sun_efficient_2025}.

Hardware has rapidly evolved to support the distributions of numerical values present in artificial networks. Representations of values specialized to those found in neural networks have been established;  for instance, brainfloat 16 and 32 define a new standard that reduces the number of bits in a floating point used for the exponent and increases the number used for the mantissa~\parencite{wang2019bfloat16}. More significant changes include the quantization of real values into integers from 32 to 4 bits, as well as the use of log representation, the latter of which dramatically changes circuits by shifting multiplications to simpler additions~\parencite{wu_understanding_2023, dally_hardware_2023}. 

Complementing this, the large proportion of zero values in neural networks has led to GPU and other hardware that natively supports data structures with high sparsity, calculating only over nonzero values and accelerating computation~\parencite{dally_hardware_2023}. These adaptations allow for the cost of representing and manipulating the values required for ANN-focused systems to be significantly reduced. 

\subsubsection{Comparison}

By the standards of modern electronics, biological neurons are large and slow: existing on the order of dozens of microns, they are electrically excitable at a maximum rate of kilohertz. In most digital architectures, a neuron does not exist physically: it is a set of software instructions executed on an arbitrary processor operating at GHz. The ``slow'' speed of a biological neuron suits it to expend the minimum possible energy to improve survival by allowing the body to react in a timely manner, in contrast to the high speed and energy dissipation of the digital processor which is an expensive technological investment which must justify itself through high throughput. 

While artificial and biological neurons vary tremendously in speed, normalization and sparsity have emerged as important features in both. Where biological neurons require sparse firing to preserve energy and avoid damage, we observe in artificial networks that these features are also important to promote effective training.

Finally, rather than being stereotyped, biological neurons take a massive variety of shapes and sizes across the brain. In contrast, artificial neurons may vary by their connectivity and layout, but often utilize identical activation functions across entire architectures. Encoding greater variety into neurons themselves can offer the opportunity for more compact functionality at the cost of introduction another variable into an already-large design space. 

\subsection{Modules}\label{sec_modules}

\begin{figure}[htbp]
    \centering
    \includegraphics[width=1\linewidth]{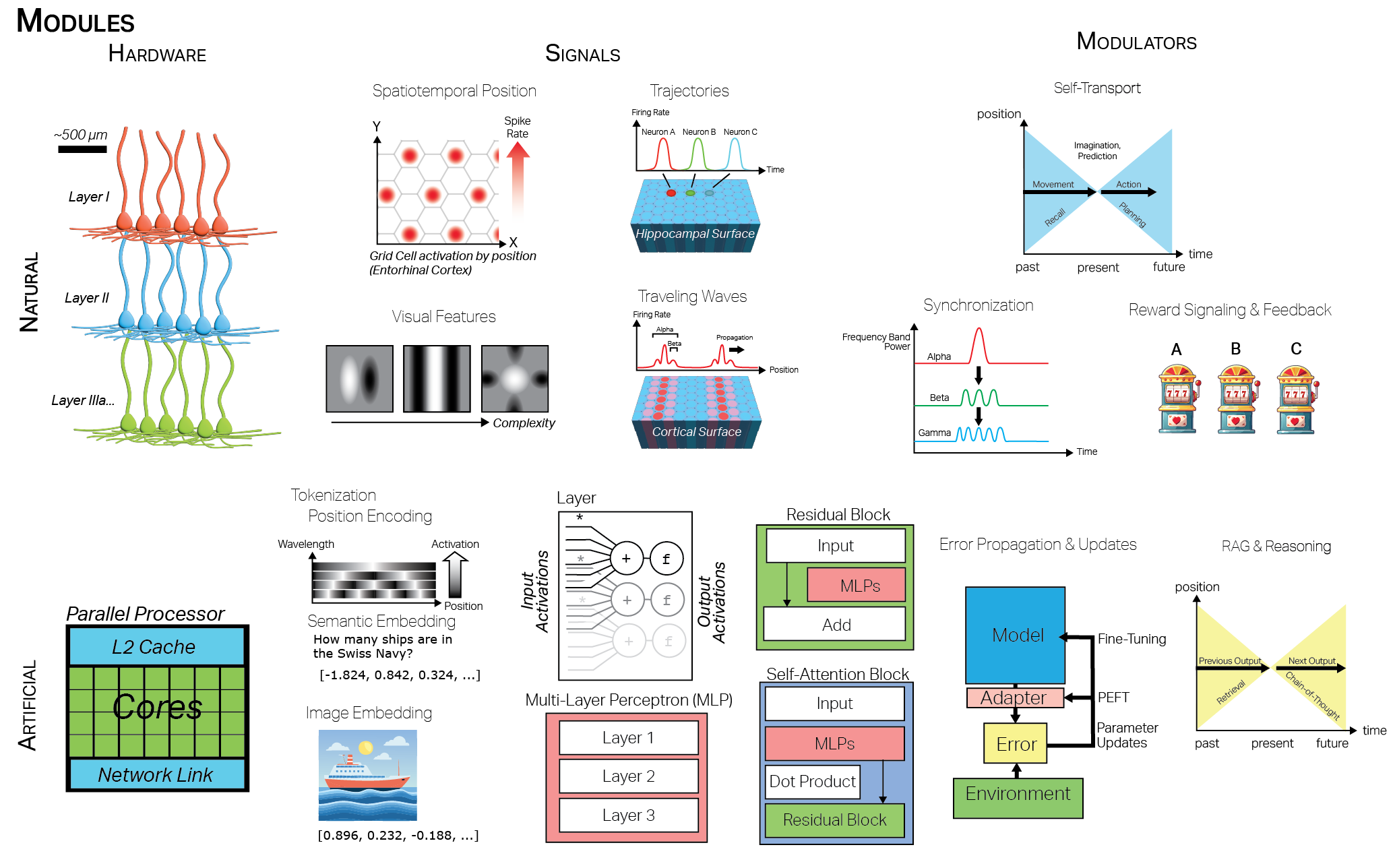}
    \caption{Structured connectivity between individual neurons allows them to collectively implement a variety of functions. Many structures within the brain have been identified, and ``layers'' of neurons (and other cells) that connect and interact in a variety of ways constitute these modules. Biological modules process and represent information such as position in the entorhinal cortex, perceived features in the visual cortex, history and paths in the hippocampus, and advanced concepts in the neocortex. Varying cell types, connectivity patterns, and other factors modulate these regions, but across the brain universal methods for controlling activity include the propagation of synchronized waves and use of chemical reward signals. Additionally, areas cross-modulate by allowing for imagined and real movements. In artificial systems, many neurons may be implemented across multiple cores with associated higher-level caches and network links. Similarly to the brain, alternate modules are used for encoding position, visual information, and high-level semantic language into an interoperable set of ``tokens'' and/or activations. Layers of artificial neurons are stacked to produce multilayer perceptrons, residual blocks, attention mechanisms, and more. These artificial modules can be modulated through the use of low-rank ``adapters'' to adjust or update behavior, as well as being directly retrained with an error signal. Artificial systems are also gaining the ability to self-regulate their actions through multiple abilities such as chain of thought and tool usage.}
    \label{fig:modules}
\end{figure}
\subsubsection{Biological}

Although individual neurons are complex, intricate systems, many are needed working in concert to robustly represent, manipulate, and transport information. And although a full map of the connectivity between neurons---the ``connectome''---has not been fully completed in mammals, advances in neural imaging have yielded fascinating insights into the organization of neurons into modular sections responsible for independent functionality ~\parencite{sporns_graph_2002, roe_columnar_2019, buzsaki_brain_2023}. Information transmitted across each module can be used in conjunction with other signals, such as neurotransmitters, to modify local neuronal activity and connectivity ~\parencite{meunier_modulation_2017}. Reward signals from the environment are known to be connected to the presence of neurotransmitters, and these chemicals can induce a variety of changes in neuronal and synaptic behavior as described previously (Sections \ref{sec_synapses}, \ref{sec_neurons})~\parencite{kudithipudi_biological_2022}.

The visual cortex is a highly studied module of the brain responsible for processing optical information received from the environment. The arrangement of the retina itself and associated regions of the brain have been elucidated and directly inspired artificial convolutional neural networks, as well as event-based cameras~\parencite{macpherson_natural_2021, benosman_event-based_2014}. Features of the brain, such as progressively detecting and representing more advanced visual features, have been replicated in artificial networks~\parencite{lecun_deep_2015}. 

Whereas the visual cortex processes optical information, the entorhinal cortex (EC) was discovered to process and represent location. In the EC, neurons become active at regular intervals in space; by combining the net activity of multiple neurons active over fields of differing sizes, a location can be decoded~\parencite{hafting_microstructure_2005, fiete_what_2008}. Furthermore, this representation corresponds to a residue number system under which a number is described by its modulo values from relatively prime factors. This allows for a highly dynamic range of values to be encoded and manipulated efficiently, with the constraint that these representations are ultimately periodic~\parencite{garner1959residue}. Multiple residue numbers can be combined to cover an arbitrarily dimensional space, enabling the representation of position in time, space, and more abstract spaces~\parencite{moser_place_2015, schoyen_hexagons_2024, klukas_efficient_2020}. 

Information from multiple regions of the brain can converge in areas such as the hippocampus (HC), where information perceived from the environment and actions can produce unique neuronal ``trajectories,'' which trace out a representation of this experience and even allow it to be played back~\parencite{denovellis_hippocampal_2020}. Notably, if the HC is removed, learning new information becomes difficult or impossible~\parencite{squire_legacy_2009}. 

The HC has rich interactions with the neocortex (NC), one of the largest regions of the human brain, formed by vertical columns of richly connected neurons that are tiled to make a large sheet with six layers. The interface between these layers provides rich synaptic connectivity  between different regions of the NC~\parencite{mountcastle_columnar_1997, roe_columnar_2019}. This large sheet is itself folded, maximizing the available area within the confines of the skull and giving the human brain its characteristic ``lumpy'' appearance. Its evolutionary expansion in mammals is believed to have enabled advanced reasoning and thought ~\parencite{hofman_evolution_2014,lui_development_2011}. 

One long-standing question in neuroscience is how information from separate regions of the brain propagates and form a common syntax that different regions can decode and modulate. Recent research has shown that waves of spiking activity that propagate across various regions of the brain could fill this role~\parencite{ benigno_what_2022, aggarwal_visual_2022}. The success of perceiving new information itself has been shown to be related to the phase of an arriving wave, gating the transport of information~\parencite{keller_traveling_2023}. This complements growing knowledge of other neural components interacting with waves to represent information on place and learned information~\parencite{buzsaki_neuronal_2004, molle_influence_2009}. 

The brain is not dominated by a single wave or frequency of waves but contains many. One of the earliest neurological discoveries was of the alpha power band: a set of frequencies emerging from the distribution of normal activity with significant power and stability~\parencite{buzsaki_neuronal_2004}.  This discovery was followed by subsequent bands that display rich informational structure and the ability for slower rhythms to modulate faster ones. In other words, as daylight influences our daily mealtimes, waves in the brain also synchronize, with slower waves aligning and modifying faster ones~\parencite{klimesch_frequency_2018}. Additionally, despite large variances between brain structure and volume in animals, the presence and hierarchy of these frequency bands remain fairly consistent~\parencite{buzsaki_brain_2023}. Traveling waves have thus been suggested as an implementation of the brain's overall ``working memory''---the ability to know one's location, recent actions, the state of the environment, and more~\parencite{keller_traveling_2023,keller_spacetime_2024}. 

Furthermore, neural rhythms play an important role during sleep. Rather than being an ``unproductive'' period, rest is a time when the brain is highly active. Its normal hierarchy of rhythms shifts; and as the brain enters slow-wave sleep (SWS), it simultaneously engages the HC and NC, replaying recently obtained information from the former and integrating it into the latter through mechanisms that avoid overwriting extant knowledge to avoid ``catastrophic forgetting'' of older information~\parencite{mcclelland_why_1995, kumaran_what_2016, yang_selection_2024, walker_role_2009}.  SWS is not observed solely in mammals but has equivalents in other species such as birds and reptiles~\parencite{miyazaki_sleep_2017}. 

\subsubsection{Artificial}

The multilayer perceptron (MLP) is the foundation of modern connectionist architectures~\parencite{rosenblatt_perceptron_1958, block_perceptron_1962, lecun_deep_2015}. The MLP architecture consists of multiple ``layers'' of artificial neurons. The output of each neuron in one ``dense'' layer is connected to each neuron in the succeeding layer. By adding successive layers with nonlinear activation functions, the MLP can serve as a universal function approximator~\parencite{augustine_survey_2024}. However, the quadratic scaling of the number of connections between each layer, as well as other numerical issues, requires modifications to scale to the very large systems employed today. These include previously mentioned modifications to activation functions, sparsification, normalization methods, and selective convolutional connections.

In particular, residual connections and attention mechanisms were crucial advances to enable today's LLMs. Residual connections were originally inspired by the ``skip'' connections observed in the columnar connectivity of the neocortex~\parencite{he_deep_2016}. These long-range connections pass information received at their inputs to deeper layers and allow for the successful training of very deep dense and convolutional networks. And as previously mentioned, attention mechanisms allow the inputs to a neuron to modulate one another, developing patterns of co-activation or inhibition across time and space ~\parencite{guo_attention_2022, vaswani_attention_2017}. Whether these modulations can be constructed between all inputs, in a temporally causal or acausal order, spatially local or global, dense or sparse, has enormous impacts on the capability of the resulting model, as well as the amount of computation it requires and amounts of data it must store over time \cite{yu_reverse_2025, sun_efficient_2025}. 

In addition to these advancements, a variety of other modules were necessary to implement the modern LLM models used today. For instance, ``tokenization'' of written language (and even images) into the activation values processed by LLMs was necessary~\parencite{liDiscreteTokenizationMultimodal2025}. Furthermore, position of each token within a sentence had to be effectively coded within a latent space. One of the most successful methods to accomplish this is rotational positional encoding (RoPE)~\parencite{su_roformer_2022}. Like grid cells, RoPE utilizes a variety of spatial frequencies that periodically modulate across different locations and can be combined to map higher-dimensional spaces. 

Self-attention blocks, positional encoding, tokenization, and other advances enabled the ``transformer'' architectures, which allow LLMs to efficiently carry out autoregression over text: Given a series of words, the model predicts the next. This prediction can be appended to the previous input, and the next word predicted. Thus, the model appears to ``remember'' what it has ``said.'' The model can then interact with a user's responses, attending to the entire ``context'' of the conversation, until its maximum capacity is reached, often limited by the span of its attentional mechanism or poor extrapolation from position encodings~\parencite{han_lm-infinite_2024}. Modified attention mechanisms, training datasets, positional encodings, and other methods are currently being applied to address this issue~\parencite{zheng_rethinking_2021, han_lm-infinite_2024}. 

Although many AI systems remain ``dense''---that is, each one of their weights must be loaded and used during each inferential or training calculation pass---several leading models have amortized this cost by employing a ``mixture of experts'' (MoE). In an MoE system, a gating network controls the routing of messages along subnetworks, reducing the number of parameters actively engaged in a single computational pass~\parencite{cai_survey_2024}. 

Various methods exist to modulate the behavior of transformers and other related ANN modules. Many of these methods are based on the ability to compute continuous gradients through these modules by the use of automatic differentiation (AD) systems. AD establishes a numerical relationship between each parameter in the model and its output, allowing numerical optimizers to reduce this error over time---``training'' the system to a task~\parencite{baydin2018automatic}. In the case of ``foundation models,'' the task is to predict the next word in a sentence over the entire corpus of collected written language, resulting in the large training costs for these systems~\parencite{bommasani_opportunities_2022}. Further modification may ``fine-tune'' the behavior of a foundation model---for instance, making it behave in a conversational manner suitable to the familiar chatbot applications deployed today~\parencite{bai_training_2022}. Further adaptations, specializing a system to a small corpus of data (such as specialized chip design language), may be accomplished by introducing a small set of new weights that modify only the model's existing behavior without changing original weights~\parencite{han_parameter-efficient_2024}. Famously, artificial neural networks are subject to catastrophic forgetting: if new training data does not incorporate previous datasets, performance on excluded examples will decrease to the point of failure \parencite{robert_m_french_catastrophic_1999}. Several methods to overcome this issue and enable ``continual learning'' have been proposed: measuring the importance of weights and making them resistant to change, incorporating new parts of layers or experts during training, or even sleep-inspired ``generative replay'' to interleave synthetic old data and new examples during training \parencite{ven_continual_2024}. 

The texts used as the input to an LLM are also increasingly being employed to modify these systems' behavior. Often, an invisible ``system prompt'' is injected into most chat systems to align their output with the goals of their developer~\parencite{bai_training_2022, wu_jailbreaking_2024}. Chain-of-thought models also modify the use of context to encourage systems to separate ``thinking'' from ``speaking,'' allowing them to ``reason out'' problems before responding to user inputs~\parencite{wei_chain--thought_2022}. Agentic systems offer an expanding set of programming and database tools for models to employ as alternative actions rather than simply ``thinking'' and ``speaking.'' For instance, many AI tools can employ vector databases to reference specific pieces of information and include them in the current context~\parencite{li_survey_2022}. Similarly to the ability of biological systems to take different actions based on context, the artificial context of LLMs is also being employed to enable them to imagine different outcomes, recall information, and set a course of action based on this information.

\subsubsection{Comparison}

The brain contains a number of distinct regions, some of which are highly specialized to certain tasks such as visual processing or spatiotemporal navigation. Encoding this information into and transporting it via electrochemical traveling waves may be the common ``syntax'' which allows these regions to inter-operate, allowing for these physically distant areas with distinct tasks to engage with one another and allow for high-level planning, learning, and action based on integrated information. In artificial networks, information is shared through successive ``layers'' of modules acting on a set of inputs. Information from the recent past may or may not be included, and recurrent storage of the recent past is not common in many architectures, though it may be added into the inputs. The common syntax of these artificial modules is usually digital representations of real numbers in a high-dimensional latent space. 

While artificial neurons are often homogenous, an enormous amount of innovation has taken place in the space of layers and modules for artificial computation: dense, sparse, convolutional, normalization, and attention layers are all ``basic'' essentials which are commonly implemented, with hundreds or more variations available. Furthermore, the ability to design these layers in software and test them on general-purpose hardware enables a rapid pace of evolution impossible in evolved, living structures such as the brain.

However, while it cannot be radically re-implemented with ease, the brain contains enormous amounts of flexibility. Homeostasis, neurogenesis, synaptic plasticity, and other mechanisms allow new information to be acquired throughout life without destroying previous knowledge. This ability to learn involves multiple structures and mechanisms, and actively addresses relevant regions for updates. Interaction with databases, historical inputs, and other mechanisms can emulate learning in artificial systems but do not induce long-term, fundamental changes in the underlying model behavior. 

\subsection{Systems}
\label{sec_systems}

\begin{figure}[htbp]
    \centering
    \includegraphics[width=1\linewidth]{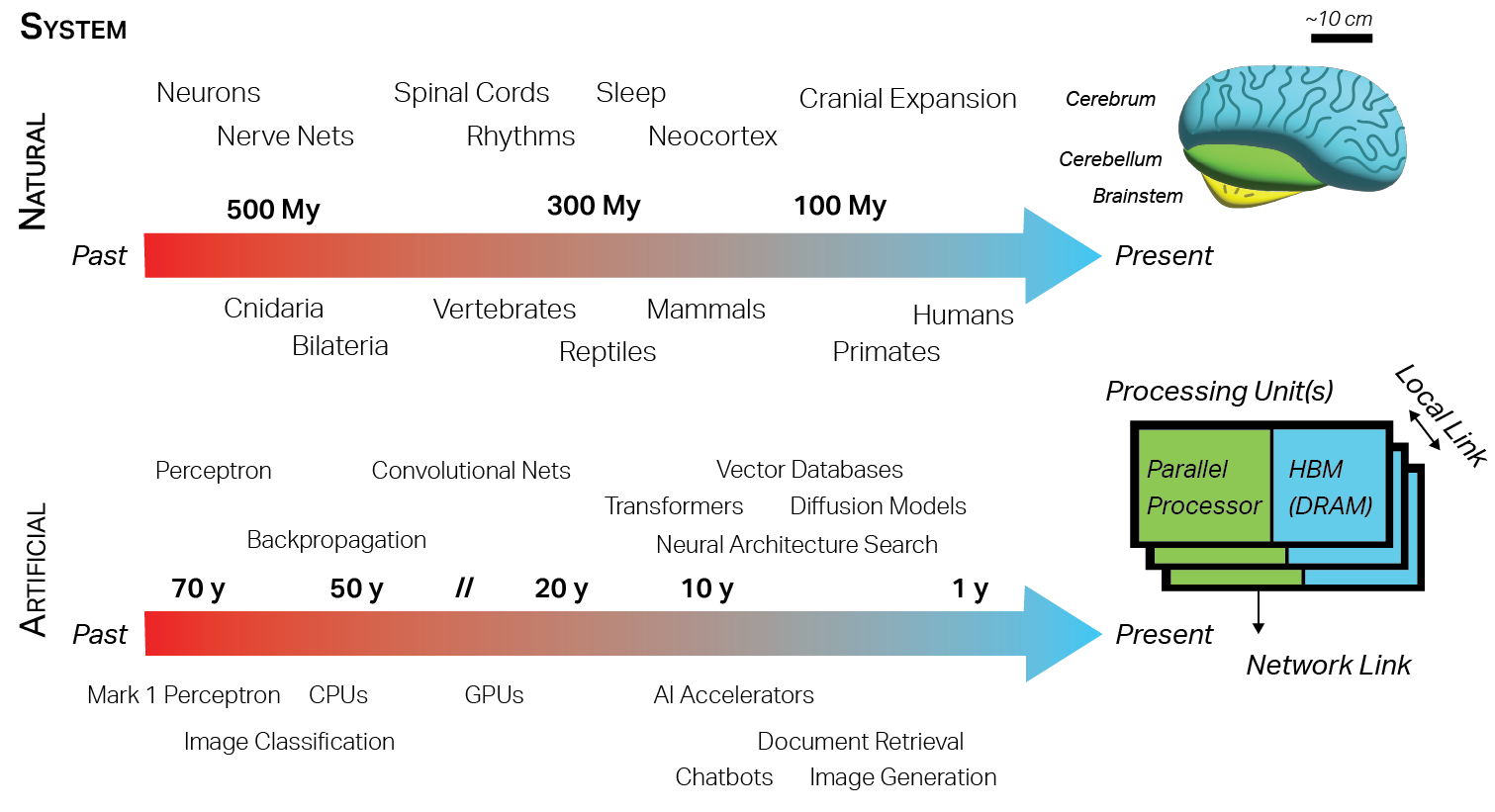}
    \caption{Both NI and AI have adapted over time to expand their capabilities. Major architectural changes have occurred gradually over hundreds of millions of years (My) in NI. Neurons preceded even bilateral organization of the body and have evolved into the highly integrated, dense, and capable structures found in the human brain. AI has undergone rapid changes since its foundational inception, as the adoption of GPUs and changing computational supplies and demands drove the growth of AI systems to the revolutionary tools found today.}
    \label{fig:systems}
\end{figure}

Cybernetics defines intelligent systems as those that can process information, learn from feedback, adapt, and regulate themselves over time~\parencite{wiener1948cybernetics}. Although NI applies these capabilities to survive in an uncertain natural environment, a major goal of AI is to replicate these abilities to navigate and manage an increasingly large world of information. As previously introduced, meeting the demands of these niches requires system-level characteristics that differ, and integrating AI with the natural environment introduces new spatiotemporal constraints in power, latency, cost, and other measures (Section \ref{sec_intro_summary}). In contrast to previous subsections \ref{sec_synapses} - \ref{sec_modules} where hardware, signals, and modulatory mechanisms are described, here we describe how the systems integrated from these components meet the demands of an environment, interact with it, and change through time. 

\subsubsection{Biological}

Evolutionary pressures have shaped NI into the wide variety of forms observed today (Figure \ref{fig:systems}). Neurons were present even before the emergence of bilaterally symmetric animals over 500 million years  ago ~\parencite{bosch_back_2017, heger_genetic_2020}. As a result, neurons can be observed in animals as small as insects and as large as blue whales ~\parencite{dorkenwald_neuronal_2024, buzsaki_scaling_2013}. Ancient species, such as cnidaria (jellyfish, sea anemones, and others), demonstrate complex signaling techniques such as synaptic signaling through multiple neurotransmitters and  share neuronal proteins with the vertebrate nervous system ~\parencite{bosch_back_2017}. These capabilities, while relatively simple, allow for the transmission and processing of information across a body to increase capabilities for sensing, movement, predation, and more. For instance, traveling waves are observed in jellyfish to coordinate swimming behavior~\parencite{pallasdies_single_2019}. Oscillations have also been consistently maintained through evolution across a wide variety of mammalian species;  the region of the brain responsible for generating traveling waves during sleep was present 320 million years ago in the common ancestor of both reptiles and mammals~\parencite{buzsaki_scaling_2013, norimoto_claustrum_2020}. The widespread presence of rhythms in neural systems suggests that their ability to regulate local behavior, synchronize neurons, and generate rhythms has been a useful primitive in NI systems for hundreds of millions of years.

Primate brains contain more neurons per unit volume than other mammals~\parencite{herculano-houzel_human_2009}. The expansion of primate brains occurred over several million years, with humans and Neanderthals reaching the highest ratio of brain volume to body mass~\parencite{puschel_hominin_2024, miller_quantitative_2019}. Despite its high efficiency (Section \ref{sec_bio_efficiency}), however, the brain is hugely costly: although  approximately 2\% of the human body's mass, the brain uses 20\% of available metabolic energy. The ability to support these high costs was likely supported through a combination of tool use, food processing, and language, leading to positive feedback in which a more capable brain becomes better suited to support itself~\parencite{herculano-houzel_scaling_2011, miller_quantitative_2019}. Many high-level behaviors such as tool use and communication are not uniquely human but have evolved in multiple species, such as corvids, with large numbers of neurons to support centralized processing~\parencite{herculano-houzel_numbers_2017}.

Although the brain is the hub of information processing, it both influences and is acted on by the body and the environment around it. Rather than being viewed as an isolated machine that produces outputs from inputs, the brain may instead be viewed as a participant in feedback loops that extend outward in space and time. For example, movement affects internal representations of position, which may in turn promote the recall of linked trajectories or other stored information in separate parts of the brain. This information may, in turn, cause us to act and move again~\parencite{buzsaki_brain_2019}. The manipulation of combined positions---both spatial and temporal, physical and virtual---has been suggested as a common, underlying ``reference frame'' of the brain~\parencite{hawkins2021thousand}. 

Furthermore, the brain can employ feedback loops to the external environment to supplement its own capabilities. Any literate person can instantly and permanently learn a new piece of information by writing it down---as long as the corresponding piece of paper isn't lost. Other tools such as slide rules, calculators, computers, and even AI allow us to expand our own capabilities through the use of interactive feedback~\parencite{slors_notebooks_2020}. 

\subsubsection{Artificial}

The evolution of AI is much shorter than that of NI but better documented. Cybernetics was founded in the era following World War II; it \ guided researchers in diverse areas from neuroscience to control systems and laid the foundations for what would be defined as ``artificial intelligence'' in the 1955 Dartmouth workshop~\parencite{wiener1948cybernetics, mccarthy_proposal_2006}. In the following years, fundamental aspects of modern connectionist architectures were established, such as multilayer neural networks (1958) and backpropagation (1976)~\parencite{rosenblatt_perceptron_1958, linnainmaa_taylor_1976}. However, only with the advent of large parallel computers and datasets did training these networks become practical. The rise of computing and the end of Dennard scaling in the early 2000s created unique economic conditions for new combinations of hardware and software to emerge, and a winning ticket for this ``hardware lottery'' was the newly found combination of parallel GPUs and neural networks; the current generation of AI research was spurred by this discovery~\parencite{thompson_economic_2017, hooker_hardware_2021, krizhevsky_imagenet_2012}. 

These early research successes in computer vision and other applications broadened interest in neural networks as a means toward general AI applications, and  numerous architectures, modules, meta-learning methods, and software tools were developed (Figure \ref{fig:systems}) ~\parencite{ren_comprehensive_2022, paszke_pytorch_2019}. Subsequent advances such as transformers, diffusion networks, and vector databases enabled commercially relevant AI skills such as text-to-image generation, video synthesis, text summarization, and document retrieval (Section \ref{sec_modules}) ~\parencite{rombach_high-resolution_2021, li_survey_2022, brooks_video_2024, dubey_llama_2024}. These led to the situation observed today, where AI applications enabled by neural networks are highly valued and commercially relevant tools~\parencite{WSJ_2025_AI_revolution, fitch_nvidia_2024, smith_ai_2024}. However, these successes remain highly dependent on NI to label datasets, mark undesirable behaviors, and train AI models into agents through reinforcement learning with human feedback (RLHF) \parencite{bai_training_2022}.

The decade-long journey from the initial successes of computer vision to wide-scale adoption of chat agents took place using primarily large-scale computing resources, spurring major investments in data centers and even power generation~\parencite{reuters_microsoft_openai_2024, patel2024multidatacenter, hiller_microsoft_2023}. Rather than being constrained by the relatively small amounts of information and power easily available at one point in space and time, AI's development was massively accelerated by enormous amounts of energy and information available in datacenters. Rather than evolving under the constraints of a natural environment, the scale-up of computational resources, power, and information has potentially enabled AI to scale farther and faster than many experts predicted~\parencite{grace_thousands_2024}. However, even given this massive scale-up of AI systems over previous years, encounter other limits, such as amount of available data, power, infrastructure, or undiscovered algorithmic hurdles are being encountered~\parencite{epoch2024trainingcomputeoffrontieraimodelsgrowsby45xperyear, villalobos_will_2024, balaraman2025transformer}. Future progress in addressing the shortcomings of current AI models or addressing niches in which they remain inapplicable may not be solved through a continued scale-up. 

Rather than directly comparing natural and artificial systems here, we extend this discussion of system-level convergences, divergences, and opportunities in the next section (Section \ref{sec_vectors}).

\subsection{Summary}

Let us summarize our brief traversal of the nervous system and paint a high-level portrait of its operation: electrical or chemical signals are modulated by synapses on dendrites. Dendrites themselves are complex, active processing structures that modify information accumulated at their host neuron's main body. Neurons receive thousands of inputs per second, but fire rarely because of inhibitory feedback. Networks of neurons produce traveling waves, oscillations that move through the neural substrate, synchronizing a nested hierarchy of activities and maintaining working memory. Working memory contains multitudes of information on place and time, both internal and external, and this information can be manipulated through real or imagined movement. Waking activities and sleep enable continual learning by replaying old and new experiences together. Natural intelligence applies these capabilities to improve survival by enabling the acquisition of new information to adjusting behavior while moving through and interacting with a rich natural environment. 

Artificial intelligence has arguably succeeded at applying many of these principles to a different context. AI systems use massively parallel, nonlinear processing similar to neurons and synapses, where activations are assumed to correlate with a ``steady-state'' activity of a biological neuron through time. AI systems actively encourage and utilize sparsity and modulation in a manner similar to NI and utilize encodings of space and time that are highly similar to grid and place cells. Short-term memory in AI emerges in the form of a self-modifiable context, but long-term memories remain expensive to modify. AI systems synchronize via digital clocks that guide all aspects of computation, including movement of data, instructions, and execution of calculations. Many leading AI models employ a ``mixture of experts,'' allowing certain subsections of the network to specialize in certain contexts and reducing the number of parameters required during inference. AI applies these capabilities to excel at parallel processing of multiple user inputs in high-cost, high-throughput data systems that are constructed at high cost and remain inflexible to many end users and application niches. 
\section{Vectors for Developing AI Hardware and Software }
\label{sec_vectors}

To begin, we showed that while under purely per-operation efficiencies, AI and NI have begun to compete with each other, this is only under different environmental conditions (\ref{sec_intro_summary}). To achieve AI systems that can be effectively embedded in a variety of environments, including real-time and individual-user applications, this gap in capabilities must be addressed. We propose  advancements along two general axes: developing hardware that can efficiently execute large models while trading off execution speed and power and greatly decreasing the amount of energy required for AI to integrate new information. 

We presented a hierarchical review of the computations in NI and compared how this knowledge relates to current AI systems (\ref{sec_comparing}). Here, we integrate these perspectives to suggest future biologically inspired vectors for research in AI hardware and software to enable embedded, agentic, and physical AI systems \parencite{luo_integration_2024}. Complementing this, we conclude by highlighting areas that may not be directly addressable by biological inspiration. 

\subsection{Vectors for Software}

\begin{figure}[htbp]
    \centering
    \includegraphics[width=1\linewidth]{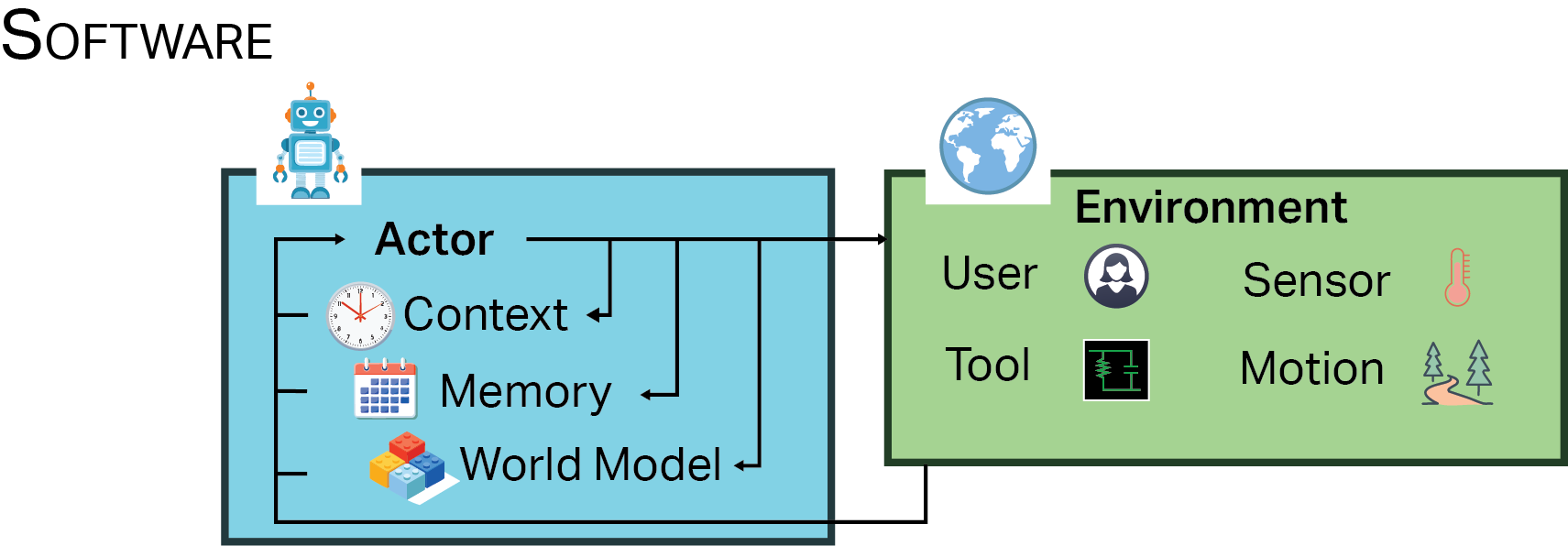}
    \caption{AI systems are becoming embedded in multiple feedback loops, rather than being simple input-output systems. By manipulating their context, storing information in databases, or querying other foundation models, their internal dynamics are enriched. Additionally, these models can grow to access external resources such as sensors, software tools, users, and potentially the external world through motion.}
    \label{fig:systems_software}
\end{figure}

Although convergent along several directions, AI and NI remain divorced in their ability for systems to self-direct their actions. Many AI systems utilize a single action: they \textit{speak} until they predict the next $<end>$ token and then wait for another input. More advanced systems are augmented with other verbs, such as \textit{think} or \textit{retrieve}, allowing them to use chain-of-thought reasoning or interact with external databases. Including additional actions for  AI to take increases the dimensionality of the space it can ``move'' through and allows for further feedback loops than simply interacting with a user. 

Recent AI systems allow for more flexible construction of artificial agents by linking multiple models and tools together. These frameworks can empower AI to utilize vector databases to store information, interact with web search tools, and even virtually move cursors and operate computers. Thus, the agent can learn in new ways by searching for clarification, moving a mouse to a different location, waiting for more data to arrive, and storing recent experiences. These actions can be enabled by decoding discrete actions from output tokens and engaging relevant peripherals, tools, and hardware. 

This separation of actions to specialized systems and components, moderated by an ``actor'' model, can potentially be  improved and better understood by examining its relationship to biology. Each component, such as the ``working memory'' of the context and ``long-term memory'' of a vector database, can be better understood and optimized toward a specific niche. AI models may also continue their specialization toward ``actor'' models, which  flexibly understand and orchestrate these actions, and ``world'' models, which incorporate general knowledge~\parencite{lecun_path_2022, team_kimi_2025}. This organization may also impart new actions, such as self-distilling new data and experiences stored in working memory toward a vector database or consolidating the memories stored within the database itself. More advanced uncertainty estimation models can empower AI to \textit{ask} users for clarification or direct itself to \textit{move} and \textit{detect} to address uncertainty (Figure \ref{fig:systems_software}). 

Additionally, while learning in artificial systems is often looked at as a monolithic, all-or-nothing action: will a network be retrained from scratch or not? However, the existence of multiple, interlinked processes for learning in the brain suggests that learning at multiple scales is possible. Further actions to implement learning could be implemented, such as distillation of databases into long-term memories and grounding of predictions with physical measurements and/or local sensors. These actions complement approaches such as reinforcement learning (RL), which provides a rich framework in which agents based on AI systems with a large variety of actions can be evaluated and improved. The potential of this approach was demonstrated through the famous DeepSeek-R1 model, which promoted reasoning behavior without the use of human-supervised data and an efficient evaluation method, and could be expended to include larger simulated environments and AI ``gyms'' \cite{DeepSeek2025, brockman_openai_2016}. 

By expanding available actions and feedback loops, AI can interact with the hardware in several ways: efficient implementation of multiple models, integration with multidomain sensors, motor control, and deciding on actions without large amounts of batching or high amounts of available power and cooling. 

Finally, by engaging in co-design --- allowing design factors to influence one another, rather than following a strict hierarchy --- approaches in software can be guided by available hardware, rather than hardware strictly implementing desired computational primitives. Available devices, such as oscillators, dynamical resistive devices, optoelectronic devices, or actual neurons and cells in xenobots, may be modeled and considered as primitives available to construct systems \parencite{csaba_coupled_2020, yang_memristive_2013, wang_memristors_2016, adair_resonate-and-fire_2025, blackiston_cellular_2021}. These models may themselves be neural networks, such as physically-informed neural networks (PINNs) which represent a compact model of a real, physical device that can be emulated quickly on a general-purpose computer, then physically realized later in hardware \cite{lagergren_biologically-informed_2020, rackauckas_universal_2021}. 

Navigating the possible design space of actions, hardware systems, and more requires iterating through many possible combinations in a manner which scales to the largest computational systems available while evaluating candidates without access to information such as gradients. Evolutionary algorithms (EA) have already been applied to explore and optimize many neural architectures and systems presented previously (Sections \ref{sec_modules} \& \ref{sec_systems}) (\parencite{ren_comprehensive_2022}). Utilizing or adapting current EA resources to address the design space for potential biologically inspired actions and systems would enable a sound theoretical footing to begin computational experimentation.

\subsection{Vectors for Hardware}

\begin{figure}[htbp]
    \centering
    \includegraphics[width=1\linewidth]{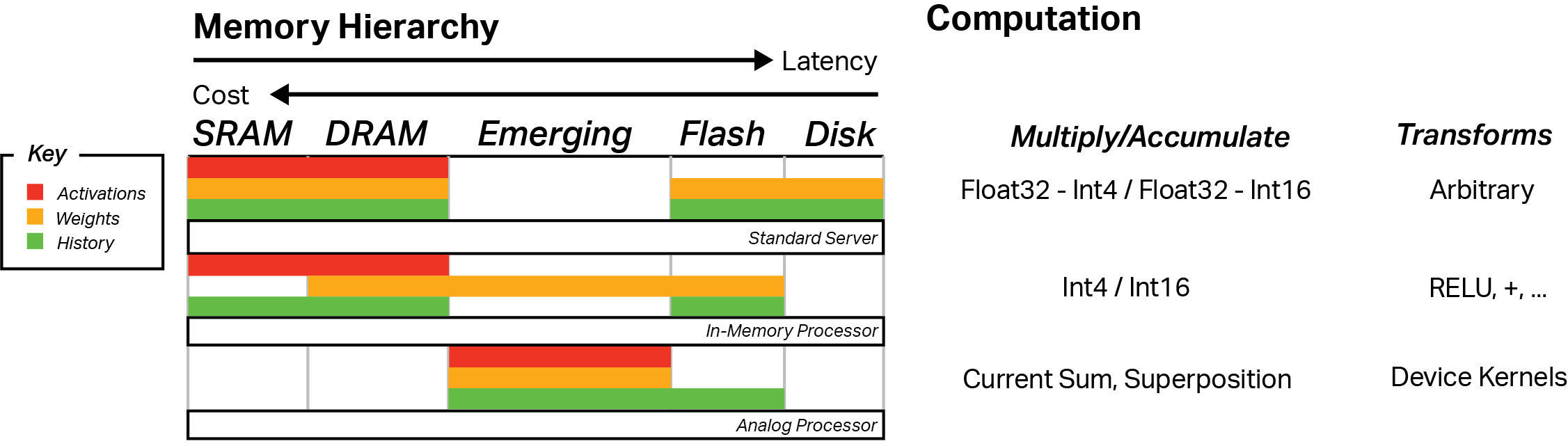}
    \caption{Current GPU systems used for AI training and inference follow a standard memory hierarchy, although in a massively parallel fashion. Local registers in SRAM fetch data from DRAM, flash memory, and disks. Process-in-memory (PIM) systems can reduce data movement and network congestion by allowing a subset of operations to take place within hardware, such as DRAM, flash, and emerging memories (e.g.,  RRAM, MRAM, PCRAM). A fully analog system may represent even highly variable data, such as activations, within inexpensive dynamical components and other emerging memories.  This can trade off cost and performance when embedded in a single-user environment.}
    \label{fig:systems_hardware}
\end{figure}

GPUs have remained the gold standard for datacenter-based inference and training of AI systems. In these systems, input, intermediate, and final activation values are stored in fast memory caches such as static and dynamic random-access memory (SRAM/DRAM). The weights used to calculate these activations must be loaded into the same fast GPU memory from solid state (flash) or spinning disks. This general-purpose architecture in which instructions and data share memory allows for general-purpose programming of arbitrary transformations on many datatypes (Figure \ref{fig:systems_hardware}).

However, embedding AI into real-world environments largely precludes batching of data and thus seriously affects the throughput and economics of architectures such as GPUs, which excel at massively parallel compute. Unbatched data cannot amortize the cost of retrieving and loading weights and other data from memory, suiting it to alternative architectures such as dataflow systems~\parencite{abts_think_2020, prabhakar_sambanova_2022, 10.1145/3414622.3432992}. These systems compile programs to intermediate representations and deploy computational subroutines that are embedded in smaller cores. This approach avoids the cost of transporting and waiting for parameters to be loaded from distant and slow memories such as DRAM. However, the density of available local memory for dataflow systems continues to be an issue. Even expensive wafer-scale systems can only implement approximately a hundred gigabytes of storage, meaning that these dataflow systems must be integrated through many die and systems and/or resort to using the external, dense memories~\parencite{hall_training_2023}. This problem is unlikely to resolve itself through simple scaling of digital technologies, since scaling of digital memory technologies such as SRAM and DRAM has slowed or ceased in recent years~\parencite{schor_iedm_2022}. As a result, denser, cheaper memories are a crucial step for embedded AI systems. 

Analog memories such as ReRAM have long been posited as a potential solution to this issue, and many significant advances have been made in this area, including multibit resistance levels, high endurance, long-term stability, and large-scale integration~\parencite{mittal_survey_2018, wan_compute--memory_2022}. However, simultaneously achieving these advances in a single, commercial-scale system continues to prove challenging and requires complex peripheral circuitry for handling the voltages and signals required to modulate the resistance of these memories.

Flash memory, in contrast, is  well established and has successfully scaled into 3D production, enabling memory densities to continue scaling. Additionally, flash memory has been integrated into experimental AI accelerators in roles similar to ReRAM to enable high-performance inferencing. In contrast to DRAM and SRAM, flash memory continues to scale vertically by adding more layers, increasing the aerial density of memory available. This comes at the cost of endurance, which is limited in flash; weights stored in this memory can  be adjusted only a finite number of times, as opposed to essentially infinite limits for SRAM and DRAM. However, by limiting Flash memories to the long-term ``world model'' section of an AI system, which is rarely reprogrammed, this limitation becomes largely irrelevant. Higher-endurance memories such as ReRAM, DRAM, and SRAM could largely be deployed on chips for long- and short-term memories involved in recording experience and executing distillation.

Novel memory architectures can integrate computing kernels within memory banks, allowing for certain operations to be processed in memory (PIM). This effectively separates data and instruction buses, avoiding the famous ``von Neumann bottleneck'' and reducing data movement, which remains an expensive overhead~\parencite{dally_high-performance_2015}. These memories may remain digital, as in the case of DRAM and flash PIM, or utilize analog encoding and transformations, as is the case for several RRAM PIMs~\parencite{lee_review_2024, kim_embedded_2022, wan_compute--memory_2022}. By avoiding data movement, PIMs can reduce the cost of synaptic operations, potentially orders of magnitude below digital approaches. However, these potential gains have been cut into by the decreasing cost of these operations in digital technologies through specialized representations (Section \ref{sec_neurons}) and the high cost of converting between analog and digital domains at the input and output of analog circuits. Thus, analog PIM approaches to synaptic operations can  succeed only if they can avoid high domain conversion costs, either by operating completely in the analog domain and/or using alternative encoding strategies with cheaper domain conversion methods,  for instance, encoding data in time or phase rather than amplitude. The area, circuitry, and instructional limitations of PIMs also may limit their capabilities to a subset of the transformations and datatypes available on other systems (Figure \ref{fig:systems_hardware}). 

Analog devices can engage such time- and phase-based coding techniques by engaging dynamics to transform signals through time. For instance, rather than working exclusively as a long-term memory, ReRAM systems can be engineered to have short-term resistance changes that dissipate quickly~\parencite{wang_memristors_2016}. These memories can be a powerful method for providing coincidence detection with low energy. Engaging these and other devices with digital pulses reminiscent of spikes can be used to construct oscillatory systems capable of advanced computations, not only encoding, but also computing with values in domains that offer different conversion strategies between analog and digital systems~\parencite{olin-ammentorp_hyperdimensional_2023}. These computational primitives could be applied to produce nonlinear activations and normalization processes in the analog domain. Additionally, these oscillatory systems and devices can recycle energy, maintaining computational ``momentum'' through time.

An additional capability of in-memory computation is in situ variance detection. That is, in some analog systems, the inputs and outputs can be switched or clamped, allowing for direct sensitivity analysis between output and input~\parencite{patrick_xiao_training_2024, pai_experimentally_2023}. This contrasts highly with digital processors, which must separately store and recalculate gradients for backwards passes. Local differentiation can support many alternative learning rules with greatly reduced costs for data transport and storage~\parencite{ororbia_brain-inspired_2023}. 

One common theme in AI systems is that the size of programs involved largely precludes single-die solutions from being effective: multiple-die, racks, and systems are involved in both training and inference. Increasingly efficient means for integrating multichip solutions through chiplets, photonics, and wafer bonding will continue to be a crucial enabling technology for both datacenter and embedded AI. These advanced interconnect technologies can overcome the limitation of photolithography, which effectively restricts fabrication of advanced circuits to two dimensions, whereas the brain can grow in three dimensions. By providing hardware that can increase the density of computation, utilize inexpensive memories, approximate novel in situ training techniques, and achieve compact integration, flexibly addressing the trade-offs required to reduce learning costs and effectively unbatched input streams can become possible. This may enable the deployment of the powerful techniques used today for language modeling in understanding and interacting with the world.

\subsection{Vectors for Supporting Research}

These initiatives require many advances to achieve the high-level objectives of creating AI architectures that can more freely self-direct actions, recall, and distillation. Supporting these objectives requires AI with different tokenization strategies and a latent space that  is closer to conceptual thought and movement than strictly adhering to language. Steps have already been taken in this direction but can be augmented to allow AI systems to take actions within their memory systems and attached peripherals~\parencite{team_large_2024}.

Integration of peripherals and multisensory data requires new benchmarks and datasets such as video content or virtual reinforcement learning ``gyms''~\parencite{brockman_openai_2016} However, creating the foundation models for such systems will likely remain expensive: open-source licensing of these systems once developed can be a powerful tool for driving further innovation and architectural experimentation. 

Furthermore, continual learning and catastrophic forgetting have previously been defined through a number of paradigms that focus primarily on classification tasks, making it difficult to adapt this framework to the autoregressive, generative nature of LLM models deployed today~\parencite{van_de_ven_three_2019}. Fine-tuning and alignment are continual learning problems: their goal is to modify the behavior of a pretrained LLM while preventing it from forgetting any knowledge gained during pretraining. New scenarios for continual learning should address this enormous need. 

We note that although continual learning techniques can enable powerful new capabilities, these may not be desirable under certain use cases. namely, for large-scale commercial applications such as today's AI chat systems. The ability of systems to self-modify in these situations is undesirable, as this could offer an attack vector for malicious users to circumvent or undo the fine-tuning and alignment processes that AI vendors implement in order to encourage their tools to be safe and useful. Continual learning may  prove useful only under certain situations where out-of-distribution data shifts are inevitable, such as in an environment where sensors may become damaged or occluded.

Developing hardware for real-world applications requires a multidisciplinary co-design  approach. Cheap, dense memories are one crucial enabling factor for future AI hardware. Novel coding strategies that can harness analog computation in these devices or engage with interesting device dynamics are highly complementary to memories. Any such novel memory or computing devices must additionally be simulated in large-scale systems and scale to address realistic, complex problems; demonstrations using MNIST---a 30-year old dataset---are insufficient to demonstrate that techniques can scale to the demands of modern AI systems~\parencite{xiao_fashion-mnist_2017}. 

The previous research vectors follow the disparities between AI and NI that we have described. However, these directions alone cannot address all issues necessary for an ``ultimate'' AI system. 

\subsection{Limitations of Biological Inspiration}

A key issue not addressed by current understandings of AI or NI is a concrete description of how these systems so effectively process natural data. Progress in understanding has been made through modeling ``infinitely wide'' neural layers, but questions such as how wide must a neural layer be, what is the most effective depth, how many attention blocks are optimal, what is an appropriate degree of overparametrization, and what are the best training strategies are still shrouded in mystery~\parencite{novak_neural_2019}. These choices are guided not by concrete theoretical understanding and constraints, but by empirical trends obtained at enormous expense \cite{dubey_llama_2024}. 

Lacking a sound theoretical underpinning, one may only guess at the true differences in capacities between two possible systems. Could current AI systems be shrunk down to a tenth or a hundredth of their current size through an architectural or mathematical trick? Or is overparametrization a key feature that allows generalization and stability? Theoretical underpinnings hint at the answers to these issues, but concrete resolution remains out of grasp~\parencite{bubeck_universal_2021}. Nonetheless, these issues must ultimately be resolved in order to allow AI systems to be truly engineered rather than brewed in a cauldron brimming with a costly brew of data, GPUs, and attention blocks. 

Furthermore, AI systems must be engineered to push the ``intelligence frontier'' beyond even the capabilities that have been demonstrated in biology: the Landauer limit of an irreversible computing operation is 18 meV at room temperature, approximately 10 orders of magnitude above today's most efficient computers~\parencite{plenio_physics_2001, Green500_2024}. By integrating new devices, physical representations, and computing methodologies, artificial computing can improve beyond even the impressive capabilities achieved by biological evolution. 

\section{Conclusion}

State-of-the-art AI systems utilizing LLMs are quickly changing the economy and driving massive investments into datacenters and related technologies. However, these AI systems face numerous limitations, including their cost to train and deploy, preventing them from many use cases. AI agents that more closely resemble natural intelligence, endowed with the ability to quickly and easily learn, explore, and develop, are necessary to overcome these challenges. Although AI has diverged from strict biological motivations, in many cases these foundations remain relevant or even convergent. We propose that by applying additional abilities to AI systems focusing on memory, movement, uncertainty, and exploration, artificial systems can develop the same strengths as natural intelligence. These advances depend on numerous developments in hardware and software, but none of these challenges appear insurmountable, and many are already underway. However, more cross-disciplinary conversation is necessary to recognize the parallels between different areas and develop common software, benchmarks, and tools. 

\section*{Acknowledgments}
AI was applied in this work to generate data analysis and plotting routines, collect recent literature, review drafts for typos and other grammatical errors, and generate several vector illustrations. 

Additional acknowledgments removed from this manuscript for anonymous review.




\printbibliography

@inproceedings{10.1145/3414622.3432992,
  title = {Wafer {{Scale Interconnect}} and {{Pathfinding}} for {{Machine Learning Hardware}} ({{Invited}})},
  booktitle = {Proceedings of the {{Workshop}} on {{System-Level Interconnect}}: {{Problems}} and {{Pathfinding Workshop}}},
  author = {Groeneveld, Patrick},
  year = {2020},
  publisher = {Association for Computing Machinery},
  address = {New York, NY, USA},
  doi = {10.1145/3414622.3432992},
  isbn = {978-1-4503-8106-2}
}

@article{siletti_transcriptomic_2023,
    title = {Transcriptomic diversity of cell types across the adult human brain},
    volume = {382},
    issn = {0036-8075, 1095-9203},
    url = {https://www.science.org/doi/10.1126/science.add7046},
    doi = {10.1126/science.add7046},
    language = {en},
    number = {6667},
    urldate = {2025-08-06},
    journal = {Science},
    author = {Siletti, Kimberly and Hodge, Rebecca and Mossi Albiach, Alejandro and Lee, Ka Wai and Ding, Song-Lin and Hu, Lijuan and Lönnerberg, Peter and Bakken, Trygve and Casper, Tamara and Clark, Michael and Dee, Nick and Gloe, Jessica and Hirschstein, Daniel and Shapovalova, Nadiya V. and Keene, C. Dirk and Nyhus, Julie and Tung, Herman and Yanny, Anna Marie and Arenas, Ernest and Lein, Ed S. and Linnarsson, Sten},
    month = oct,
    year = {2023},
    pages = {eadd7046},
}

@article{harnack_stability_2015,
    title = {Stability of {Neuronal} {Networks} with {Homeostatic} {Regulation}},
    volume = {11},
    issn = {1553-7358},
    url = {https://dx.plos.org/10.1371/journal.pcbi.1004357},
    doi = {10.1371/journal.pcbi.1004357},
    language = {en},
    number = {7},
    urldate = {2025-08-06},
    journal = {PLOS Computational Biology},
    author = {Harnack, Daniel and Pelko, Miha and Chaillet, Antoine and Chitour, Yacine and Van Rossum, Mark C.W.},
    editor = {Gutkin, Boris S.},
    month = jul,
    year = {2015},
    pages = {e1004357},
}

@article{oleary_neuronal_2011,
    title = {Neuronal homeostasis: time for a change?},
    volume = {589},
    copyright = {http://onlinelibrary.wiley.com/termsAndConditions\#vor},
    issn = {0022-3751, 1469-7793},
    shorttitle = {Neuronal homeostasis},
    url = {https://physoc.onlinelibrary.wiley.com/doi/10.1113/jphysiol.2011.210179},
    doi = {10.1113/jphysiol.2011.210179},
    language = {en},
    number = {20},
    urldate = {2025-08-07},
    journal = {The Journal of Physiology},
    author = {O’Leary, Timothy and Wyllie, David J. A.},
    month = oct,
    year = {2011},
    pages = {4811--4826},
}

@article{deneve_efficient_2016,
    title = {Efficient codes and balanced networks},
    volume = {19},
    issn = {15461726},
    doi = {10.1038/nn.4243},
    number = {3},
    journal = {Nature Neuroscience},
    author = {Denève, Sophie and Machens, Christian K.},
    month = feb,
    year = {2016},
    note = {Publisher: Nature Publishing Group},
    pages = {375--382},
}

@article{linnainmaa_taylor_1976,
    title = {Taylor expansion of the accumulated rounding error},
    volume = {16},
    copyright = {http://www.springer.com/tdm},
    issn = {0006-3835, 1572-9125},
    url = {http://link.springer.com/10.1007/BF01931367},
    doi = {10.1007/BF01931367},
    language = {en},
    number = {2},
    urldate = {2025-08-27},
    journal = {BIT},
    author = {Linnainmaa, Seppo},
    month = jun,
    year = {1976},
    pages = {146--160},
}

@article{hooker_hardware_2021,
    title = {The hardware lottery},
    volume = {64},
    issn = {15577317},
    doi = {10.1145/3467017},
    number = {12},
    journal = {Communications of the ACM},
    author = {Hooker, Sara},
    year = {2021},
    note = {arXiv: 2009.06489},
    pages = {58--65},
}

@article{WSJ_2025_AI_revolution,
    title = {Here's how big the {AI} revolution really is, in four charts},
    url = {https://www.wsj.com/tech/ai/ai-boom-companies-afb8c7e0},
    journal = {The Wall Street Journal},
    author = {{Rattner, Nate} and Seetharaman, Deepa},
    month = apr,
    year = {2025},
}

@article{balaraman2025transformer,
    title = {{US} faces transformer supply shortfall as power demand surges, {WoodMac} says},
    url = {https://www.reuters.com/business/energy/us-faces-transformer-supply-shortfall-power-demand-surges-woodmac-says-2025-08-14/},
    author = {Balaraman, Kavya},
    year = {2025},
}

@misc{paszke_pytorch_2019,
    title = {{PyTorch}: {An} {Imperative} {Style}, {High}-{Performance} {Deep} {Learning} {Library}},
    shorttitle = {{PyTorch}},
    url = {http://arxiv.org/abs/1912.01703},
    doi = {10.48550/arXiv.1912.01703},
    language = {en},
    urldate = {2025-08-27},
    publisher = {arXiv},
    author = {Paszke, Adam and Gross, Sam and Massa, Francisco and Lerer, Adam and Bradbury, James and Chanan, Gregory and Killeen, Trevor and Lin, Zeming and Gimelshein, Natalia and Antiga, Luca and Desmaison, Alban and Köpf, Andreas and Yang, Edward and DeVito, Zach and Raison, Martin and Tejani, Alykhan and Chilamkurthy, Sasank and Steiner, Benoit and Fang, Lu and Bai, Junjie and Chintala, Soumith},
    month = dec,
    year = {2019},
    note = {arXiv:1912.01703 [cs]},
}

@article{lecun_deep_2015,
    title = {Deep learning},
    volume = {521},
    issn = {14764687},
    doi = {10.1038/nature14539},
    number = {7553},
    journal = {Nature},
    author = {Lecun, Yann and Bengio, Yoshua and Hinton, Geoffrey},
    year = {2015},
    pmid = {26017442},
    pages = {436--444},
}

@misc{rackauckas_universal_2021,
	title = {Universal {Differential} {Equations} for {Scientific} {Machine} {Learning}},
	url = {http://arxiv.org/abs/2001.04385},
	doi = {10.48550/arXiv.2001.04385},
	language = {en},
	urldate = {2026-01-26},
	publisher = {arXiv},
	author = {Rackauckas, Christopher and Ma, Yingbo and Martensen, Julius and Warner, Collin and Zubov, Kirill and Supekar, Rohit and Skinner, Dominic and Ramadhan, Ali and Edelman, Alan},
	month = nov,
	year = {2021},
	note = {arXiv:2001.04385 [cs]},
}

@article{lagergren_biologically-informed_2020,
	title = {Biologically-informed neural networks guide mechanistic modeling from sparse experimental data},
	volume = {16},
	issn = {1553-7358},
	url = {https://dx.plos.org/10.1371/journal.pcbi.1008462},
	doi = {10.1371/journal.pcbi.1008462},
	language = {en},
	number = {12},
	urldate = {2026-01-26},
	journal = {PLOS Computational Biology},
	author = {Lagergren, John H. and Nardini, John T. and Baker, Ruth E. and Simpson, Matthew J. and Flores, Kevin B.},
	editor = {Lavrik, Inna},
	month = dec,
	year = {2020},
	pages = {e1008462},
}

@article{blackiston_cellular_2021,
	title = {A cellular platform for the development of synthetic living machines},
	volume = {6},
	issn = {2470-9476},
	url = {https://www.science.org/doi/10.1126/scirobotics.abf1571},
	doi = {10.1126/scirobotics.abf1571},
	language = {en},
	number = {52},
	urldate = {2026-01-26},
	journal = {Science Robotics},
	author = {Blackiston, Douglas and Lederer, Emma and Kriegman, Sam and Garnier, Simon and Bongard, Joshua and Levin, Michael},
	month = mar,
	year = {2021},
	pages = {eabf1571},
}

@article{guo_attention_2022,
	title = {Attention mechanisms in computer vision: {A} survey},
	volume = {8},
	issn = {2096-0662, 2096-0433},
	shorttitle = {Attention mechanisms in computer vision},
	url = {https://ieeexplore.ieee.org/document/10897531/},
	doi = {10.1007/s41095-022-0271-y},
	language = {en},
	number = {3},
	urldate = {2026-01-26},
	journal = {Computational Visual Media},
	author = {Guo, Meng-Hao and Xu, Tian-Xing and Liu, Jiang-Jiang and Liu, Zheng-Ning and Jiang, Peng-Tao and Mu, Tai-Jiang and Zhang, Song-Hai and Martin, Ralph R. and Cheng, Ming-Ming and Hu, Shi-Min},
	month = sep,
	year = {2022},
	pages = {331--368},
}

@misc{sun_efficient_2025,
	title = {Efficient {Attention} {Mechanisms} for {Large} {Language} {Models}: {A} {Survey}},
	shorttitle = {Efficient {Attention} {Mechanisms} for {Large} {Language} {Models}},
	url = {http://arxiv.org/abs/2507.19595},
	doi = {10.48550/arXiv.2507.19595},
	language = {en},
	urldate = {2026-01-26},
	publisher = {arXiv},
	author = {Sun, Yutao and Li, Zhenyu and Zhang, Yike and Pan, Tengyu and Dong, Bowen and Guo, Yuyi and Wang, Jianyong},
	month = aug,
	year = {2025},
	note = {arXiv:2507.19595 [cs]},
}

@article{henry_basal_2005,
	title = {Basal metabolic rate studies in humans: measurement and development of new equations},
	volume = {8},
	copyright = {https://www.cambridge.org/core/terms},
	issn = {1368-9800, 1475-2727},
	shorttitle = {Basal metabolic rate studies in humans},
	url = {https://www.cambridge.org/core/product/identifier/S1368980005001394/type/journal_article},
	doi = {10.1079/PHN2005801},
	language = {en},
	number = {7a},
	urldate = {2026-01-25},
	journal = {Public Health Nutrition},
	author = {Henry, Cjk},
	month = oct,
	year = {2005},
	pages = {1133--1152},
}

@misc{anil_gemini_2023,
	title = {Gemini: {A} {Family} of {Highly} {Capable} {Multimodal} {Models}},
	shorttitle = {Gemini},
	url = {http://arxiv.org/abs/2312.11805},
	language = {en},
	urldate = {2024-02-26},
	publisher = {arXiv},
	author = {Anil, Rohan and Borgeaud, Sebastian and Wu, Yonghui and Alayrac, Jean-Baptiste and Yu, Jiahui and Soricut, Radu and Schalkwyk, Johan and Dai, Andrew M. and Hauth, Anja and Millican, Katie and Silver, David and Petrov, Slav and Johnson, Melvin and Antonoglou, Ioannis and Schrittwieser, Julian and Glaese, Amelia and Chen, Jilin and Pitler, Emily and Lillicrap, Timothy and Lazaridou, Angeliki and Firat, Orhan and Molloy, James and Isard, Michael and Barham, Paul R. and Hennigan, Tom and Lee, Benjamin and Viola, Fabio and Reynolds, Malcolm and Xu, Yuanzhong and Doherty, Ryan and Collins, Eli and Meyer, Clemens and Rutherford, Eliza and Moreira, Erica and Ayoub, Kareem and Goel, Megha and Tucker, George and Piqueras, Enrique and Krikun, Maxim and Barr, Iain and Savinov, Nikolay and Danihelka, Ivo and Roelofs, Becca and White, Anaïs and Andreassen, Anders and von Glehn, Tamara and Yagati, Lakshman and Kazemi, Mehran and Gonzalez, Lucas and Khalman, Misha and Sygnowski, Jakub and Frechette, Alexandre and Smith, Charlotte and Culp, Laura and Proleev, Lev and Luan, Yi and Chen, Xi and Lottes, James and Schucher, Nathan and Lebron, Federico and Rrustemi, Alban and Clay, Natalie and Crone, Phil and Kocisky, Tomas and Zhao, Jeffrey and Perz, Bartek and Yu, Dian and Howard, Heidi and Bloniarz, Adam and Rae, Jack W. and Lu, Han and Sifre, Laurent and Maggioni, Marcello and Alcober, Fred and Garrette, Dan and Barnes, Megan and Thakoor, Shantanu and Austin, Jacob and Barth-Maron, Gabriel and Wong, William and Joshi, Rishabh and Chaabouni, Rahma and Fatiha, Deeni and Ahuja, Arun and Liu, Ruibo and Li, Yunxuan and Cogan, Sarah and Chen, Jeremy and Jia, Chao and Gu, Chenjie and Zhang, Qiao and Grimstad, Jordan and Hartman, Ale Jakse and Chadwick, Martin and Tomar, Gaurav Singh and Garcia, Xavier and Senter, Evan and Taropa, Emanuel and Pillai, Thanumalayan Sankaranarayana and Devlin, Jacob and Laskin, Michael and Casas, Diego de Las and Valter, Dasha and Tao, Connie and Blanco, Lorenzo and Badia, Adrià Puigdomènech and Reitter, David and Chen, Mianna and Brennan, Jenny and Rivera, Clara and Brin, Sergey and Iqbal, Shariq and Surita, Gabriela and Labanowski, Jane and Rao, Abhi and Winkler, Stephanie and Parisotto, Emilio and Gu, Yiming and Olszewska, Kate and Zhang, Yujing and Addanki, Ravi and Miech, Antoine and Louis, Annie and Shafey, Laurent El and Teplyashin, Denis and Brown, Geoff and Catt, Elliot and Attaluri, Nithya and Balaguer, Jan and Xiang, Jackie and Wang, Pidong and Ashwood, Zoe and Briukhov, Anton and Webson, Albert and Ganapathy, Sanjay and Sanghavi, Smit and Kannan, Ajay and Chang, Ming-Wei and Stjerngren, Axel and Djolonga, Josip and Sun, Yuting and Bapna, Ankur and Aitchison, Matthew and Pejman, Pedram and Michalewski, Henryk and Yu, Tianhe and Wang, Cindy and Love, Juliette and Ahn, Junwhan and Bloxwich, Dawn and Han, Kehang and Humphreys, Peter and Sellam, Thibault and Bradbury, James and Godbole, Varun and Samangooei, Sina and Damoc, Bogdan and Kaskasoli, Alex and Arnold, Sébastien M. R. and Vasudevan, Vijay and Agrawal, Shubham and Riesa, Jason and Lepikhin, Dmitry and Tanburn, Richard and Srinivasan, Srivatsan and Lim, Hyeontaek and Hodkinson, Sarah and Shyam, Pranav and Ferret, Johan and Hand, Steven and Garg, Ankush and Paine, Tom Le and Li, Jian and Li, Yujia and Giang, Minh and Neitz, Alexander and Abbas, Zaheer and York, Sarah and Reid, Machel and Cole, Elizabeth and Chowdhery, Aakanksha and Das, Dipanjan and Rogozińska, Dominika and Nikolaev, Vitaly and Sprechmann, Pablo and Nado, Zachary and Zilka, Lukas and Prost, Flavien and He, Luheng and Monteiro, Marianne and Mishra, Gaurav and Welty, Chris and Newlan, Josh and Jia, Dawei and Allamanis, Miltiadis and Hu, Clara Huiyi and de Liedekerke, Raoul and Gilmer, Justin and Saroufim, Carl and Rijhwani, Shruti and Hou, Shaobo and Shrivastava, Disha and Baddepudi, Anirudh and Goldin, Alex and Ozturel, Adnan and Cassirer, Albin and Xu, Yunhan and Sohn, Daniel and Sachan, Devendra and Amplayo, Reinald Kim and Swanson, Craig and Petrova, Dessie and Narayan, Shashi and Guez, Arthur and Brahma, Siddhartha and Landon, Jessica and Patel, Miteyan and Zhao, Ruizhe and Villela, Kevin and Wang, Luyu and Jia, Wenhao and Rahtz, Matthew and Giménez, Mai and Yeung, Legg and Lin, Hanzhao and Keeling, James and Georgiev, Petko and Mincu, Diana and Wu, Boxi and Haykal, Salem and Saputro, Rachel and Vodrahalli, Kiran and Qin, James and Cankara, Zeynep and Sharma, Abhanshu and Fernando, Nick and Hawkins, Will and Neyshabur, Behnam and Kim, Solomon and Hutter, Adrian and Agrawal, Priyanka and Castro-Ros, Alex and Driessche, George van den and Wang, Tao and Yang, Fan and Chang, Shuo-yiin and Komarek, Paul and McIlroy, Ross and Lučić, Mario and Zhang, Guodong and Farhan, Wael and Sharman, Michael and Natsev, Paul and Michel, Paul and Cheng, Yong and Bansal, Yamini and Qiao, Siyuan and Cao, Kris and Shakeri, Siamak and Butterfield, Christina and Chung, Justin and Rubenstein, Paul Kishan and Agrawal, Shivani and Mensch, Arthur and Soparkar, Kedar and Lenc, Karel and Chung, Timothy and Pope, Aedan and Maggiore, Loren and Kay, Jackie and Jhakra, Priya and Wang, Shibo and Maynez, Joshua and Phuong, Mary and Tobin, Taylor and Tacchetti, Andrea and Trebacz, Maja and Robinson, Kevin and Katariya, Yash and Riedel, Sebastian and Bailey, Paige and Xiao, Kefan and Ghelani, Nimesh and Aroyo, Lora and Slone, Ambrose and Houlsby, Neil and Xiong, Xuehan and Yang, Zhen and Gribovskaya, Elena and Adler, Jonas and Wirth, Mateo and Lee, Lisa and Li, Music and Kagohara, Thais and Pavagadhi, Jay and Bridgers, Sophie and Bortsova, Anna and Ghemawat, Sanjay and Ahmed, Zafarali and Liu, Tianqi and Powell, Richard and Bolina, Vijay and Iinuma, Mariko and Zablotskaia, Polina and Besley, James and Chung, Da-Woon and Dozat, Timothy and Comanescu, Ramona and Si, Xiance and Greer, Jeremy and Su, Guolong and Polacek, Martin and Kaufman, Raphaël Lopez and Tokumine, Simon and Hu, Hexiang and Buchatskaya, Elena and Miao, Yingjie and Elhawaty, Mohamed and Siddhant, Aditya and Tomasev, Nenad and Xing, Jinwei and Greer, Christina and Miller, Helen and Ashraf, Shereen and Roy, Aurko and Zhang, Zizhao and Ma, Ada and Filos, Angelos and Besta, Milos and Blevins, Rory and Klimenko, Ted and Yeh, Chih-Kuan and Changpinyo, Soravit and Mu, Jiaqi and Chang, Oscar and Pajarskas, Mantas and Muir, Carrie and Cohen, Vered and Lan, Charline Le and Haridasan, Krishna and Marathe, Amit and Hansen, Steven and Douglas, Sholto and Samuel, Rajkumar and Wang, Mingqiu and Austin, Sophia and Lan, Chang and Jiang, Jiepu and Chiu, Justin and Lorenzo, Jaime Alonso and Sjösund, Lars Lowe and Cevey, Sébastien and Gleicher, Zach and Avrahami, Thi and Boral, Anudhyan and Srinivasan, Hansa and Selo, Vittorio and May, Rhys and Aisopos, Konstantinos and Hussenot, Léonard and Soares, Livio Baldini and Baumli, Kate and Chang, Michael B. and Recasens, Adrià and Caine, Ben and Pritzel, Alexander and Pavetic, Filip and Pardo, Fabio and Gergely, Anita and Frye, Justin and Ramasesh, Vinay and Horgan, Dan and Badola, Kartikeya and Kassner, Nora and Roy, Subhrajit and Dyer, Ethan and Campos, Víctor and Tomala, Alex and Tang, Yunhao and Badawy, Dalia El and White, Elspeth and Mustafa, Basil and Lang, Oran and Jindal, Abhishek and Vikram, Sharad and Gong, Zhitao and Caelles, Sergi and Hemsley, Ross and Thornton, Gregory and Feng, Fangxiaoyu and Stokowiec, Wojciech and Zheng, Ce and Thacker, Phoebe and Ünlü, Çağlar and Zhang, Zhishuai and Saleh, Mohammad and Svensson, James and Bileschi, Max and Patil, Piyush and Anand, Ankesh and Ring, Roman and Tsihlas, Katerina and Vezer, Arpi and Selvi, Marco and Shevlane, Toby and Rodriguez, Mikel and Kwiatkowski, Tom and Daruki, Samira and Rong, Keran and Dafoe, Allan and FitzGerald, Nicholas and Gu-Lemberg, Keren and Khan, Mina and Hendricks, Lisa Anne and Pellat, Marie and Feinberg, Vladimir and Cobon-Kerr, James and Sainath, Tara and Rauh, Maribeth and Hashemi, Sayed Hadi and Ives, Richard and Hasson, Yana and Li, YaGuang and Noland, Eric and Cao, Yuan and Byrd, Nathan and Hou, Le and Wang, Qingze and Sottiaux, Thibault and Paganini, Michela and Lespiau, Jean-Baptiste and Moufarek, Alexandre and Hassan, Samer and Shivakumar, Kaushik and van Amersfoort, Joost and Mandhane, Amol and Joshi, Pratik and Goyal, Anirudh and Tung, Matthew and Brock, Andrew and Sheahan, Hannah and Misra, Vedant and Li, Cheng and Rakićević, Nemanja and Dehghani, Mostafa and Liu, Fangyu and Mittal, Sid and Oh, Junhyuk and Noury, Seb and Sezener, Eren and Huot, Fantine and Lamm, Matthew and De Cao, Nicola and Chen, Charlie and Elsayed, Gamaleldin and Chi, Ed and Mahdieh, Mahdis and Tenney, Ian and Hua, Nan and Petrychenko, Ivan and Kane, Patrick and Scandinaro, Dylan and Jain, Rishub and Uesato, Jonathan and Datta, Romina and Sadovsky, Adam and Bunyan, Oskar and Rabiej, Dominik and Wu, Shimu and Zhang, John and Vasudevan, Gautam and Leurent, Edouard and Alnahlawi, Mahmoud and Georgescu, Ionut and Wei, Nan and Zheng, Ivy and Chan, Betty and Rabinovitch, Pam G. and Stanczyk, Piotr and Zhang, Ye and Steiner, David and Naskar, Subhajit and Azzam, Michael and Johnson, Matthew and Paszke, Adam and Chiu, Chung-Cheng and Elias, Jaume Sanchez and Mohiuddin, Afroz and Muhammad, Faizan and Miao, Jin and Lee, Andrew and Vieillard, Nino and Potluri, Sahitya and Park, Jane and Davoodi, Elnaz and Zhang, Jiageng and Stanway, Jeff and Garmon, Drew and Karmarkar, Abhijit and Dong, Zhe and Lee, Jong and Kumar, Aviral and Zhou, Luowei and Evens, Jonathan and Isaac, William and Chen, Zhe and Jia, Johnson and Levskaya, Anselm and Zhu, Zhenkai and Gorgolewski, Chris and Grabowski, Peter and Mao, Yu and Magni, Alberto and Yao, Kaisheng and Snaider, Javier and Casagrande, Norman and Suganthan, Paul and Palmer, Evan and Irving, Geoffrey and Loper, Edward and Faruqui, Manaal and Arkatkar, Isha and Chen, Nanxin and Shafran, Izhak and Fink, Michael and Castaño, Alfonso and Giannoumis, Irene and Kim, Wooyeol and Rybiński, Mikołaj and Sreevatsa, Ashwin and Prendki, Jennifer and Soergel, David and Goedeckemeyer, Adrian and Gierke, Willi and Jafari, Mohsen and Gaba, Meenu and Wiesner, Jeremy and Wright, Diana Gage and Wei, Yawen and Vashisht, Harsha and Kulizhskaya, Yana and Hoover, Jay and Le, Maigo and Li, Lu and Iwuanyanwu, Chimezie and Liu, Lu and Ramirez, Kevin and Khorlin, Andrey and Cui, Albert and LIN, Tian and Georgiev, Marin and Wu, Marcus and Aguilar, Ricardo and Pallo, Keith and Chakladar, Abhishek and Repina, Alena and Wu, Xihui and van der Weide, Tom and Ponnapalli, Priya and Kaplan, Caroline and Simsa, Jiri and Li, Shuangfeng and Dousse, Olivier and Yang, Fan and Piper, Jeff and Ie, Nathan and Lui, Minnie and Pasumarthi, Rama and Lintz, Nathan and Vijayakumar, Anitha and Thiet, Lam Nguyen and Andor, Daniel and Valenzuela, Pedro and Paduraru, Cosmin and Peng, Daiyi and Lee, Katherine and Zhang, Shuyuan and Greene, Somer and Nguyen, Duc Dung and Kurylowicz, Paula and Velury, Sarmishta and Krause, Sebastian and Hardin, Cassidy and Dixon, Lucas and Janzer, Lili and Choo, Kiam and Feng, Ziqiang and Zhang, Biao and Singhal, Achintya and Latkar, Tejasi and Zhang, Mingyang and Le, Quoc and Abellan, Elena Allica and Du, Dayou and McKinnon, Dan and Antropova, Natasha and Bolukbasi, Tolga and Keller, Orgad and Reid, David and Finchelstein, Daniel and Raad, Maria Abi and Crocker, Remi and Hawkins, Peter and Dadashi, Robert and Gaffney, Colin and Lall, Sid and Franko, Ken and Filonov, Egor and Bulanova, Anna and Leblond, Rémi and Yadav, Vikas and Chung, Shirley and Askham, Harry and Cobo, Luis C. and Xu, Kelvin and Fischer, Felix and Xu, Jun and Sorokin, Christina and Alberti, Chris and Lin, Chu-Cheng and Evans, Colin and Zhou, Hao and Dimitriev, Alek and Forbes, Hannah and Banarse, Dylan and Tung, Zora and Liu, Jeremiah and Omernick, Mark and Bishop, Colton and Kumar, Chintu and Sterneck, Rachel and Foley, Ryan and Jain, Rohan and Mishra, Swaroop and Xia, Jiawei and Bos, Taylor and Cideron, Geoffrey and Amid, Ehsan and Piccinno, Francesco and Wang, Xingyu and Banzal, Praseem and Gurita, Petru and Noga, Hila and Shah, Premal and Mankowitz, Daniel J. and Polozov, Alex and Kushman, Nate and Krakovna, Victoria and Brown, Sasha and Bateni, MohammadHossein and Duan, Dennis and Firoiu, Vlad and Thotakuri, Meghana and Natan, Tom and Mohananey, Anhad and Geist, Matthieu and Mudgal, Sidharth and Girgin, Sertan and Li, Hui and Ye, Jiayu and Roval, Ofir and Tojo, Reiko and Kwong, Michael and Lee-Thorp, James and Yew, Christopher and Yuan, Quan and Bagri, Sumit and Sinopalnikov, Danila and Ramos, Sabela and Mellor, John and Sharma, Abhishek and Severyn, Aliaksei and Lai, Jonathan and Wu, Kathy and Cheng, Heng-Tze and Miller, David and Sonnerat, Nicolas and Vnukov, Denis and Greig, Rory and Beattie, Jennifer and Caveness, Emily and Bai, Libin and Eisenschlos, Julian and Korchemniy, Alex and Tsai, Tomy and Jasarevic, Mimi and Kong, Weize and Dao, Phuong and Zheng, Zeyu and Liu, Frederick and Yang, Fan and Zhu, Rui and Geller, Mark and Teh, Tian Huey and Sanmiya, Jason and Gladchenko, Evgeny and Trdin, Nejc and Sozanschi, Andrei and Toyama, Daniel and Rosen, Evan and Tavakkol, Sasan and Xue, Linting and Elkind, Chen and Woodman, Oliver and Carpenter, John and Papamakarios, George and Kemp, Rupert and Kafle, Sushant and Grunina, Tanya and Sinha, Rishika and Talbert, Alice and Goyal, Abhimanyu and Wu, Diane and Owusu-Afriyie, Denese and Du, Cosmo and Thornton, Chloe and Pont-Tuset, Jordi and Narayana, Pradyumna and Li, Jing and Fatehi, Sabaer and Wieting, John and Ajmeri, Omar and Uria, Benigno and Zhu, Tao and Ko, Yeongil and Knight, Laura and Héliou, Amélie and Niu, Ning and Gu, Shane and Pang, Chenxi and Tran, Dustin and Li, Yeqing and Levine, Nir and Stolovich, Ariel and Kalb, Norbert and Santamaria-Fernandez, Rebeca and Goenka, Sonam and Yustalim, Wenny and Strudel, Robin and Elqursh, Ali and Lakshminarayanan, Balaji and Deck, Charlie and Upadhyay, Shyam and Lee, Hyo and Dusenberry, Mike and Li, Zonglin and Wang, Xuezhi and Levin, Kyle and Hoffmann, Raphael and Holtmann-Rice, Dan and Bachem, Olivier and Yue, Summer and Arora, Sho and Malmi, Eric and Mirylenka, Daniil and Tan, Qijun and Koh, Christy and Yeganeh, Soheil Hassas and Põder, Siim and Zheng, Steven and Pongetti, Francesco and Tariq, Mukarram and Sun, Yanhua and Ionita, Lucian and Seyedhosseini, Mojtaba and Tafti, Pouya and Kotikalapudi, Ragha and Liu, Zhiyu and Gulati, Anmol and Liu, Jasmine and Ye, Xinyu and Chrzaszcz, Bart and Wang, Lily and Sethi, Nikhil and Li, Tianrun and Brown, Ben and Singh, Shreya and Fan, Wei and Parisi, Aaron and Stanton, Joe and Kuang, Chenkai and Koverkathu, Vinod and Choquette-Choo, Christopher A. and Li, Yunjie and Lu, T. J. and Ittycheriah, Abe and Shroff, Prakash and Sun, Pei and Varadarajan, Mani and Bahargam, Sanaz and Willoughby, Rob and Gaddy, David and Dasgupta, Ishita and Desjardins, Guillaume and Cornero, Marco and Robenek, Brona and Mittal, Bhavishya and Albrecht, Ben and Shenoy, Ashish and Moiseev, Fedor and Jacobsson, Henrik and Ghaffarkhah, Alireza and Rivière, Morgane and Walton, Alanna and Crepy, Clément and Parrish, Alicia and Liu, Yuan and Zhou, Zongwei and Farabet, Clement and Radebaugh, Carey and Srinivasan, Praveen and van der Salm, Claudia and Fidjeland, Andreas and Scellato, Salvatore and Latorre-Chimoto, Eri and Klimczak-Plucińska, Hanna and Bridson, David and de Cesare, Dario and Hudson, Tom and Mendolicchio, Piermaria and Walker, Lexi and Morris, Alex and Penchev, Ivo and Mauger, Matthew and Guseynov, Alexey and Reid, Alison and Odoom, Seth and Loher, Lucia and Cotruta, Victor and Yenugula, Madhavi and Grewe, Dominik and Petrushkina, Anastasia and Duerig, Tom and Sanchez, Antonio and Yadlowsky, Steve and Shen, Amy and Globerson, Amir and Kurzrok, Adam and Webb, Lynette and Dua, Sahil and Li, Dong and Lahoti, Preethi and Bhupatiraju, Surya and Hurt, Dan and Qureshi, Haroon and Agarwal, Ananth and Shani, Tomer and Eyal, Matan and Khare, Anuj and Belle, Shreyas Rammohan and Wang, Lei and Tekur, Chetan and Kale, Mihir Sanjay and Wei, Jinliang and Sang, Ruoxin and Saeta, Brennan and Liechty, Tyler and Sun, Yi and Zhao, Yao and Lee, Stephan and Nayak, Pandu and Fritz, Doug and Vuyyuru, Manish Reddy and Aslanides, John and Vyas, Nidhi and Wicke, Martin and Ma, Xiao and Bilal, Taylan and Eltyshev, Evgenii and Balle, Daniel and Martin, Nina and Cate, Hardie and Manyika, James and Amiri, Keyvan and Kim, Yelin and Xiong, Xi and Kang, Kai and Luisier, Florian and Tripuraneni, Nilesh and Madras, David and Guo, Mandy and Waters, Austin and Wang, Oliver and Ainslie, Joshua and Baldridge, Jason and Zhang, Han and Pruthi, Garima and Bauer, Jakob and Yang, Feng and Mansour, Riham and Gelman, Jason and Xu, Yang and Polovets, George and Liu, Ji and Cai, Honglong and Chen, Warren and Sheng, XiangHai and Xue, Emily and Ozair, Sherjil and Yu, Adams and Angermueller, Christof and Li, Xiaowei and Wang, Weiren and Wiesinger, Julia and Koukoumidis, Emmanouil and Tian, Yuan and Iyer, Anand and Gurumurthy, Madhu and Goldenson, Mark and Shah, Parashar and Blake, M. K. and Yu, Hongkun and Urbanowicz, Anthony and Palomaki, Jennimaria and Fernando, Chrisantha and Brooks, Kevin and Durden, Ken and Mehta, Harsh and Momchev, Nikola and Rahimtoroghi, Elahe and Georgaki, Maria and Raul, Amit and Ruder, Sebastian and Redshaw, Morgan and Lee, Jinhyuk and Jalan, Komal and Li, Dinghua and Perng, Ginger and Hechtman, Blake and Schuh, Parker and Nasr, Milad and Chen, Mia and Milan, Kieran and Mikulik, Vladimir and Strohman, Trevor and Franco, Juliana and Green, Tim and Hassabis, Demis and Kavukcuoglu, Koray and Dean, Jeffrey and Vinyals, Oriol},
	month = dec,
	year = {2023},
	note = {arXiv:2312.11805 [cs]},
}

@misc{google_gemini_tokens,
	title = {Understanding and counting tokens},
	url = {https://ai.google.dev/gemini-api/docs/tokens?lang=python},
	author = {{Google AI}},
	year = {2024},
}

@misc{gptspace_gpt_tokens_guide,
	title = {Understanding {OpenAI} {GPT} tokens: a comprehensive guide},
	url = {https://gpt.space/blog/understanding-openai-gpt-tokens-a-comprehensive-guide},
	author = {{OpenAI}},
	year = {2024},
}

@inproceedings{jun_hbm_2017,
	address = {Monterey, CA, USA},
	title = {{HBM} ({High} {Bandwidth} {Memory}) {DRAM} {Technology} and {Architecture}},
	isbn = {978-1-5090-3274-7},
	url = {http://ieeexplore.ieee.org/document/7939084/},
	doi = {10.1109/IMW.2017.7939084},
	language = {en},
	urldate = {2026-01-21},
	booktitle = {2017 {IEEE} {International} {Memory} {Workshop} ({IMW})},
	publisher = {IEEE},
	author = {Jun, Hongshin and Cho, Jinhee and Lee, Kangseol and Son, Ho-Young and Kim, Kwiwook and Jin, Hanho and Kim, Keith},
	month = may,
	year = {2017},
	pages = {1--4},
}

@techreport{nvidia_h100_gpu_whitepaper,
	title = {{NVIDIA} {H100} {GPU} whitepaper},
	url = {https://resources.nvidia.com/en-us-hopper-architecture/nvidia-h100-tensor-c},
	institution = {NVIDIA Corporation},
	author = {{NVIDIA Corporation}},
	year = {2022},
}

@misc{EpochMachineLearningHardware2024,
	title = {Data on machine learning hardware},
	url = {https://epoch.ai/data/machine-learning-hardware},
	author = {{Epoch AI}},
	month = oct,
	year = {2024},
}

@incollection{Koch2024,
	address = {Cham},
	title = {The brain: {Neurons}, synapses, and neural networks},
	isbn = {978-3-031-38971-9},
	url = {https://doi.org/10.1007/978-3-031-38971-9_734-1},
	booktitle = {Encyclopedia of religious psychology and behavior},
	publisher = {Springer Nature Switzerland},
	author = {Koch, Christopher},
	editor = {Shackelford, Todd K.},
	year = {2024},
	doi = {10.1007/978-3-031-38971-9_734-1},
	pages = {1--4},
}

@inproceedings{luo_integration_2024,
	address = {Changsha China},
	title = {Integration of {LLMs} and the {Physical} {World}: {Research} and {Application}},
	isbn = {979-8-4007-1011-7},
	shorttitle = {Integration of {LLMs} and the {Physical} {World}},
	url = {https://dl.acm.org/doi/10.1145/3674399.3674402},
	doi = {10.1145/3674399.3674402},
	language = {en},
	urldate = {2026-01-07},
	booktitle = {{ACM} {Turing} {Award} {Celebration} {Conference} 2024},
	publisher = {ACM},
	author = {Luo, Xiaoyu and Liu, Daping and Dang, Fan and Luo, Hanjiang},
	month = jul,
	year = {2024},
	pages = {1--5},
}

@inproceedings{mlenergy-neuripsdb25,
	title = {The {ML}.{ENERGY} benchmark: {Toward} automated inference energy measurement and optimization},
	booktitle = {{NeurIPS} datasets and benchmarks},
	author = {Chung, Jae-Won and Ma, Jeff J. and Wu, Ruofan and Liu, Jiachen and Kweon, Oh Jun and Xia, Yuxuan and Wu, Zhiyu and Chowdhury, Mosharaf},
	year = {2025},
}

@article{zheng_unbearable_2025,
	title = {The unbearable slowness of being: {Why} do we live at 10 bits/s?},
	volume = {113},
	issn = {08966273},
	shorttitle = {The unbearable slowness of being},
	url = {https://linkinghub.elsevier.com/retrieve/pii/S0896627324008080},
	doi = {10.1016/j.neuron.2024.11.008},
	language = {en},
	number = {2},
	urldate = {2025-12-17},
	journal = {Neuron},
	author = {Zheng, Jieyu and Meister, Markus},
	month = jan,
	year = {2025},
	pages = {192--204},
}

@misc{adair_resonate-and-fire_2025,
	title = {Resonate-and-{Fire} {Photonic}-{Electronic} {Spiking} {Neurons} for {Fast} and {Efficient} {Light}-{Enabled} {Neuromorphic} {Processing} {Systems}},
	url = {http://arxiv.org/abs/2510.14515},
	doi = {10.48550/arXiv.2510.14515},
	language = {en},
	urldate = {2025-12-17},
	publisher = {arXiv},
	author = {Adair, Andrew and Owen-Newns, Dafydd and Donati, Giovanni and Robertson, Joshua and Figueiredo, José and Wasige, Eduard and Al-Taai, Qusay and Romeira, Bruno and Hejda, Matěj and Hurtado, Antonio},
	month = oct,
	year = {2025},
	note = {arXiv:2510.14515 [physics]},
}

@article{wu_understanding_2023,
	title = {Understanding {INT4} {Quantization} for {Language} {Models}: {Latency} {Speedup}, {Composability}, and {Failure} {Cases}},
	url = {https://arxiv.org/abs/2301.12017},
	language = {en},
	author = {Wu, Xiaoxia and Li, Cheng and Aminabadi, Reza Yazdani and Yao, Zhewei and He, Yuxiong},
	year = {2023},
}

@article{mccarthy_proposal_2006,
	title = {A {Proposal} for the {Dartmouth} {Summer} {Research} {Project} on {Artificial} {Intelligence}},
	url = {https://ojs.aaai.org/aimagazine/index.php/aimagazine/article/view/1904},
	language = {en},
	author = {McCarthy, John},
	year = {2006},
}

@article{hering2001dentritic,
	title = {Dentritic spines: structure, dynamics and regulation},
	volume = {2},
	number = {12},
	journal = {Nature Reviews Neuroscience},
	author = {Hering, Heike and Sheng, Morgan},
	year = {2001},
	note = {Publisher: Nature Publishing Group UK London},
	pages = {880--888},
}

@article{hall_training_2023,
	title = {Training {Giant} {Neural} {Networks} {Using} {Weight} {Streaming} on {Cerebras} {Wafer}-{Scale} {Clusters}},
	url = {https://8968533.fs1.hubspotusercontent-na1.net/hubfs/8968533/Virtual%2520Booth%2520Docs/CS%2520Weight%2520Streaming%2520White%2520Paper.pdf&ved=2ahUKEwjCxJrhqN6PAxWakokEHVnAFEUQFnoECBgQAQ&usg=AOvVaw2hCbdbt_sg9CtMABk9YF_P},
	language = {en},
	author = {Hall, Stewart and Schreiber, Rob and Lie, Sean and Systems, Cerebras},
	year = {2023},
}

@article{baydin2018automatic,
	title = {Automatic {Differentiation} in {Machine} {Learning}: {A} {Survey}},
	volume = {18},
	number = {153},
	journal = {Journal of machine learning research},
	author = {Baydin, Atilim Gunes and Pearlmutter, Barak A and Radul, Alexey Andreyevich and Siskind, Jeffrey Mark},
	year = {2018},
	pages = {1--43},
}

@article{patrick_xiao_training_2024,
	title = {Training neural networks using physical equations of motion},
	volume = {121},
	issn = {0027-8424, 1091-6490},
	url = {https://pnas.org/doi/10.1073/pnas.2411913121},
	doi = {10.1073/pnas.2411913121},
	language = {en},
	number = {30},
	urldate = {2025-08-29},
	journal = {Proceedings of the National Academy of Sciences},
	author = {Patrick Xiao, T.},
	month = jul,
	year = {2024},
	pages = {e2411913121},
}

@article{pai_experimentally_2023,
	title = {Experimentally realized in situ backpropagation for deep learning in photonic neural networks},
	volume = {380},
	issn = {0036-8075, 1095-9203},
	url = {https://www.science.org/doi/10.1126/science.ade8450},
	doi = {10.1126/science.ade8450},
	language = {en},
	number = {6643},
	urldate = {2025-08-29},
	journal = {Science},
	author = {Pai, Sunil and Sun, Zhanghao and Hughes, Tyler W. and Park, Taewon and Bartlett, Ben and Williamson, Ian A. D. and Minkov, Momchil and Milanizadeh, Maziyar and Abebe, Nathnael and Morichetti, Francesco and Melloni, Andrea and Fan, Shanhui and Solgaard, Olav and Miller, David A. B.},
	month = apr,
	year = {2023},
	pages = {398--404},
}

@misc{brockman_openai_2016,
	title = {{OpenAI} {Gym}},
	url = {http://arxiv.org/abs/1606.01540},
	doi = {10.48550/arXiv.1606.01540},
	language = {en},
	urldate = {2025-08-29},
	publisher = {arXiv},
	author = {Brockman, Greg and Cheung, Vicki and Pettersson, Ludwig and Schneider, Jonas and Schulman, John and Tang, Jie and Zaremba, Wojciech},
	month = jun,
	year = {2016},
	note = {arXiv:1606.01540 [cs]},
}

@misc{team_kimi_2025,
	title = {Kimi {K2}: {Open} {Agentic} {Intelligence}},
	shorttitle = {Kimi {K2}},
	url = {http://arxiv.org/abs/2507.20534},
	doi = {10.48550/arXiv.2507.20534},
	language = {en},
	urldate = {2025-08-29},
	publisher = {arXiv},
	author = {Team, Kimi and Bai, Yifan and Bao, Yiping and Chen, Guanduo and Chen, Jiahao and Chen, Ningxin and Chen, Ruijue and Chen, Yanru and Chen, Yuankun and Chen, Yutian and Chen, Zhuofu and Cui, Jialei and Ding, Hao and Dong, Mengnan and Du, Angang and Du, Chenzhuang and Du, Dikang and Du, Yulun and Fan, Yu and Feng, Yichen and Fu, Kelin and Gao, Bofei and Gao, Hongcheng and Gao, Peizhong and Gao, Tong and Gu, Xinran and Guan, Longyu and Guo, Haiqing and Guo, Jianhang and Hu, Hao and Hao, Xiaoru and He, Tianhong and He, Weiran and He, Wenyang and Hong, Chao and Hu, Yangyang and Hu, Zhenxing and Huang, Weixiao and Huang, Zhiqi and Huang, Zihao and Jiang, Tao and Jiang, Zhejun and Jin, Xinyi and Kang, Yongsheng and Lai, Guokun and Li, Cheng and Li, Fang and Li, Haoyang and Li, Ming and Li, Wentao and Li, Yanhao and Li, Yiwei and Li, Zhaowei and Li, Zheming and Lin, Hongzhan and Lin, Xiaohan and Lin, Zongyu and Liu, Chengyin and Liu, Chenyu and Liu, Hongzhang and Liu, Jingyuan and Liu, Junqi and Liu, Liang and Liu, Shaowei and Liu, T. Y. and Liu, Tianwei and Liu, Weizhou and Liu, Yangyang and Liu, Yibo and Liu, Yiping and Liu, Yue and Liu, Zhengying and Lu, Enzhe and Lu, Lijun and Ma, Shengling and Ma, Xinyu and Ma, Yingwei and Mao, Shaoguang and Mei, Jie and Men, Xin and Miao, Yibo and Pan, Siyuan and Peng, Yebo and Qin, Ruoyu and Qu, Bowen and Shang, Zeyu and Shi, Lidong and Shi, Shengyuan and Song, Feifan and Su, Jianlin and Su, Zhengyuan and Sun, Xinjie and Sung, Flood and Tang, Heyi and Tao, Jiawen and Teng, Qifeng and Wang, Chensi and Wang, Dinglu and Wang, Feng and Wang, Haiming and Wang, Jianzhou and Wang, Jiaxing and Wang, Jinhong and Wang, Shengjie and Wang, Shuyi and Wang, Yao and Wang, Yejie and Wang, Yiqin and Wang, Yuxin and Wang, Yuzhi and Wang, Zhaoji and Wang, Zhengtao and Wang, Zhexu and Wei, Chu and Wei, Qianqian and Wu, Wenhao and Wu, Xingzhe and Wu, Yuxin and Xiao, Chenjun and Xie, Xiaotong and Xiong, Weimin and Xu, Boyu and Xu, Jing and Xu, Jinjing and Xu, L. H. and Xu, Lin and Xu, Suting and Xu, Weixin and Xu, Xinran and Xu, Yangchuan and Xu, Ziyao and Yan, Junjie and Yan, Yuzi and Yang, Xiaofei and Yang, Ying and Yang, Zhen and Yang, Zhilin and Yang, Zonghan and Yao, Haotian and Yao, Xingcheng and Ye, Wenjie and Ye, Zhuorui and Yin, Bohong and Yu, Longhui and Yuan, Enming and Yuan, Hongbang and Yuan, Mengjie and Zhan, Haobing and Zhang, Dehao and Zhang, Hao and Zhang, Wanlu and Zhang, Xiaobin and Zhang, Yangkun and Zhang, Yizhi and Zhang, Yongting and Zhang, Yu and Zhang, Yutao and Zhang, Yutong and Zhang, Zheng and Zhao, Haotian and Zhao, Yikai and Zheng, Huabin and Zheng, Shaojie and Zhou, Jianren and Zhou, Xinyu and Zhou, Zaida and Zhu, Zhen and Zhuang, Weiyu and Zu, Xinxing},
	month = jul,
	year = {2025},
	note = {arXiv:2507.20534 [cs]},
}

@article{lecun_path_2022,
	title = {A {Path} {Towards} {Autonomous} {Machine} {Intelligence} {Version} 0.9.2, 2022-06-27},
	language = {en},
	author = {LeCun, Yann},
	month = jun,
	year = {2022},
}

@article{lee_review_2024,
	title = {Review of neuromorphic computing based on {NAND} flash memory},
	volume = {9},
	issn = {2055-6756, 2055-6764},
	url = {https://xlink.rsc.org/?DOI=D3NH00532A},
	doi = {10.1039/D3NH00532A},
	language = {en},
	number = {9},
	urldate = {2025-08-29},
	journal = {Nanoscale Horizons},
	author = {Lee, Sung-Tae and Lee, Jong-Ho},
	year = {2024},
	pages = {1475--1492},
}

@article{kim_embedded_2022,
	title = {An {Embedded} nand {Flash}-{Based} {Compute}-{In}-{Memory} {Array} {Demonstrated} in a {Standard} {Logic} {Process}},
	volume = {57},
	copyright = {https://ieeexplore.ieee.org/Xplorehelp/downloads/license-information/IEEE.html},
	issn = {0018-9200, 1558-173X},
	url = {https://ieeexplore.ieee.org/document/9501165/},
	doi = {10.1109/JSSC.2021.3098671},
	language = {en},
	number = {2},
	urldate = {2025-08-29},
	journal = {IEEE Journal of Solid-State Circuits},
	author = {Kim, Minsu and Liu, Muqing and Everson, Luke R. and Kim, Chris H.},
	month = feb,
	year = {2022},
	pages = {625--638},
}

@article{slors_notebooks_2020,
	title = {From {Notebooks} to {Institutions}: {The} {Case} for {Symbiotic} {Cognition}},
	volume = {11},
	issn = {1664-1078},
	shorttitle = {From {Notebooks} to {Institutions}},
	url = {https://www.frontiersin.org/article/10.3389/fpsyg.2020.00674/full},
	doi = {10.3389/fpsyg.2020.00674},
	language = {en},
	urldate = {2025-08-27},
	journal = {Frontiers in Psychology},
	author = {Slors, Marc},
	month = apr,
	year = {2020},
	pages = {674},
}

@article{pallasdies_single_2019,
	title = {From single neurons to behavior in the jellyfish {Aurelia} aurita},
	volume = {8},
	issn = {2050-084X},
	url = {https://elifesciences.org/articles/50084},
	doi = {10.7554/eLife.50084},
	language = {en},
	urldate = {2025-08-26},
	journal = {eLife},
	author = {Pallasdies, Fabian and Goedeke, Sven and Braun, Wilhelm and Memmesheimer, Raoul-Martin},
	month = dec,
	year = {2019},
	pages = {e50084},
}

@article{herculano-houzel_scaling_2011,
	title = {Scaling of {Brain} {Metabolism} with a {Fixed} {Energy} {Budget} per {Neuron}: {Implications} for {Neuronal} {Activity}, {Plasticity} and {Evolution}},
	volume = {6},
	issn = {1932-6203},
	shorttitle = {Scaling of {Brain} {Metabolism} with a {Fixed} {Energy} {Budget} per {Neuron}},
	url = {https://dx.plos.org/10.1371/journal.pone.0017514},
	doi = {10.1371/journal.pone.0017514},
	language = {en},
	number = {3},
	urldate = {2025-08-26},
	journal = {PLoS ONE},
	author = {Herculano-Houzel, Suzana},
	editor = {Perc, Matjaz},
	month = mar,
	year = {2011},
	pages = {e17514},
}

@article{herculano-houzel_numbers_2017,
	title = {Numbers of neurons as biological correlates of cognitive capability},
	volume = {16},
	issn = {23521546},
	url = {https://linkinghub.elsevier.com/retrieve/pii/S2352154616302637},
	doi = {10.1016/j.cobeha.2017.02.004},
	language = {en},
	urldate = {2025-08-26},
	journal = {Current Opinion in Behavioral Sciences},
	author = {Herculano-Houzel, Suzana},
	month = aug,
	year = {2017},
	pages = {1--7},
}

@article{puschel_hominin_2024,
	title = {Hominin brain size increase has emerged from within-species encephalization},
	volume = {121},
	issn = {0027-8424, 1091-6490},
	url = {https://pnas.org/doi/10.1073/pnas.2409542121},
	doi = {10.1073/pnas.2409542121},
	language = {en},
	number = {49},
	urldate = {2025-08-26},
	journal = {Proceedings of the National Academy of Sciences},
	author = {Püschel, Thomas A. and Nicholson, Samuel L. and Baker, Joanna and Barton, Robert A. and Venditti, Chris},
	month = dec,
	year = {2024},
	pages = {e2409542121},
}

@article{miller_quantitative_2019,
	title = {Quantitative uniqueness of human brain evolution revealed through phylogenetic comparative analysis},
	volume = {8},
	copyright = {http://creativecommons.org/licenses/by/4.0/},
	issn = {2050-084X},
	url = {https://elifesciences.org/articles/41250},
	doi = {10.7554/eLife.41250},
	language = {en},
	urldate = {2025-08-26},
	journal = {eLife},
	author = {Miller, Ian F and Barton, Robert A and Nunn, Charles L},
	month = jan,
	year = {2019},
	pages = {e41250},
}

@article{heger_genetic_2020,
	title = {The genetic factors of bilaterian evolution},
	volume = {9},
	issn = {2050-084X},
	url = {https://elifesciences.org/articles/45530},
	doi = {10.7554/eLife.45530},
	language = {en},
	urldate = {2025-08-26},
	journal = {eLife},
	author = {Heger, Peter and Zheng, Wen and Rottmann, Anna and Panfilio, Kristen A and Wiehe, Thomas},
	month = jul,
	year = {2020},
	pages = {e45530},
}

@article{bosch_back_2017,
	title = {Back to the {Basics}: {Cnidarians} {Start} to {Fire}},
	volume = {40},
	issn = {01662236},
	shorttitle = {Back to the {Basics}},
	url = {https://linkinghub.elsevier.com/retrieve/pii/S0166223616301680},
	doi = {10.1016/j.tins.2016.11.005},
	language = {en},
	number = {2},
	urldate = {2025-08-26},
	journal = {Trends in Neurosciences},
	author = {Bosch, Thomas C.G. and Klimovich, Alexander and Domazet-Lošo, Tomislav and Gründer, Stefan and Holstein, Thomas W. and Jékely, Gáspár and Miller, David J. and Murillo-Rincon, Andrea P. and Rentzsch, Fabian and Richards, Gemma S. and Schröder, Katja and Technau, Ulrich and Yuste, Rafael},
	month = feb,
	year = {2017},
	pages = {92--105},
}

@article{buzsaki_scaling_2013,
	title = {Scaling {Brain} {Size}, {Keeping} {Timing}: {Evolutionary} {Preservation} of {Brain} {Rhythms}},
	volume = {80},
	issn = {08966273},
	shorttitle = {Scaling {Brain} {Size}, {Keeping} {Timing}},
	url = {https://linkinghub.elsevier.com/retrieve/pii/S0896627313009045},
	doi = {10.1016/j.neuron.2013.10.002},
	language = {en},
	number = {3},
	urldate = {2025-08-20},
	journal = {Neuron},
	author = {Buzsáki, György and Logothetis, Nikos and Singer, Wolf},
	month = oct,
	year = {2013},
	pages = {751--764},
}

@article{lemon_classification_2021,
	title = {Classification of {Cortical} {Neurons} by {Spike} {Shape} and the {Identification} of {Pyramidal} {Neurons}},
	volume = {31},
	copyright = {http://creativecommons.org/licenses/by/4.0/},
	issn = {1047-3211, 1460-2199},
	url = {https://academic.oup.com/cercor/article/31/11/5131/6297166},
	doi = {10.1093/cercor/bhab147},
	language = {en},
	number = {11},
	urldate = {2025-08-19},
	journal = {Cerebral Cortex},
	author = {Lemon, Roger N and Baker, Stuart N and Kraskov, Alexander},
	month = oct,
	year = {2021},
	pages = {5131--5138},
}

@article{santuy_estimation_2020,
	title = {Estimation of the number of synapses in the hippocampus and brain-wide by volume electron microscopy and genetic labeling},
	volume = {10},
	issn = {2045-2322},
	url = {https://www.nature.com/articles/s41598-020-70859-5},
	doi = {10.1038/s41598-020-70859-5},
	language = {en},
	number = {1},
	urldate = {2025-08-19},
	journal = {Scientific Reports},
	author = {Santuy, Andrea and Tomás-Roca, Laura and Rodríguez, José-Rodrigo and González-Soriano, Juncal and Zhu, Fei and Qiu, Zhen and Grant, Seth G. N. and DeFelipe, Javier and Merchan-Perez, Angel},
	month = aug,
	year = {2020},
	pages = {14014},
}

@article{wiener1948cybernetics,
	title = {Cybernetics},
	volume = {179},
	number = {5},
	journal = {Scientific American},
	author = {Wiener, Norbert},
	year = {1948},
	note = {Publisher: JSTOR},
	pages = {14--19},
}

@article{lui_development_2011,
	title = {Development and {Evolution} of the {Human} {Neocortex}},
	volume = {146},
	issn = {00928674},
	url = {https://linkinghub.elsevier.com/retrieve/pii/S0092867411007057},
	doi = {10.1016/j.cell.2011.06.030},
	language = {en},
	number = {1},
	urldate = {2025-08-18},
	journal = {Cell},
	author = {Lui, Jan H. and Hansen, David V. and Kriegstein, Arnold R.},
	month = jul,
	year = {2011},
	pages = {18--36},
}

@article{wei_chain--thought_2022,
	title = {Chain-of-{Thought} {Prompting} {Elicits} {Reasoning} in {Large} {Language} {Models}},
	language = {en},
	author = {Wei, Jason and Wang, Xuezhi and Schuurmans, Dale and Bosma, Maarten and Ichter, Brian and Xia, Fei and Chi, Ed H and Le, Quoc V and Zhou, Denny},
	year = {2022},
}

@misc{liDiscreteTokenizationMultimodal2025,
	title = {Discrete {Tokenization} for {Multimodal} {LLMs}: {A} {Comprehensive} {Survey}},
	shorttitle = {Discrete {Tokenization} for {Multimodal} {LLMs}},
	url = {http://arxiv.org/abs/2507.22920},
	doi = {10.48550/arXiv.2507.22920},
	language = {en},
	urldate = {2025-08-15},
	publisher = {arXiv},
	author = {Li, Jindong and Fu, Yali and Liu, Jiahong and Cao, Linxiao and Ji, Wei and Yang, Menglin and King, Irwin and Yang, Ming-Hsuan},
	month = jul,
	year = {2025},
	note = {arXiv:2507.22920 [cs]},
}

@inproceedings{garner1959residue,
	title = {The residue number system},
	booktitle = {Papers presented at the the {March} 3-5, 1959, western joint computer conference},
	author = {Garner, Harvey L},
	year = {1959},
	pages = {146--153},
}

@misc{epoch2023announcingupdatedpcddatabase,
	title = {Announcing epoch ai’s updated parameter, compute and data trends database},
	url = {https://epoch.ai/blog/announcing-updated-pcd-database},
	author = {{Epoch AI}},
	year = {2023},
}

@misc{yu_reverse_2025,
	title = {Reverse {Modeling} in {Large} {Language} {Models}},
	url = {http://arxiv.org/abs/2410.09817},
	doi = {10.48550/arXiv.2410.09817},
	language = {en},
	urldate = {2025-08-04},
	publisher = {arXiv},
	author = {Yu, Sicheng and Xu, Yuanchen and Du, Cunxiao and Zhou, Yanying and Qiu, Minghui and Sun, Qianru and Zhang, Hao and Wu, Jiawei},
	month = feb,
	year = {2025},
	note = {arXiv:2410.09817 [cs]},
}

@misc{cai_survey_2024,
	title = {A {Survey} on {Mixture} of {Experts}},
	url = {http://arxiv.org/abs/2407.06204},
	doi = {10.48550/arXiv.2407.06204},
	language = {en},
	urldate = {2025-01-31},
	publisher = {arXiv},
	author = {Cai, Weilin and Jiang, Juyong and Wang, Fan and Tang, Jing and Kim, Sunghun and Huang, Jiayi},
	month = aug,
	year = {2024},
	note = {arXiv:2407.06204 [cs]},
}

@article{hofman_evolution_2014,
	title = {Evolution of the human brain: when bigger is better},
	volume = {8},
	issn = {1662-5129},
	shorttitle = {Evolution of the human brain},
	url = {http://journal.frontiersin.org/article/10.3389/fnana.2014.00015/abstract},
	doi = {10.3389/fnana.2014.00015},
	language = {en},
	urldate = {2025-01-31},
	journal = {Frontiers in Neuroanatomy},
	author = {Hofman, Michel A.},
	month = mar,
	year = {2014},
}

@article{roe_columnar_2019,
	title = {Columnar connectome: toward a mathematics of brain function},
	volume = {3},
	issn = {2472-1751},
	shorttitle = {Columnar connectome},
	url = {https://direct.mit.edu/netn/article/3/3/779-791/2180},
	doi = {10.1162/netn_a_00088},
	language = {en},
	number = {3},
	urldate = {2025-01-31},
	journal = {Network Neuroscience},
	author = {Roe, Anna Wang},
	month = jan,
	year = {2019},
	pages = {779--791},
}

@article{mountcastle_columnar_1997,
	title = {The columnar organization of the neocortex},
	volume = {120},
	issn = {14602156},
	url = {https://academic.oup.com/brain/article-lookup/doi/10.1093/brain/120.4.701},
	doi = {10.1093/brain/120.4.701},
	language = {en},
	number = {4},
	urldate = {2025-01-31},
	journal = {Brain},
	author = {Mountcastle, V.},
	month = apr,
	year = {1997},
	pages = {701--722},
}

@book{hawkins2021thousand,
	title = {A thousand brains: {A} new theory of intelligence},
	publisher = {Basic Books},
	author = {Hawkins, Jeff},
	year = {2021},
}

@misc{ven_continual_2024,
	title = {Continual {Learning} and {Catastrophic} {Forgetting}},
	url = {http://arxiv.org/abs/2403.05175},
	doi = {10.48550/arXiv.2403.05175},
	language = {en},
	urldate = {2025-01-31},
	publisher = {arXiv},
	author = {Ven, Gido M. van de and Soures, Nicholas and Kudithipudi, Dhireesha},
	month = mar,
	year = {2024},
	note = {arXiv:2403.05175 [cs]},
}

@article{dorkenwald_neuronal_2024,
	title = {Neuronal wiring diagram of an adult brain},
	volume = {634},
	issn = {0028-0836, 1476-4687},
	url = {https://www.nature.com/articles/s41586-024-07558-y},
	doi = {10.1038/s41586-024-07558-y},
	language = {en},
	number = {8032},
	urldate = {2025-01-31},
	journal = {Nature},
	author = {Dorkenwald, Sven and Matsliah, Arie and Sterling, Amy R. and Schlegel, Philipp and Yu, Szi-chieh and McKellar, Claire E. and Lin, Albert and Costa, Marta and Eichler, Katharina and Yin, Yijie and Silversmith, Will and Schneider-Mizell, Casey and Jordan, Chris S. and Brittain, Derrick and Halageri, Akhilesh and Kuehner, Kai and Ogedengbe, Oluwaseun and Morey, Ryan and Gager, Jay and Kruk, Krzysztof and Perlman, Eric and Yang, Runzhe and Deutsch, David and Bland, Doug and Sorek, Marissa and Lu, Ran and Macrina, Thomas and Lee, Kisuk and Bae, J. Alexander and Mu, Shang and Nehoran, Barak and Mitchell, Eric and Popovych, Sergiy and Wu, Jingpeng and Jia, Zhen and Castro, Manuel A. and Kemnitz, Nico and Ih, Dodam and Bates, Alexander Shakeel and Eckstein, Nils and Funke, Jan and Collman, Forrest and Bock, Davi D. and Jefferis, Gregory S. X. E. and Seung, H. Sebastian and Murthy, Mala and {The FlyWire Consortium} and Lenizo, Zairene and Burke, Austin T. and Willie, Kyle Patrick and Serafetinidis, Nikitas and Hadjerol, Nashra and Willie, Ryan and Silverman, Ben and Ocho, John Anthony and Bañez, Joshua and Candilada, Rey Adrian and Kristiansen, Anne and Panes, Nelsie and Yadav, Arti and Tancontian, Remer and Serona, Shirleyjoy and Dolorosa, Jet Ivan and Vinson, Kendrick Joules and Garner, Dustin and Salem, Regine and Dagohoy, Ariel and Skelton, Jaime and Lopez, Mendell and Capdevila, Laia Serratosa and Badalamente, Griffin and Stocks, Thomas and Pandey, Anjali and Akiatan, Darrel Jay and Hebditch, James and David, Celia and Sapkal, Dharini and Monungolh, Shaina Mae and Sane, Varun and Pielago, Mark Lloyd and Albero, Miguel and Laude, Jacquilyn and Dos Santos, Márcia and Vohra, Zeba and Wang, Kaiyu and Gogo, Allien Mae and Kind, Emil and Mandahay, Alvin Josh and Martinez, Chereb and Asis, John David and Nair, Chitra and Patel, Dhwani and Manaytay, Marchan and Tamimi, Imaan F. M. and Lim, Clyde Angelo and Ampo, Philip Lenard and Pantujan, Michelle Darapan and Javier, Alexandre and Bautista, Daril and Rana, Rashmita and Seguido, Jansen and Parmar, Bhargavi and Saguimpa, John Clyde and Moore, Merlin and Pleijzier, Markus William and Larson, Mark and Hsu, Joseph and Joshi, Itisha and Kakadiya, Dhara and Braun, Amalia and Pilapil, Cathy and Gkantia, Marina and Parmar, Kaushik and Vanderbeck, Quinn and Salgarella, Irene and Dunne, Christopher and Munnelly, Eva and Kang, Chan Hyuk and Lörsch, Lena and Lee, Jinmook and Kmecova, Lucia and Sancer, Gizem and Baker, Christa and Joroff, Jenna and Calle, Steven and Patel, Yashvi and Sato, Olivia and Fang, Siqi and Salocot, Janice and Salman, Farzaan and Molina-Obando, Sebastian and Brooks, Paul and Bui, Mai and Lichtenberger, Matthew and Tamboboy, Edward and Molloy, Katie and Santana-Cruz, Alexis E. and Hernandez, Anthony and Yu, Seongbong and Diwan, Arzoo and Patel, Monika and Aiken, Travis R. and Morejohn, Sarah and Koskela, Sanna and Yang, Tansy and Lehmann, Daniel and Chojetzki, Jonas and Sisodiya, Sangeeta and Koolman, Selden and Shiu, Philip K. and Cho, Sky and Bast, Annika and Reicher, Brian and Blanquart, Marlon and Houghton, Lucy and Choi, Hyungjun and Ioannidou, Maria and Collie, Matt and Eckhardt, Joanna and Gorko, Benjamin and Guo, Li and Zheng, Zhihao and Poh, Alisa and Lin, Marina and Taisz, István and Murfin, Wes and Díez, Álvaro Sanz and Reinhard, Nils and Gibb, Peter and Patel, Nidhi and Kumar, Sandeep and Yun, Minsik and Wang, Megan and Jones, Devon and Encarnacion-Rivera, Lucas and Oswald, Annalena and Jadia, Akanksha and Erginkaya, Mert and Drummond, Nik and Walter, Leonie and Tastekin, Ibrahim and Zhong, Xin and Mabuchi, Yuta and Figueroa Santiago, Fernando J. and Verma, Urja and Byrne, Nick and Kunze, Edda and Crahan, Thomas and Margossian, Ryan and Kim, Haein and Georgiev, Iliyan and Szorenyi, Fabianna and Adachi, Atsuko and Bargeron, Benjamin and Stürner, Tomke and Demarest, Damian and Gür, Burak and Becker, Andrea N. and Turnbull, Robert and Morren, Ashley and Sandoval, Andrea and Moreno-Sanchez, Anthony and Pacheco, Diego A. and Samara, Eleni and Croke, Haley and Thomson, Alexander and Laughland, Connor and Dutta, Suchetana B. and De Antón, Paula Guiomar Alarcón and Huang, Binglin and Pujols, Patricia and Haber, Isabel and González-Segarra, Amanda and Choe, Daniel T. and Lukyanova, Veronika and Mancini, Nino and Liu, Zequan and Okubo, Tatsuo and Flynn, Miriam A. and Vitelli, Gianna and Laturney, Meghan and Li, Feng and Cao, Shuo and Manyari-Diaz, Carolina and Yim, Hyunsoo and Duc Le, Anh and Maier, Kate and Yu, Seungyun and Nam, Yeonju and Bąba, Daniel and Abusaif, Amanda and Francis, Audrey and Gayk, Jesse and Huntress, Sommer S. and Barajas, Raquel and Kim, Mindy and Cui, Xinyue and Sterne, Gabriella R. and Li, Anna and Park, Keehyun and Dempsey, Georgia and Mathew, Alan and Kim, Jinseong and Kim, Taewan and Wu, Guan-ting and Dhawan, Serene and Brotas, Margarida and Zhang, Cheng-hao and Bailey, Shanice and Del Toro, Alexander and Yang, Runzhe and Gerhard, Stephan and Champion, Andrew and Anderson, David J. and Behnia, Rudy and Bidaye, Salil S. and Borst, Alexander and Chiappe, Eugenia and Colodner, Kenneth J. and Dacks, Andrew and Dickson, Barry and Garcia, Denise and Hampel, Stefanie and Hartenstein, Volker and Hassan, Bassem and Helfrich-Forster, Charlotte and Huetteroth, Wolf and Kim, Jinseop and Kim, Sung Soo and Kim, Young-Joon and Kwon, Jae Young and Lee, Wei-Chung and Linneweber, Gerit A. and Maimon, Gaby and Mann, Richard and Noselli, Stéphane and Pankratz, Michael and Prieto-Godino, Lucia and Read, Jenny and Reiser, Michael and Von Reyn, Katie and Ribeiro, Carlos and Scott, Kristin and Seeds, Andrew M. and Selcho, Mareike and Silies, Marion and Simpson, Julie and Waddell, Scott and Wernet, Mathias F. and Wilson, Rachel I. and Wolf, Fred W. and Yao, Zepeng and Yapici, Nilay and Zandawala, Meet},
	month = oct,
	year = {2024},
	pages = {124--138},
}

@article{yang_selection_2024,
	title = {Selection of experience for memory by hippocampal sharp wave ripples},
	volume = {383},
	issn = {0036-8075, 1095-9203},
	url = {https://www.science.org/doi/10.1126/science.adk8261},
	doi = {10.1126/science.adk8261},
	language = {en},
	number = {6690},
	urldate = {2025-01-30},
	journal = {Science},
	author = {Yang, Wannan and Sun, Chen and Huszár, Roman and Hainmueller, Thomas and Kiselev, Kirill and Buzsáki, György},
	month = mar,
	year = {2024},
	pages = {1478--1483},
}

@article{buzsaki_neuronal_2004,
	title = {Neuronal {Oscillations} in {Cortical} {Networks}},
	volume = {304},
	issn = {0036-8075, 1095-9203},
	url = {https://www.science.org/doi/10.1126/science.1099745},
	doi = {10.1126/science.1099745},
	language = {en},
	number = {5679},
	urldate = {2025-01-28},
	journal = {Science},
	author = {Buzsáki, György and Draguhn, Andreas},
	month = jun,
	year = {2004},
	pages = {1926--1929},
}

@article{huang_normalization_2023,
	title = {Normalization {Techniques} in {Training} {DNNs}: {Methodology}, {Analysis} and {Application}},
	volume = {45},
	copyright = {https://ieeexplore.ieee.org/Xplorehelp/downloads/license-information/IEEE.html},
	issn = {0162-8828, 2160-9292, 1939-3539},
	shorttitle = {Normalization {Techniques} in {Training} {DNNs}},
	url = {https://ieeexplore.ieee.org/document/10056354/},
	doi = {10.1109/TPAMI.2023.3250241},
	language = {en},
	number = {8},
	urldate = {2025-01-28},
	journal = {IEEE Transactions on Pattern Analysis and Machine Intelligence},
	author = {Huang, Lei and Qin, Jie and Zhou, Yi and Zhu, Fan and Liu, Li and Shao, Ling},
	month = aug,
	year = {2023},
	pages = {10173--10196},
}

@misc{kunc_three_2024,
	title = {Three {Decades} of {Activations}: {A} {Comprehensive} {Survey} of 400 {Activation} {Functions} for {Neural} {Networks}},
	shorttitle = {Three {Decades} of {Activations}},
	url = {http://arxiv.org/abs/2402.09092},
	doi = {10.48550/arXiv.2402.09092},
	language = {en},
	urldate = {2025-01-28},
	publisher = {arXiv},
	author = {Kunc, Vladimír and Kléma, Jiří},
	month = feb,
	year = {2024},
	note = {arXiv:2402.09092 [cs]},
}

@article{timofeev_neocortical_2004,
	title = {Neocortical seizures: initiation, development and cessation},
	volume = {123},
	copyright = {https://www.elsevier.com/tdm/userlicense/1.0/},
	issn = {03064522},
	shorttitle = {Neocortical seizures},
	url = {https://linkinghub.elsevier.com/retrieve/pii/S0306452203006857},
	doi = {10.1016/j.neuroscience.2003.08.051},
	language = {en},
	number = {2},
	urldate = {2025-01-28},
	journal = {Neuroscience},
	author = {Timofeev, I and Steriade, M},
	month = jan,
	year = {2004},
	pages = {299--336},
}

@misc{Chrysafides2024,
	title = {Physiology, resting potential},
	url = {https://www.ncbi.nlm.nih.gov/books/NBK538338/},
	urldate = {2025-01-28},
	publisher = {StatPearls Publishing},
	author = {Chrysafides, Steven and Bordes, Stephen and Sharma, Sandeep},
	year = {2024},
	note = {Place: Treasure Island (FL)
tex.entrytype: electronic},
}

@article{holtmaat_transient_2005,
	title = {Transient and {Persistent} {Dendritic} {Spines} in the {Neocortex} {In} {Vivo}},
	volume = {45},
	copyright = {https://www.elsevier.com/tdm/userlicense/1.0/},
	issn = {08966273},
	url = {https://linkinghub.elsevier.com/retrieve/pii/S0896627305000048},
	doi = {10.1016/j.neuron.2005.01.003},
	language = {en},
	number = {2},
	urldate = {2025-01-28},
	journal = {Neuron},
	author = {Holtmaat, Anthony J.G.D. and Trachtenberg, Joshua T. and Wilbrecht, Linda and Shepherd, Gordon M. and Zhang, Xiaoqun and Knott, Graham W. and Svoboda, Karel},
	month = jan,
	year = {2005},
	pages = {279--291},
}

@article{meunier_modulation_2017,
	title = {Modulation of {Synaptic} {Plasticity} in the {Cortex} {Needs} to {Understand} {All} the {Players}},
	volume = {9},
	issn = {1663-3563},
	url = {http://journal.frontiersin.org/article/10.3389/fnsyn.2017.00002/full},
	doi = {10.3389/fnsyn.2017.00002},
	language = {en},
	urldate = {2025-01-28},
	journal = {Frontiers in Synaptic Neuroscience},
	author = {Meunier, Claire N. J. and Chameau, Pascal and Fossier, Philippe M.},
	month = feb,
	year = {2017},
}

@article{bartol_nanoconnectomic_2015,
	title = {Nanoconnectomic upper bound on the variability of synaptic plasticity},
	volume = {4},
	copyright = {http://creativecommons.org/licenses/by/4.0/},
	issn = {2050-084X},
	url = {https://elifesciences.org/articles/10778},
	doi = {10.7554/eLife.10778},
	language = {en},
	urldate = {2025-01-28},
	journal = {eLife},
	author = {Bartol, Thomas M and Bromer, Cailey and Kinney, Justin and Chirillo, Michael A and Bourne, Jennifer N and Harris, Kristen M and Sejnowski, Terrence J},
	month = nov,
	year = {2015},
	pages = {e10778},
}

@article{plenio_physics_2001,
	title = {The physics of forgetting: {Landauer}'s erasure principle and information theory},
	volume = {42},
	issn = {0010-7514, 1366-5812},
	shorttitle = {The physics of forgetting},
	url = {http://arxiv.org/abs/quant-ph/0103108},
	doi = {10.1080/00107510010018916},
	language = {en},
	number = {1},
	urldate = {2025-01-24},
	journal = {Contemporary Physics},
	author = {Plenio, M. B. and Vitelli, V.},
	month = jan,
	year = {2001},
	note = {arXiv:quant-ph/0103108},
	pages = {25--60},
}

@misc{Green500_2024,
	title = {The {GREEN500} list - november 2024},
	url = {https://top500.org/lists/green500/2024/11/},
	author = {{TOP500}},
	month = nov,
	year = {2024},
}

@misc{epoch2024trainingcomputeoffrontieraimodelsgrowsby45xperyear,
	title = {Training compute of frontier {AI} models grows by 4-5x per year},
	url = {https://epoch.ai/blog/training-compute-of-frontier-ai-models-grows-by-4-5x-per-year},
	author = {Sevilla, Jaime and Roldán, Edu},
	year = {2024},
}

@misc{han_parameter-efficient_2024,
	title = {Parameter-{Efficient} {Fine}-{Tuning} for {Large} {Models}: {A} {Comprehensive} {Survey}},
	shorttitle = {Parameter-{Efficient} {Fine}-{Tuning} for {Large} {Models}},
	url = {http://arxiv.org/abs/2403.14608},
	doi = {10.48550/arXiv.2403.14608},
	language = {en},
	urldate = {2025-01-23},
	publisher = {arXiv},
	author = {Han, Zeyu and Gao, Chao and Liu, Jinyang and Zhang, Jeff and Zhang, Sai Qian},
	month = sep,
	year = {2024},
	note = {arXiv:2403.14608 [cs]},
}

@article{DeepSeek2025,
	title = {{DeepSeek}-{R1}: {Incentivizing} reasoning capability in llms via reinforcement learning},
	journal = {arXiv preprint},
	author = {{DeepSeek-AI}},
	year = {2025},
}

@article{pashler1994dual,
	title = {Dual-task interference in simple tasks: data and theory.},
	volume = {116},
	number = {2},
	journal = {Psychological bulletin},
	author = {Pashler, Harold},
	year = {1994},
	note = {Publisher: American Psychological Association},
	pages = {220},
}

@misc{semianalysis_inference_race_2023,
	title = {Inference race to the bottom: {Make} or break for {AI} startups},
	url = {https://semianalysis.com/2023/12/18/inference-race-to-the-bottom-make/},
	author = {Patel, Dylan and Nishball, Daniel},
	month = dec,
	year = {2023},
	note = {tex.howpublished: SemiAnalysis},
}

@misc{team_large_2024,
	title = {Large {Concept} {Models}: {Language} {Modeling} in a {Sentence} {Representation} {Space}},
	shorttitle = {Large {Concept} {Models}},
	url = {http://arxiv.org/abs/2412.08821},
	doi = {10.48550/arXiv.2412.08821},
	language = {en},
	urldate = {2025-01-21},
	publisher = {arXiv},
	author = {team, L. C. M. and Barrault, Loïc and Duquenne, Paul-Ambroise and Elbayad, Maha and Kozhevnikov, Artyom and Alastruey, Belen and Andrews, Pierre and Coria, Mariano and Couairon, Guillaume and Costa-jussà, Marta R. and Dale, David and Elsahar, Hady and Heffernan, Kevin and Janeiro, João Maria and Tran, Tuan and Ropers, Christophe and Sánchez, Eduardo and Roman, Robin San and Mourachko, Alexandre and Saleem, Safiyyah and Schwenk, Holger},
	month = dec,
	year = {2024},
	note = {arXiv:2412.08821 [cs]},
}

@article{reuters_microsoft_openai_2024,
	title = {Microsoft, {OpenAI} planning \$100 billion data center project - {Information}},
	url = {https://www.reuters.com/technology/microsoft-openai-planning-100-billion-data-center-project-information-reports-2024-03-29/},
	journal = {Reuters},
	author = {{Reuters}},
	month = mar,
	year = {2024},
}

@misc{dubey_llama_2024,
	title = {The {Llama} 3 {Herd} of {Models}},
	url = {http://arxiv.org/abs/2407.21783},
	language = {en},
	urldate = {2024-11-19},
	publisher = {arXiv},
	author = {Dubey, Abhimanyu and Jauhri, Abhinav and Pandey, Abhinav and Kadian, Abhishek and Al-Dahle, Ahmad and Letman, Aiesha and Mathur, Akhil and Schelten, Alan and Yang, Amy and Fan, Angela and Goyal, Anirudh and Hartshorn, Anthony and Yang, Aobo and Mitra, Archi and Sravankumar, Archie and Korenev, Artem and Hinsvark, Arthur and Rao, Arun and Zhang, Aston and Rodriguez, Aurelien and Gregerson, Austen and Spataru, Ava and Roziere, Baptiste and Biron, Bethany and Tang, Binh and Chern, Bobbie and Caucheteux, Charlotte and Nayak, Chaya and Bi, Chloe and Marra, Chris and McConnell, Chris and Keller, Christian and Touret, Christophe and Wu, Chunyang and Wong, Corinne and Ferrer, Cristian Canton and Nikolaidis, Cyrus and Allonsius, Damien and Song, Daniel and Pintz, Danielle and Livshits, Danny and Esiobu, David and Choudhary, Dhruv and Mahajan, Dhruv and Garcia-Olano, Diego and Perino, Diego and Hupkes, Dieuwke and Lakomkin, Egor and AlBadawy, Ehab and Lobanova, Elina and Dinan, Emily and Smith, Eric Michael and Radenovic, Filip and Zhang, Frank and Synnaeve, Gabriel and Lee, Gabrielle and Anderson, Georgia Lewis and Nail, Graeme and Mialon, Gregoire and Pang, Guan and Cucurell, Guillem and Nguyen, Hailey and Korevaar, Hannah and Xu, Hu and Touvron, Hugo and Zarov, Iliyan and Ibarra, Imanol Arrieta and Kloumann, Isabel and Misra, Ishan and Evtimov, Ivan and Copet, Jade and Lee, Jaewon and Geffert, Jan and Vranes, Jana and Park, Jason and Mahadeokar, Jay and Shah, Jeet and Linde, Jelmer van der and Billock, Jennifer and Hong, Jenny and Lee, Jenya and Fu, Jeremy and Chi, Jianfeng and Huang, Jianyu and Liu, Jiawen and Wang, Jie and Yu, Jiecao and Bitton, Joanna and Spisak, Joe and Park, Jongsoo and Rocca, Joseph and Johnstun, Joshua and Saxe, Joshua and Jia, Junteng and Alwala, Kalyan Vasuden and Upasani, Kartikeya and Plawiak, Kate and Li, Ke and Heafield, Kenneth and Stone, Kevin and El-Arini, Khalid and Iyer, Krithika and Malik, Kshitiz and Chiu, Kuenley and Bhalla, Kunal and Rantala-Yeary, Lauren and Maaten, Laurens van der and Chen, Lawrence and Tan, Liang and Jenkins, Liz and Martin, Louis and Madaan, Lovish and Malo, Lubo and Blecher, Lukas and Landzaat, Lukas and Oliveira, Luke de and Muzzi, Madeline and Pasupuleti, Mahesh and Singh, Mannat and Paluri, Manohar and Kardas, Marcin and Oldham, Mathew and Rita, Mathieu and Pavlova, Maya and Kambadur, Melanie and Lewis, Mike and Si, Min and Singh, Mitesh Kumar and Hassan, Mona and Goyal, Naman and Torabi, Narjes and Bashlykov, Nikolay and Bogoychev, Nikolay and Chatterji, Niladri and Duchenne, Olivier and Çelebi, Onur and Alrassy, Patrick and Zhang, Pengchuan and Li, Pengwei and Vasic, Petar and Weng, Peter and Bhargava, Prajjwal and Dubal, Pratik and Krishnan, Praveen and Koura, Punit Singh and Xu, Puxin and He, Qing and Dong, Qingxiao and Srinivasan, Ragavan and Ganapathy, Raj and Calderer, Ramon and Cabral, Ricardo Silveira and Stojnic, Robert and Raileanu, Roberta and Girdhar, Rohit and Patel, Rohit and Sauvestre, Romain and Polidoro, Ronnie and Sumbaly, Roshan and Taylor, Ross and Silva, Ruan and Hou, Rui and Wang, Rui and Hosseini, Saghar and Chennabasappa, Sahana and Singh, Sanjay and Bell, Sean and Kim, Seohyun Sonia and Edunov, Sergey and Nie, Shaoliang and Narang, Sharan and Raparthy, Sharath and Shen, Sheng and Wan, Shengye and Bhosale, Shruti and Zhang, Shun and Vandenhende, Simon and Batra, Soumya and Whitman, Spencer and Sootla, Sten and Collot, Stephane and Gururangan, Suchin and Borodinsky, Sydney and Herman, Tamar and Fowler, Tara and Sheasha, Tarek and Georgiou, Thomas and Scialom, Thomas and Speckbacher, Tobias and Mihaylov, Todor and Xiao, Tong and Karn, Ujjwal and Goswami, Vedanuj and Gupta, Vibhor and Ramanathan, Vignesh and Kerkez, Viktor and Gonguet, Vincent and Do, Virginie and Vogeti, Vish and Petrovic, Vladan and Chu, Weiwei and Xiong, Wenhan and Fu, Wenyin and Meers, Whitney and Martinet, Xavier and Wang, Xiaodong and Tan, Xiaoqing Ellen and Xie, Xinfeng and Jia, Xuchao and Wang, Xuewei and Goldschlag, Yaelle and Gaur, Yashesh and Babaei, Yasmine and Wen, Yi and Song, Yiwen and Zhang, Yuchen and Li, Yue and Mao, Yuning and Coudert, Zacharie Delpierre and Yan, Zheng and Chen, Zhengxing and Papakipos, Zoe and Singh, Aaditya and Grattafiori, Aaron and Jain, Abha and Kelsey, Adam and Shajnfeld, Adam and Gangidi, Adithya and Victoria, Adolfo and Goldstand, Ahuva and Menon, Ajay and Sharma, Ajay and Boesenberg, Alex and Vaughan, Alex and Baevski, Alexei and Feinstein, Allie and Kallet, Amanda and Sangani, Amit and Yunus, Anam and Lupu, Andrei and Alvarado, Andres and Caples, Andrew and Gu, Andrew and Ho, Andrew and Poulton, Andrew and Ryan, Andrew and Ramchandani, Ankit and Franco, Annie and Saraf, Aparajita and Chowdhury, Arkabandhu and Gabriel, Ashley and Bharambe, Ashwin and Eisenman, Assaf and Yazdan, Azadeh and James, Beau and Maurer, Ben and Leonhardi, Benjamin and Huang, Bernie and Loyd, Beth and Paola, Beto De and Paranjape, Bhargavi and Liu, Bing and Wu, Bo and Ni, Boyu and Hancock, Braden and Wasti, Bram and Spence, Brandon and Stojkovic, Brani and Gamido, Brian and Montalvo, Britt and Parker, Carl and Burton, Carly and Mejia, Catalina and Wang, Changhan and Kim, Changkyu and Zhou, Chao and Hu, Chester and Chu, Ching-Hsiang and Cai, Chris and Tindal, Chris and Feichtenhofer, Christoph and Civin, Damon and Beaty, Dana and Kreymer, Daniel and Li, Daniel and Wyatt, Danny and Adkins, David and Xu, David and Testuggine, Davide and David, Delia and Parikh, Devi and Liskovich, Diana and Foss, Didem and Wang, Dingkang and Le, Duc and Holland, Dustin and Dowling, Edward and Jamil, Eissa and Montgomery, Elaine and Presani, Eleonora and Hahn, Emily and Wood, Emily and Brinkman, Erik and Arcaute, Esteban and Dunbar, Evan and Smothers, Evan and Sun, Fei and Kreuk, Felix and Tian, Feng and Ozgenel, Firat and Caggioni, Francesco and Guzmán, Francisco and Kanayet, Frank and Seide, Frank and Florez, Gabriela Medina and Schwarz, Gabriella and Badeer, Gada and Swee, Georgia and Halpern, Gil and Thattai, Govind and Herman, Grant and Sizov, Grigory and Guangyi and Zhang and Lakshminarayanan, Guna and Shojanazeri, Hamid and Zou, Han and Wang, Hannah and Zha, Hanwen and Habeeb, Haroun and Rudolph, Harrison and Suk, Helen and Aspegren, Henry and Goldman, Hunter and Damlaj, Ibrahim and Molybog, Igor and Tufanov, Igor and Veliche, Irina-Elena and Gat, Itai and Weissman, Jake and Geboski, James and Kohli, James and Asher, Japhet and Gaya, Jean-Baptiste and Marcus, Jeff and Tang, Jeff and Chan, Jennifer and Zhen, Jenny and Reizenstein, Jeremy and Teboul, Jeremy and Zhong, Jessica and Jin, Jian and Yang, Jingyi and Cummings, Joe and Carvill, Jon and Shepard, Jon and McPhie, Jonathan and Torres, Jonathan and Ginsburg, Josh and Wang, Junjie and Wu, Kai and U, Kam Hou and Saxena, Karan and Prasad, Karthik and Khandelwal, Kartikay and Zand, Katayoun and Matosich, Kathy and Veeraraghavan, Kaushik and Michelena, Kelly and Li, Keqian and Huang, Kun and Chawla, Kunal and Lakhotia, Kushal and Huang, Kyle and Chen, Lailin and Garg, Lakshya and A, Lavender and Silva, Leandro and Bell, Lee and Zhang, Lei and Guo, Liangpeng and Yu, Licheng and Moshkovich, Liron and Wehrstedt, Luca and Khabsa, Madian and Avalani, Manav and Bhatt, Manish and Tsimpoukelli, Maria and Mankus, Martynas and Hasson, Matan and Lennie, Matthew and Reso, Matthias and Groshev, Maxim and Naumov, Maxim and Lathi, Maya and Keneally, Meghan and Seltzer, Michael L. and Valko, Michal and Restrepo, Michelle and Patel, Mihir and Vyatskov, Mik and Samvelyan, Mikayel and Clark, Mike and Macey, Mike and Wang, Mike and Hermoso, Miquel Jubert and Metanat, Mo and Rastegari, Mohammad and Bansal, Munish and Santhanam, Nandhini and Parks, Natascha and White, Natasha and Bawa, Navyata and Singhal, Nayan and Egebo, Nick and Usunier, Nicolas and Laptev, Nikolay Pavlovich and Dong, Ning and Zhang, Ning and Cheng, Norman and Chernoguz, Oleg and Hart, Olivia and Salpekar, Omkar and Kalinli, Ozlem and Kent, Parkin and Parekh, Parth and Saab, Paul and Balaji, Pavan and Rittner, Pedro and Bontrager, Philip and Roux, Pierre and Dollar, Piotr and Zvyagina, Polina and Ratanchandani, Prashant and Yuvraj, Pritish and Liang, Qian and Alao, Rachad and Rodriguez, Rachel and Ayub, Rafi and Murthy, Raghotham and Nayani, Raghu and Mitra, Rahul and Li, Raymond and Hogan, Rebekkah and Battey, Robin and Wang, Rocky and Maheswari, Rohan and Howes, Russ and Rinott, Ruty and Bondu, Sai Jayesh and Datta, Samyak and Chugh, Sara and Hunt, Sara and Dhillon, Sargun and Sidorov, Sasha and Pan, Satadru and Verma, Saurabh and Yamamoto, Seiji and Ramaswamy, Sharadh and Lindsay, Shaun and Lindsay, Shaun and Feng, Sheng and Lin, Shenghao and Zha, Shengxin Cindy and Shankar, Shiva and Zhang, Shuqiang and Zhang, Shuqiang and Wang, Sinong and Agarwal, Sneha and Sajuyigbe, Soji and Chintala, Soumith and Max, Stephanie and Chen, Stephen and Kehoe, Steve and Satterfield, Steve and Govindaprasad, Sudarshan and Gupta, Sumit and Cho, Sungmin and Virk, Sunny and Subramanian, Suraj and Choudhury, Sy and Goldman, Sydney and Remez, Tal and Glaser, Tamar and Best, Tamara and Kohler, Thilo and Robinson, Thomas and Li, Tianhe and Zhang, Tianjun and Matthews, Tim and Chou, Timothy and Shaked, Tzook and Vontimitta, Varun and Ajayi, Victoria and Montanez, Victoria and Mohan, Vijai and Kumar, Vinay Satish and Mangla, Vishal and Albiero, Vítor and Ionescu, Vlad and Poenaru, Vlad and Mihailescu, Vlad Tiberiu and Ivanov, Vladimir and Li, Wei and Wang, Wenchen and Jiang, Wenwen and Bouaziz, Wes and Constable, Will and Tang, Xiaocheng and Wang, Xiaofang and Wu, Xiaojian and Wang, Xiaolan and Xia, Xide and Wu, Xilun and Gao, Xinbo and Chen, Yanjun and Hu, Ye and Jia, Ye and Qi, Ye and Li, Yenda and Zhang, Yilin and Zhang, Ying and Adi, Yossi and Nam, Youngjin and Yu and Wang and Hao, Yuchen and Qian, Yundi and He, Yuzi and Rait, Zach and DeVito, Zachary and Rosnbrick, Zef and Wen, Zhaoduo and Yang, Zhenyu and Zhao, Zhiwei},
	month = aug,
	year = {2024},
	note = {arXiv:2407.21783 [cs]},
}

@article{ren_comprehensive_2022,
	title = {A {Comprehensive} {Survey} of {Neural} {Architecture} {Search}: {Challenges} and {Solutions}},
	volume = {54},
	issn = {0360-0300, 1557-7341},
	shorttitle = {A {Comprehensive} {Survey} of {Neural} {Architecture} {Search}},
	url = {https://dl.acm.org/doi/10.1145/3447582},
	doi = {10.1145/3447582},
	language = {en},
	number = {4},
	urldate = {2024-11-19},
	journal = {ACM Computing Surveys},
	author = {Ren, Pengzhen and Xiao, Yun and Chang, Xiaojun and Huang, Po-yao and Li, Zhihui and Chen, Xiaojiang and Wang, Xin},
	month = may,
	year = {2022},
	pages = {1--34},
}

@inproceedings{han_lm-infinite_2024,
	address = {Mexico City, Mexico},
	title = {{LM}-{Infinite}: {Zero}-{Shot} {Extreme} {Length} {Generalization} for {Large} {Language} {Models}},
	shorttitle = {{LM}-{Infinite}},
	url = {https://aclanthology.org/2024.naacl-long.222},
	doi = {10.18653/v1/2024.naacl-long.222},
	language = {en},
	urldate = {2024-11-18},
	booktitle = {Proceedings of the 2024 {Conference} of the {North} {American} {Chapter} of the {Association} for {Computational} {Linguistics}: {Human} {Language} {Technologies} ({Volume} 1: {Long} {Papers})},
	publisher = {Association for Computational Linguistics},
	author = {Han, Chi and Wang, Qifan and Peng, Hao and Xiong, Wenhan and Chen, Yu and Ji, Heng and Wang, Sinong},
	year = {2024},
	pages = {3991--4008},
}

@misc{bai_training_2022,
	title = {Training a {Helpful} and {Harmless} {Assistant} with {Reinforcement} {Learning} from {Human} {Feedback}},
	url = {http://arxiv.org/abs/2204.05862},
	language = {en},
	urldate = {2024-11-18},
	publisher = {arXiv},
	author = {Bai, Yuntao and Jones, Andy and Ndousse, Kamal and Askell, Amanda and Chen, Anna and DasSarma, Nova and Drain, Dawn and Fort, Stanislav and Ganguli, Deep and Henighan, Tom and Joseph, Nicholas and Kadavath, Saurav and Kernion, Jackson and Conerly, Tom and El-Showk, Sheer and Elhage, Nelson and Hatfield-Dodds, Zac and Hernandez, Danny and Hume, Tristan and Johnston, Scott and Kravec, Shauna and Lovitt, Liane and Nanda, Neel and Olsson, Catherine and Amodei, Dario and Brown, Tom and Clark, Jack and McCandlish, Sam and Olah, Chris and Mann, Ben and Kaplan, Jared},
	month = apr,
	year = {2022},
	note = {arXiv:2204.05862 [cs]},
}

@misc{wu_jailbreaking_2024,
	title = {Jailbreaking {GPT}-{4V} via {Self}-{Adversarial} {Attacks} with {System} {Prompts}},
	url = {http://arxiv.org/abs/2311.09127},
	language = {en},
	urldate = {2024-11-18},
	publisher = {arXiv},
	author = {Wu, Yuanwei and Li, Xiang and Liu, Yixin and Zhou, Pan and Sun, Lichao},
	month = jan,
	year = {2024},
	note = {arXiv:2311.09127 [cs]},
}

@misc{patel2024multidatacenter,
	title = {Multi-datacenter training: {OpenAI}'s ambitious plan to beat {Google}'s infrastructure},
	url = {https://semianalysis.com/2024/09/04/multi-datacenter-training-openais/},
	publisher = {SemiAnalysis},
	author = {Patel, Dylan and Ontiveros, Jeremie Eliahou and Nishball, Daniel},
	month = sep,
	year = {2024},
}

@misc{villalobos_will_2024,
	title = {Will we run out of data? {Limits} of {LLM} scaling based on human-generated data},
	shorttitle = {Will we run out of data?},
	url = {http://arxiv.org/abs/2211.04325},
	language = {en},
	urldate = {2024-11-15},
	publisher = {arXiv},
	author = {Villalobos, Pablo and Ho, Anson and Sevilla, Jaime and Besiroglu, Tamay and Heim, Lennart and Hobbhahn, Marius},
	month = jun,
	year = {2024},
	note = {arXiv:2211.04325 [cs]},
}

@misc{wang2019bfloat16,
	title = {{BFloat16}: {The} secret to high performance on {Cloud} {TPUs}},
	url = {https://cloud.google.com/blog/products/ai-machine-learning/bfloat16-the-secret-to-high-performance-on-cloud-tpus},
	publisher = {Google Cloud Blog},
	author = {Wang, Shibo and Kanwar, Pankaj},
	year = {2019},
}

@article{smart_regulation_2000,
	title = {Regulation of dendritic spine stability},
	volume = {10},
	copyright = {http://doi.wiley.com/10.1002/tdm\_license\_1.1},
	issn = {1050-9631, 1098-1063},
	url = {https://onlinelibrary.wiley.com/doi/10.1002/1098-1063(2000)10:5<542::AID-HIPO4>3.0.CO;2-7},
	doi = {10.1002/1098-1063(2000)10:5<542::AID-HIPO4>3.0.CO;2-7},
	language = {en},
	number = {5},
	urldate = {2024-11-14},
	journal = {Hippocampus},
	author = {Smart, Fiona M. and Halpain, Shelley},
	year = {2000},
	pages = {542--554},
}

@article{squire_legacy_2009,
	title = {The {Legacy} of {Patient} {H}.{M}. for {Neuroscience}},
	volume = {61},
	issn = {08966273},
	url = {https://linkinghub.elsevier.com/retrieve/pii/S0896627308010957},
	doi = {10.1016/j.neuron.2008.12.023},
	language = {en},
	number = {1},
	urldate = {2024-11-13},
	journal = {Neuron},
	author = {Squire, Larry R.},
	month = jan,
	year = {2009},
	pages = {6--9},
}

@article{walker_role_2009,
	title = {The {Role} of {Slow} {Wave} {Sleep} in {Memory} {Processing}},
	volume = {5},
	issn = {1550-9389, 1550-9397},
	url = {http://jcsm.aasm.org/doi/10.5664/jcsm.5.2S.S20},
	doi = {10.5664/jcsm.5.2S.S20},
	language = {en},
	number = {2 suppl},
	urldate = {2024-11-13},
	journal = {Journal of Clinical Sleep Medicine},
	author = {Walker, Matthew P.},
	month = apr,
	year = {2009},
}

@article{norimoto_claustrum_2020,
	title = {A claustrum in reptiles and its role in slow-wave sleep},
	volume = {578},
	issn = {0028-0836, 1476-4687},
	url = {https://www.nature.com/articles/s41586-020-1993-6},
	doi = {10.1038/s41586-020-1993-6},
	language = {en},
	number = {7795},
	urldate = {2024-11-13},
	journal = {Nature},
	author = {Norimoto, Hiroaki and Fenk, Lorenz A. and Li, Hsing-Hsi and Tosches, Maria Antonietta and Gallego-Flores, Tatiana and Hain, David and Reiter, Sam and Kobayashi, Riho and Macias, Angeles and Arends, Anja and Klinkmann, Michaela and Laurent, Gilles},
	month = feb,
	year = {2020},
	pages = {413--418},
}

@article{miyazaki_sleep_2017,
	title = {Sleep in vertebrate and invertebrate animals, and insights into the function and evolution of sleep},
	volume = {118},
	issn = {01680102},
	url = {https://linkinghub.elsevier.com/retrieve/pii/S0168010217302225},
	doi = {10.1016/j.neures.2017.04.017},
	language = {en},
	urldate = {2024-11-13},
	journal = {Neuroscience Research},
	author = {Miyazaki, Shinichi and Liu, Chih-Yao and Hayashi, Yu},
	month = may,
	year = {2017},
	pages = {3--12},
}

@article{klukas_efficient_2020,
	title = {Efficient and flexible representation of higher-dimensional cognitive variables with grid cells},
	volume = {16},
	issn = {1553-7358},
	url = {https://dx.plos.org/10.1371/journal.pcbi.1007796},
	doi = {10.1371/journal.pcbi.1007796},
	language = {en},
	number = {4},
	urldate = {2024-10-31},
	journal = {PLOS Computational Biology},
	author = {Klukas, Mirko and Lewis, Marcus and Fiete, Ila},
	editor = {Bush, Daniel},
	month = apr,
	year = {2020},
	pages = {e1007796},
}

@misc{schoyen_hexagons_2024,
	title = {Hexagons all the way down: {Grid} cells as a conformal isometric map of space},
	shorttitle = {Hexagons all the way down},
	url = {http://biorxiv.org/lookup/doi/10.1101/2024.02.02.578585},
	doi = {10.1101/2024.02.02.578585},
	language = {en},
	urldate = {2024-10-31},
	author = {Schøyen, Vemund and Bechkov, Constantin and Pettersen, Markus Borud and Hermansen, Erik and Holzhausen, Konstantin and Malthe-Sørenssen, Anders and Fyhn, Marianne and Lepperød, Mikkel Elle},
	month = feb,
	year = {2024},
}

@misc{keller_spacetime_2024,
	title = {A {Spacetime} {Perspective} on {Dynamical} {Computation} in {Neural} {Information} {Processing} {Systems}},
	url = {http://arxiv.org/abs/2409.13669},
	language = {en},
	urldate = {2024-10-22},
	publisher = {arXiv},
	author = {Keller, T. Anderson and Muller, Lyle and Sejnowski, Terrence J. and Welling, Max},
	month = sep,
	year = {2024},
	note = {arXiv:2409.13669 [q-bio]},
}

@misc{augustine_survey_2024,
	title = {A {Survey} on {Universal} {Approximation} {Theorems}},
	url = {http://arxiv.org/abs/2407.12895},
	language = {en},
	urldate = {2024-10-04},
	publisher = {arXiv},
	author = {Augustine, Midhun T.},
	month = jul,
	year = {2024},
	note = {arXiv:2407.12895 [cs, eess]},
}

@misc{ororbia_brain-inspired_2023,
	title = {Brain-{Inspired} {Machine} {Intelligence}: {A} {Survey} of {Neurobiologically}-{Plausible} {Credit} {Assignment}},
	shorttitle = {Brain-{Inspired} {Machine} {Intelligence}},
	url = {http://arxiv.org/abs/2312.09257},
	language = {en},
	urldate = {2024-09-30},
	publisher = {arXiv},
	author = {Ororbia, Alexander G.},
	month = dec,
	year = {2023},
	note = {arXiv:2312.09257 [cs, q-bio]},
}

@article{herculano-houzel_human_2009,
	title = {The human brain in numbers: a linearly scaled-up primate brain},
	volume = {3},
	issn = {16625161},
	shorttitle = {The human brain in numbers},
	url = {http://journal.frontiersin.org/article/10.3389/neuro.09.031.2009/abstract},
	doi = {10.3389/neuro.09.031.2009},
	language = {en},
	urldate = {2024-09-30},
	journal = {Frontiers in Human Neuroscience},
	author = {Herculano-Houzel, Suzana},
	year = {2009},
}

@article{christensen_2022_2022,
	title = {2022 roadmap on neuromorphic computing and engineering},
	volume = {2},
	issn = {2634-4386},
	url = {https://iopscience.iop.org/article/10.1088/2634-4386/ac4a83},
	doi = {10.1088/2634-4386/ac4a83},
	language = {en},
	number = {2},
	urldate = {2024-09-30},
	journal = {Neuromorphic Computing and Engineering},
	author = {Christensen, Dennis V and Dittmann, Regina and Linares-Barranco, Bernabe and Sebastian, Abu and Le Gallo, Manuel and Redaelli, Andrea and Slesazeck, Stefan and Mikolajick, Thomas and Spiga, Sabina and Menzel, Stephan and Valov, Ilia and Milano, Gianluca and Ricciardi, Carlo and Liang, Shi-Jun and Miao, Feng and Lanza, Mario and Quill, Tyler J and Keene, Scott T and Salleo, Alberto and Grollier, Julie and Marković, Danijela and Mizrahi, Alice and Yao, Peng and Yang, J Joshua and Indiveri, Giacomo and Strachan, John Paul and Datta, Suman and Vianello, Elisa and Valentian, Alexandre and Feldmann, Johannes and Li, Xuan and Pernice, Wolfram H P and Bhaskaran, Harish and Furber, Steve and Neftci, Emre and Scherr, Franz and Maass, Wolfgang and Ramaswamy, Srikanth and Tapson, Jonathan and Panda, Priyadarshini and Kim, Youngeun and Tanaka, Gouhei and Thorpe, Simon and Bartolozzi, Chiara and Cleland, Thomas A and Posch, Christoph and Liu, ShihChii and Panuccio, Gabriella and Mahmud, Mufti and Mazumder, Arnab Neelim and Hosseini, Morteza and Mohsenin, Tinoosh and Donati, Elisa and Tolu, Silvia and Galeazzi, Roberto and Christensen, Martin Ejsing and Holm, Sune and Ielmini, Daniele and Pryds, N},
	month = jun,
	year = {2022},
	pages = {022501},
}

@article{juusola_coding_2007,
	title = {Coding with spike shapes and graded potentials in cortical networks},
	volume = {29},
	issn = {0265-9247, 1521-1878},
	url = {https://onlinelibrary.wiley.com/doi/10.1002/bies.20532},
	doi = {10.1002/bies.20532},
	language = {en},
	number = {2},
	urldate = {2024-09-30},
	journal = {BioEssays},
	author = {Juusola, Mikko and Robinson, Hugh P.C. and De Polavieja, Gonzalo G.},
	month = feb,
	year = {2007},
	pages = {178--187},
}

@article{doya_metalearning_2002,
	title = {Metalearning and neuromodulation},
	volume = {15},
	copyright = {https://www.elsevier.com/tdm/userlicense/1.0/},
	issn = {08936080},
	url = {https://linkinghub.elsevier.com/retrieve/pii/S0893608002000448},
	doi = {10.1016/S0893-6080(02)00044-8},
	language = {en},
	number = {4-6},
	urldate = {2024-09-30},
	journal = {Neural Networks},
	author = {Doya, Kenji},
	month = jun,
	year = {2002},
	pages = {495--506},
}

@misc{reuther_lincoln_2023,
	title = {Lincoln {AI} {Computing} {Survey} ({LAICS}) {Update}},
	url = {http://arxiv.org/abs/2310.09145},
	language = {en},
	urldate = {2024-08-21},
	publisher = {arXiv},
	author = {Reuther, Albert and Michaleas, Peter and Jones, Michael and Gadepally, Vijay and Samsi, Siddharth and Kepner, Jeremy},
	month = oct,
	year = {2023},
	note = {arXiv:2310.09145 [cs]},
}

@inproceedings{dally_hardware_2023,
	title = {Hardware for {Deep} {Learning}},
	booktitle = {2023 {IEEE} {Hot} {Chips} 35 {Symposium} ({HCS})},
	publisher = {IEEE Computer Society},
	author = {Dally, Bill},
	year = {2023},
	pages = {1--58},
}

@article{chow_intermediate_2013,
	title = {Intermediate {Representation}: {The} increasing significance of intermediate representations in compilers},
	volume = {11},
	issn = {1542-7730, 1542-7749},
	shorttitle = {Intermediate {Representation}},
	url = {https://dl.acm.org/doi/10.1145/2542661.2544374},
	doi = {10.1145/2542661.2544374},
	language = {en},
	number = {10},
	urldate = {2024-05-06},
	journal = {Queue},
	author = {Chow, Fred},
	month = oct,
	year = {2013},
	pages = {30--37},
}

@article{block_perceptron_1962,
	title = {The {Perceptron}: {A} {Model} for {Brain} {Functioning}. {I}},
	volume = {34},
	issn = {0034-6861},
	shorttitle = {The {Perceptron}},
	url = {https://link.aps.org/doi/10.1103/RevModPhys.34.123},
	doi = {10.1103/RevModPhys.34.123},
	language = {en},
	number = {1},
	urldate = {2024-03-12},
	journal = {Reviews of Modern Physics},
	author = {Block, H. D.},
	month = jan,
	year = {1962},
	pages = {123--135},
}

@article{rosenblatt_perceptron_1958,
	title = {The perceptron: a probabilistic model for information storage and organization in the brain.},
	volume = {65},
	number = {6},
	journal = {Psychological review},
	author = {Rosenblatt, Frank},
	year = {1958},
	note = {Publisher: American Psychological Association},
	pages = {386},
}

@article{buzsaki_brain_2023,
	title = {Brain rhythms have come of age},
	volume = {111},
	issn = {08966273},
	url = {https://linkinghub.elsevier.com/retrieve/pii/S0896627323002143},
	doi = {10.1016/j.neuron.2023.03.018},
	language = {en},
	number = {7},
	urldate = {2024-03-12},
	journal = {Neuron},
	author = {Buzsáki, György and Vöröslakos, Mihály},
	month = apr,
	year = {2023},
	pages = {922--926},
}

@article{balasubramanian_brain_2021,
	title = {Brain power},
	volume = {118},
	issn = {1091-6490},
	doi = {10.1073/pnas.2107022118},
	language = {eng},
	number = {32},
	journal = {Proceedings of the National Academy of Sciences of the United States of America},
	author = {Balasubramanian, Vijay},
	month = aug,
	year = {2021},
	pmid = {34341108},
	pmcid = {PMC8364152},
	pages = {e2107022118},
}

@inproceedings{pennington_glove_2014,
	address = {Doha, Qatar},
	title = {Glove: {Global} {Vectors} for {Word} {Representation}},
	shorttitle = {Glove},
	url = {http://aclweb.org/anthology/D14-1162},
	doi = {10.3115/v1/D14-1162},
	language = {en},
	urldate = {2024-02-28},
	booktitle = {Proceedings of the 2014 {Conference} on {Empirical} {Methods} in {Natural} {Language} {Processing} ({EMNLP})},
	publisher = {Association for Computational Linguistics},
	author = {Pennington, Jeffrey and Socher, Richard and Manning, Christopher},
	year = {2014},
	pages = {1532--1543},
}

@misc{wu_googles_2016,
	title = {Google's {Neural} {Machine} {Translation} {System}: {Bridging} the {Gap} between {Human} and {Machine} {Translation}},
	shorttitle = {Google's {Neural} {Machine} {Translation} {System}},
	url = {http://arxiv.org/abs/1609.08144},
	urldate = {2024-02-28},
	publisher = {arXiv},
	author = {Wu, Yonghui and Schuster, Mike and Chen, Zhifeng and Le, Quoc V. and Norouzi, Mohammad and Macherey, Wolfgang and Krikun, Maxim and Cao, Yuan and Gao, Qin and Macherey, Klaus and Klingner, Jeff and Shah, Apurva and Johnson, Melvin and Liu, Xiaobing and Kaiser, Łukasz and Gouws, Stephan and Kato, Yoshikiyo and Kudo, Taku and Kazawa, Hideto and Stevens, Keith and Kurian, George and Patil, Nishant and Wang, Wei and Young, Cliff and Smith, Jason and Riesa, Jason and Rudnick, Alex and Vinyals, Oriol and Corrado, Greg and Hughes, Macduff and Dean, Jeffrey},
	month = oct,
	year = {2016},
	note = {arXiv:1609.08144 [cs]},
}

@inproceedings{yuan_towards_2006,
	title = {Towards an integrated understanding of speaking rate in conversation},
	url = {https://www.isca-archive.org/interspeech_2006/yuan06_interspeech.html},
	doi = {10.21437/Interspeech.2006-204},
	language = {en},
	urldate = {2024-02-28},
	booktitle = {Interspeech 2006},
	publisher = {ISCA},
	author = {Yuan, Jiahong and Liberman, Mark and Cieri, Christopher},
	month = sep,
	year = {2006},
	pages = {paper 1795--Mon3A3O.1--0},
}

@article{hiller_microsoft_2023,
	title = {Microsoft {Targets} {Nuclear} to {Power} {AI} {Operations}},
	url = {https://www.wsj.com/tech/ai/microsoft-targets-nuclear-to-power-ai-operations-e10ff798},
	journal = {Wall Street Journal},
	author = {Hiller, Jennifer},
	month = dec,
	year = {2023},
}

@article{economist_staff_data_2024,
	title = {Data centres improved greatly in energy efficiency as they grew massively larger},
	url = {https://www.economist.com/technology-quarterly/2024/01/29/data-centres-improved-greatly-in-energy-efficiency-as-they-grew-massively-larger},
	journal = {The Economist},
	author = {Economist Staff},
	month = feb,
	year = {2024},
}

@inproceedings{prabhakar_sambanova_2022,
	title = {{SambaNova} {SN10} {RDU}: {A} 7nm {Dataflow} {Architecture} to {Accelerate} {Software} 2.0},
	volume = {65},
	doi = {10.1109/ISSCC42614.2022.9731612},
	booktitle = {2022 {IEEE} {International} {Solid}-{State} {Circuits} {Conference} ({ISSCC})},
	author = {Prabhakar, Raghu and Jairath, Sumti and Shin, Jinuk Luke},
	year = {2022},
	pages = {350--352},
}

@inproceedings{abts_groq_2022,
	title = {The {Groq} {Software}-defined {Scale}-out {Tensor} {Streaming} {Multiprocessor} : {From} chips-to-systems architectural overview},
	doi = {10.1109/HCS55958.2022.9895630},
	booktitle = {2022 {IEEE} {Hot} {Chips} 34 {Symposium} ({HCS})},
	author = {Abts, Dennis and Kim, John and Kimmell, Garrin and Boyd, Matthew and Kang, Kris and Parmar, Sahil and Ling, Andrew and Bitar, Andrew and Ahmed, Ibrahim and Ross, Jonathan},
	year = {2022},
	pages = {1--69},
}

@article{fitch_nvidia_2024,
	title = {Nvidia {Hits} \$2 {Trillion} {Valuation} on {Insatiable} {AI} {Chip} {Demand}},
	url = {https://www.wsj.com/tech/ai/nvidia-stock-market-cap-2-trillion-b1c839c8?mod=livecoverage_web},
	language = {English},
	urldate = {2024-02-26},
	journal = {Wall Street Journal},
	author = {Fitch, Asa},
	month = feb,
	year = {2024},
}

@article{brooks_video_2024,
	title = {Video generation models as world simulators},
	url = {https://openai.com/research/video-generation-models-as-world-simulators},
	author = {Brooks, Tim and Peebles, Bill and Homes, Connor and DePue, Will and Guo, Yufei and Jing, Li and Schnurr, David and Taylor, Joe and Luhman, Troy and Luhman, Eric and Ng, Clarence and Wang, Ricky and Ramesh, Aditya},
	year = {2024},
}

@article{katz_gpt-4_2023,
	title = {{GPT}-4 {Passes} the {Bar} {Exam}},
	issn = {1556-5068},
	url = {https://www.ssrn.com/abstract=4389233},
	doi = {10.2139/ssrn.4389233},
	language = {en},
	urldate = {2024-02-26},
	journal = {SSRN Electronic Journal},
	author = {Katz, Daniel Martin and Bommarito, Michael James and Gao, Shang and Arredondo, Pablo},
	year = {2023},
}

@misc{grace_thousands_2024,
	title = {Thousands of {AI} {Authors} on the {Future} of {AI}},
	url = {http://arxiv.org/abs/2401.02843},
	language = {en},
	urldate = {2024-02-26},
	publisher = {arXiv},
	author = {Grace, Katja and Stewart, Harlan and Sandkühler, Julia Fabienne and Thomas, Stephen and Weinstein-Raun, Ben and Brauner, Jan},
	month = jan,
	year = {2024},
	note = {arXiv:2401.02843 [cs]},
}

@article{smith_ai_2024,
	title = {{AI} {Is} {Starting} to {Threaten} {White}-{Collar} {Jobs}. {Few} {Industries} {Are} {Immune}.},
	url = {https://www.wsj.com/lifestyle/careers/ai-is-starting-to-threaten-white-collar-jobs-few-industries-are-immune-9cdbcb90?mod=hp_lead_pos10},
	language = {en},
	urldate = {2024-02-12},
	journal = {Wall Street Journal},
	author = {Smith, Ray A},
	month = feb,
	year = {2024},
}

@misc{olin-ammentorp_hyperdimensional_2023,
	title = {Hyperdimensional {Computing} {Provides} a {Programming} {Paradigm} for {Oscillatory} {Systems}},
	url = {http://arxiv.org/abs/2312.11783},
	language = {en},
	urldate = {2024-02-09},
	publisher = {arXiv},
	author = {Olin-Ammentorp, Wilkie},
	month = dec,
	year = {2023},
	note = {arXiv:2312.11783 [math]},
}

@misc{hendrycks_measuring_2021,
	title = {Measuring {Massive} {Multitask} {Language} {Understanding}},
	url = {http://arxiv.org/abs/2009.03300},
	language = {en},
	urldate = {2023-12-20},
	publisher = {arXiv},
	author = {Hendrycks, Dan and Burns, Collin and Basart, Steven and Zou, Andy and Mazeika, Mantas and Song, Dawn and Steinhardt, Jacob},
	month = jan,
	year = {2021},
	note = {arXiv:2009.03300 [cs]},
}

@misc{li_survey_2022,
	title = {A {Survey} on {Retrieval}-{Augmented} {Text} {Generation}},
	url = {http://arxiv.org/abs/2202.01110},
	language = {en},
	urldate = {2023-12-18},
	publisher = {arXiv},
	author = {Li, Huayang and Su, Yixuan and Cai, Deng and Wang, Yan and Liu, Lemao},
	month = feb,
	year = {2022},
	note = {arXiv:2202.01110 [cs]},
}

@article{kudithipudi_biological_2022,
	title = {Biological underpinnings for lifelong learning machines},
	volume = {4},
	issn = {2522-5839},
	url = {https://www.nature.com/articles/s42256-022-00452-0},
	doi = {10.1038/s42256-022-00452-0},
	language = {en},
	number = {3},
	urldate = {2023-11-22},
	journal = {Nature Machine Intelligence},
	author = {Kudithipudi, Dhireesha and Aguilar-Simon, Mario and Babb, Jonathan and Bazhenov, Maxim and Blackiston, Douglas and Bongard, Josh and Brna, Andrew P. and Chakravarthi Raja, Suraj and Cheney, Nick and Clune, Jeff and Daram, Anurag and Fusi, Stefano and Helfer, Peter and Kay, Leslie and Ketz, Nicholas and Kira, Zsolt and Kolouri, Soheil and Krichmar, Jeffrey L. and Kriegman, Sam and Levin, Michael and Madireddy, Sandeep and Manicka, Santosh and Marjaninejad, Ali and McNaughton, Bruce and Miikkulainen, Risto and Navratilova, Zaneta and Pandit, Tej and Parker, Alice and Pilly, Praveen K. and Risi, Sebastian and Sejnowski, Terrence J. and Soltoggio, Andrea and Soures, Nicholas and Tolias, Andreas S. and Urbina-Meléndez, Darío and Valero-Cuevas, Francisco J. and Van De Ven, Gido M. and Vogelstein, Joshua T. and Wang, Felix and Weiss, Ron and Yanguas-Gil, Angel and Zou, Xinyun and Siegelmann, Hava},
	month = mar,
	year = {2022},
	pages = {196--210},
}

@misc{keller_traveling_2023,
	title = {Traveling {Waves} {Encode} the {Recent} {Past} and {Enhance} {Sequence} {Learning}},
	url = {http://arxiv.org/abs/2309.08045},
	language = {en},
	urldate = {2023-11-22},
	publisher = {arXiv},
	author = {Keller, T. Anderson and Muller, Lyle and Sejnowski, Terrence and Welling, Max},
	month = sep,
	year = {2023},
	note = {arXiv:2309.08045 [cs]},
}

@article{macpherson_natural_2021,
	title = {Natural and {Artificial} {Intelligence}: {A} brief introduction to the interplay between {AI} and neuroscience research},
	volume = {144},
	issn = {08936080},
	shorttitle = {Natural and {Artificial} {Intelligence}},
	url = {https://linkinghub.elsevier.com/retrieve/pii/S0893608021003683},
	doi = {10.1016/j.neunet.2021.09.018},
	language = {en},
	urldate = {2023-11-22},
	journal = {Neural Networks},
	author = {Macpherson, Tom and Churchland, Anne and Sejnowski, Terry and DiCarlo, James and Kamitani, Yukiyasu and Takahashi, Hidehiko and Hikida, Takatoshi},
	month = dec,
	year = {2021},
	pages = {603--613},
}

@article{dally_high-performance_2015,
	title = {High-performance hardware for machine learning},
	volume = {2},
	journal = {Nips Tutorial},
	author = {Dally, William},
	year = {2015},
	pages = {3},
}

@article{schor_iedm_2022,
	title = {{IEDM} 2022: {Did} {We} {Just} {Witness} {The} {Death} {Of} {SRAM}?},
	url = {https://fuse.wikichip.org/news/7343/iedm-2022-did-we-just-witness-the-death-of-sram/},
	journal = {WikiChip Fuse},
	author = {Schor, David},
	month = sep,
	year = {2022},
}

@techreport{benigno_what_2022,
	type = {preprint},
	title = {What computations can be done with traveling waves in visual cortex?},
	url = {https://www.researchsquare.com/article/rs-1903144/v1},
	language = {en},
	urldate = {2023-04-26},
	institution = {In Review},
	author = {Benigno, Gabriel and Budzinski, Roberto and Davis, Zachary and Reynolds, John and Muller, Lyle},
	month = aug,
	year = {2022},
	doi = {10.21203/rs.3.rs-1903144/v1},
}

@article{aggarwal_visual_2022,
	title = {Visual evoked feedforward–feedback traveling waves organize neural activity across the cortical hierarchy in mice},
	volume = {13},
	issn = {2041-1723},
	url = {https://www.nature.com/articles/s41467-022-32378-x},
	doi = {10.1038/s41467-022-32378-x},
	language = {en},
	number = {1},
	urldate = {2023-04-26},
	journal = {Nature Communications},
	author = {Aggarwal, Adeeti and Brennan, Connor and Luo, Jennifer and Chung, Helen and Contreras, Diego and Kelz, Max B. and Proekt, Alex},
	month = aug,
	year = {2022},
	pages = {4754},
}

@article{iascone_whole-neuron_2020,
	title = {Whole-{Neuron} {Synaptic} {Mapping} {Reveals} {Spatially} {Precise} {Excitatory}/{Inhibitory} {Balance} {Limiting} {Dendritic} and {Somatic} {Spiking}},
	volume = {106},
	issn = {08966273},
	url = {https://linkinghub.elsevier.com/retrieve/pii/S0896627320301380},
	doi = {10.1016/j.neuron.2020.02.015},
	language = {en},
	number = {4},
	urldate = {2023-04-26},
	journal = {Neuron},
	author = {Iascone, Daniel Maxim and Li, Yujie and Sümbül, Uygar and Doron, Michael and Chen, Hanbo and Andreu, Valentine and Goudy, Finola and Blockus, Heike and Abbott, Larry F. and Segev, Idan and Peng, Hanchuan and Polleux, Franck},
	month = may,
	year = {2020},
	pages = {566--578.e8},
}

@article{klimesch_frequency_2018,
	title = {The frequency architecture of brain and brain body oscillations: an analysis},
	volume = {48},
	issn = {0953-816X, 1460-9568},
	shorttitle = {The frequency architecture of brain and brain body oscillations},
	url = {https://onlinelibrary.wiley.com/doi/10.1111/ejn.14192},
	doi = {10.1111/ejn.14192},
	language = {en},
	number = {7},
	urldate = {2023-04-10},
	journal = {European Journal of Neuroscience},
	author = {Klimesch, Wolfgang},
	month = oct,
	year = {2018},
	pages = {2431--2453},
}

@misc{gale_state_2019,
	title = {The {State} of {Sparsity} in {Deep} {Neural} {Networks}},
	url = {http://arxiv.org/abs/1902.09574},
	language = {en},
	urldate = {2023-04-07},
	publisher = {arXiv},
	author = {Gale, Trevor and Elsen, Erich and Hooker, Sara},
	month = feb,
	year = {2019},
	note = {arXiv:1902.09574 [cs, stat]},
}

@misc{bubeck_sparks_2023,
	title = {Sparks of {Artificial} {General} {Intelligence}: {Early} experiments with {GPT}-4},
	shorttitle = {Sparks of {Artificial} {General} {Intelligence}},
	url = {http://arxiv.org/abs/2303.12712},
	language = {en},
	urldate = {2023-03-29},
	publisher = {arXiv},
	author = {Bubeck, Sébastien and Chandrasekaran, Varun and Eldan, Ronen and Gehrke, Johannes and Horvitz, Eric and Kamar, Ece and Lee, Peter and Lee, Yin Tat and Li, Yuanzhi and Lundberg, Scott and Nori, Harsha and Palangi, Hamid and Ribeiro, Marco Tulio and Zhang, Yi},
	month = mar,
	year = {2023},
	note = {arXiv:2303.12712 [cs]},
}

@misc{su_roformer_2022,
	title = {{RoFormer}: {Enhanced} {Transformer} with {Rotary} {Position} {Embedding}},
	shorttitle = {{RoFormer}},
	url = {http://arxiv.org/abs/2104.09864},
	language = {en},
	urldate = {2023-03-29},
	publisher = {arXiv},
	author = {Su, Jianlin and Lu, Yu and Pan, Shengfeng and Murtadha, Ahmed and Wen, Bo and Liu, Yunfeng},
	month = aug,
	year = {2022},
	note = {arXiv:2104.09864 [cs]},
}
\end{document}